\documentclass[aos]{imsart}
\pdfoutput=1
\RequirePackage{amsthm,amsmath,amsfonts,amssymb}
\RequirePackage[authoryear]{natbib}
\RequirePackage{graphicx}

\usepackage{xr}
\makeatletter
\def\journal@name{}
\makeatother

\startlocaldefs
\theoremstyle{plain}
\newtheorem{theorem}{Theorem}
\newtheorem{proposition}{Proposition}
\newtheorem{lemma}{Lemma}
\theoremstyle{remark}
\newtheorem{definition}{Definition}
\def\thelonglist{C\arabic{longlist}}
\def\labellonglist{(\thelonglist)}
\graphicspath{{fig/}}
\def\bB{\mathbf{B}}
\def\bD{\mathbf{D}}
\def\bH{\mathbf{H}}
\def\bI{\mathbf{I}}
\def\bK{\mathbf{K}}
\def\bX{\mathbf{X}}
\def\ba{\mathbf{a}}
\def\bv{\mathbf{v}}
\def\bw{\mathbf{w}}
\def\bx{\mathbf{x}}
\def\by{\mathbf{y}}
\def\bz{\mathbf{z}}
\def\bzero{\mathbf{0}}
\def\B{\mathbb{B}}
\def\E{\mathbb{E}}
\def\R{\mathbb{R}}
\def\S{\mathbb{S}}
\def\cB{\mathcal{B}}
\def\cF{\mathcal{F}}
\def\cG{\mathcal{G}}
\def\cH{\mathcal{H}}
\def\cM{\mathcal{M}}
\def\cS{\mathcal{S}}
\def\cT{\mathcal{T}}
\def\ve{\varepsilon}
\def\bTheta{\boldsymbol{\Theta}}
\def\bbeta{\boldsymbol{\beta}}
\def\bve{\boldsymbol{\varepsilon}}
\def\btheta{\boldsymbol{\theta}}
\DeclareMathOperator*\argmin{\arg\min}
\DeclareMathOperator\diag{diag}
\DeclareMathOperator\sgn{sgn}
\DeclareMathOperator\tr{tr}
\newcommand\relph[1][=]{\mathrel{\phantom{#1}}}
\endlocaldefs

\begin{document}

\begin{frontmatter}
\title{Nonasymptotic theory for two-layer neural networks: Beyond the bias--variance trade-off}
\runtitle{Two-layer neural networks}

\begin{aug}
\author{\fnms{Huiyuan}~\snm{Wang}\ead[label=e1]{huiyuan.wang@pku.edu.cn}}
\and
\author{\fnms{Wei}~\snm{Lin}\ead[label=e2]{weilin@math.pku.edu.cn}}

\address{School of Mathematical Sciences and Center for Statistical Science, Peking University\printead[presep={,\ }]{e1,e2}}
\end{aug}

\begin{abstract}
Large neural networks have proved remarkably effective in modern deep learning practice, even in the overparametrized regime where the number of active parameters is large relative to the sample size. This contradicts the classical perspective that a machine learning model must trade off bias and variance for optimal generalization. To resolve this conflict, we present a nonasymptotic generalization theory for two-layer neural networks with ReLU activation function by incorporating scaled variation regularization. Interestingly, the regularizer is equivalent to ridge regression from the angle of gradient-based optimization, but plays a similar role to the group lasso in controlling the model complexity. By exploiting this ``ridge--lasso duality,'' we obtain new prediction bounds for all network widths, which reproduce the double descent phenomenon. Moreover, the overparametrized minimum risk is lower than its underparametrized counterpart when the signal is strong, and is nearly minimax optimal over a suitable class of functions. By contrast, we show that overparametrized random feature models suffer from the curse of dimensionality and thus are suboptimal.
\end{abstract}



\end{frontmatter}

\section{Introduction}
During the past decade, deep learning has demonstrated superiority over traditional machine learning techniques for representation learning and prediction in a wide variety of tasks, including object recognition in computer vision \citep{he2016deep}, machine translation and text generation in natural language processing \citep{sutskever2014sequence}, general game playing \citep{schrittwieser2020mastering}, and disease diagnosis in clinical research \citep{esteva2017dermatologist}. Many such successful applications build on large neural networks that operate in the overparametrized regime, where the number of parameters is relatively large compared to the number of training samples. For instance, the convolutional neural network AlexNet \citep{krizhevsky2012imagenet} had 60 million parameters trained on 1.2 million images; the more recent large language model GPT-3 was trained with 175 billion parameters and 300 billion training tokens \citep{brown2020language}.

Theoretical insights into overparametrized neural networks have been obtained from the optimization viewpoint \citep{arora2018optimization,soltanolkotabi2019theoretical}, suggesting that overparametrization can speed up convergence or improve the optimization landscape. The benefits of overparametrization to generalization in deep learning, however, remain mysterious. Numerical evidence indicates that deep neural networks easily fit random labels but still generalize well even without explicit regularization \citep{zhang2021understanding}. These empirical findings deeply challenge the conventional wisdom that optimal generalization should be achieved by trading off bias (or approximation error) and variance (or estimation error). The so-called ``double descent'' curve \citep{belkin2019reconciling} was proposed and conjectured as a ubiquitous phenomenon for unifying the generalization behaviors of machine learning models across the underparametrized and overparametrized regimes, but so far has not been theoretically justified for realistic neural networks.

While the notion of overparametrization is not new and has long been studied in high-dimensional statistics \citep{wainwright2019high}, there are some fundamental differences between the usual high-dimensional models and overparametrized deep learning models. In high-dimensional problems, although the number of parameters can be large or even exponentially growing, it is almost always assumed that certain parsimonious structures (e.g., sparsity and low-rankness) exist and can be exploited. For example, recent work has shown that minimum norm interpolators have near-optimal prediction risk and hence overfitting is not detrimental in linear regression when the parameters are sparse or the design matrix is low-rank \citep{bartlett2020benign,muthukumar2020harmless,hastie2022surprises,chinot2022robustness}. Such parsimony and the regularization for achieving it play two roles: (i) to control the model complexity for balancing bias and variance, and (ii) to ensure model identifiability so that prediction and estimation are essentially equivalent. These ideas, however, do not readily extend to overparametrized neural networks, because: (i) sparsity-inducing regularization is often not required in deep learning or not strong enough (e.g., in dropout) to bring the dimensionality down to a level below the sample size \citep{srivastava2014dropout}; and (ii) neural networks are intrinsically unidentifiable owing to weight space symmetry and many other equivalent parametrizations \citep[p.~277]{goodfellow2016deep}.

Neural networks are pure prediction algorithms in the sense of \citet{efron2020prediction}, which operate in a nonparametric and nonparsimonious way. The nonparametric view of neural networks was pioneered by \citet{barron1994approximation}, who derived risk bounds in terms of the network width for complexity-regularized two-layer sigmoidal networks. For different function classes and the now popular ReLU activation function, recent developments have shown that deep neural networks can deliver fast and near-minimax rates of convergence and circumvent the curse of dimensionality \citep{schmidt-hieber2020nonparametric,hayakawa2020minimax,farrell2021deep,kohler2021rate}. The architectural constraints imposed by this line of work, however, require the networks to be sparse or of small size, restricting the number of nonzero or active parameters to a smaller order than the sample size. Therefore, although these results demonstrate the efficiency of deep architectures, they are still confined to the underparametrized regime and do not go beyond the bias--variance trade-off.

Another line of work controls the model complexity of neural networks via norm-based regularization and obtains complexity and risk bounds in terms of various norms of the estimated network parameters. \citet{neyshabur2015norm} and \citet{golowich2020size}, among others, considered group norm and matrix norm regularization and derived size-independent bounds on the Rademacher complexity. However, as observed empirically by \citet{neyshabur2019role}, these complexity measures increase with the network size and do not correlate with the test error. As a result, they may lead to vacuous bounds for large networks and are not sufficient to explain the role of overparametrization. Recognizing these gaps, \citet{neyshabur2019role} presented complexity bounds that empirically decrease with the network size and could potentially explain the benefits of large networks. Nevertheless, norm-based complexity measures implicitly depend on the network size and the training process, which are difficult to analyze precisely and control tightly.

This paper contributes to the ongoing debate about the role of overparametrization in deep learning by developing a nonasymptotic theory for two-layer neural networks across the underparametrized and overparametrized regimes. Our theory is intended to be as transparent as possible, relying on no sparsity assumptions and giving rise to sharp risk bounds in terms of the sample size, dimensionality, and network width. Building on this theory, we aim to gain insight into the following questions:
\begin{itemize}
\item How does the network perform in the overparametrized regime differently from in the underparametrized regime?
\item How does the overparametrized minimum risk compare with its underparametrized counterpart and how far is it from optimal?
\end{itemize}

Specifically, suppose that we observe predictors $\bx_i\in\R^d$ and responses $y_i\in\R$ generated from the nonparametric regression model
\begin{equation}\label{eq:model}
y_i=f^*(\bx_i)+\ve_i,\quad i=1,\dots,n,
\end{equation}
where $f^*$ is an unknown function to be estimated and $\ve_i$ are random errors. Let $\sigma(z)=\max(z,0)$ be the rectified linear unit (ReLU) activation function \citep{jarrett2009best}. We consider a two-layer neural network with $m$ hidden units, $g(\cdot;\btheta)\colon\R^d\to\R$, of the form
\begin{equation}\label{eq:net}
g(\bx;\btheta)=\sum_{k=1}^m a_k\sigma(\bv_k^T\bx+b_k)
\end{equation}
with parameters $\btheta=(a_1,\dots,a_m,\bv_1^T,\dots,\bv_m^T,b_1,\dots,b_m)^T$. By appropriately restricting the function class to which $f^*$ belongs, we do not include an intercept in the output. Assumptions on $f^*$, $\bx_i$, and $\ve_i$ are detailed in Section \ref{ssec:assumpt}.

\begin{figure}
\centering
\includegraphics[width=.45\textwidth]{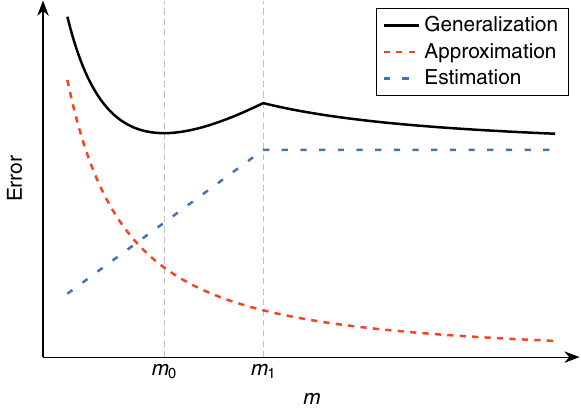}\quad
\includegraphics[width=.45\textwidth]{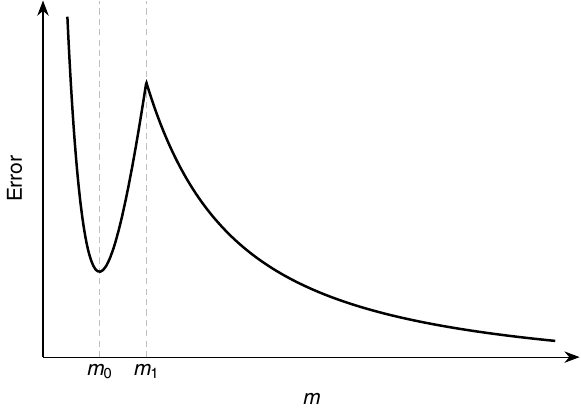}
\caption{Risk curves for varying network width $m$ from the prediction bound \eqref{eq:bound} with $\|f^*\|_\cS^2/\sigma_\ve^2=1$, $d=6$, and $n=1000$. The left panel shows the decomposition of prediction error into approximation and estimation errors. The right panel shows the same plot but with larger $m$, from which it is apparent that the second valley is lower than the first.}\label{fig:curve}
\end{figure}

By incorporating a \emph{scaled variation} regularizer to be defined in Section \ref{ssec:net}, our main result (Theorem \ref{thm:all_m}) shows that the prediction (or generalization) error of the regularized network estimator $g(\cdot;\widehat\btheta)$ is of order
\begin{equation}\label{eq:bound}
\|f^*\|_\cS^2m^{-(d+3)/d}+(\sigma_\ve^2+\|f^*\|_\cS^2)\min\biggl(\frac{md\log n}{n},\sqrt{\frac{d\log n}{n}}\biggr),
\end{equation}
where $\|f^*\|_\cS$ is the $\cS$-norm of $f^*$ (Definition \ref{def:Snorm}) and $\sigma_\ve^2$ is the variance of $\ve_i$. We emphasize that this result holds for all $m\ge1$ and any global minimizer of the regularized empirical risk. The prediction bound \eqref{eq:bound} consists of two terms: the first term represents the approximation error, which decreases with the network width $m$, while the second term represents the estimation error, which increases with $m$ up to some critical point $m_1\asymp\sqrt{n/(d\log n)}$ and thereafter stays constant. An intriguing consequence of this unusual trade-off is a double descent risk curve, as shown in Figure \ref{fig:curve}. To answer our question regarding optimality, we find the first valley or underparametrized minimum risk to be $O((d\log n/n)^{(d+3)/(2d+3)})$, which occurs at $m_0\asymp(n/(d\log n))^{d/(2d+3)}$, by matching the approximation and estimation errors in \eqref{eq:bound}. While this rate is slightly better than that of the second valley or overparametrized minimum risk, $O(\sqrt{d\log n/n})$, the asymptotic comparison can be reversed in finite samples, as shown in the right panel of Figure \ref{fig:curve}. When the signal-to-noise ratio $\|f^*\|_\cS^2/\sigma_\ve^2$ is large, the second valley tends to be lower than the first; a precise condition is given in \eqref{eq:kappa}. We further prove that the overparametrized minimum risk is nearly minimax rate-optimal over a suitable class of functions (Theorem \ref{thm:minimax}). By contrast, overparametrized random feature models suffer from the curse of dimensionality and thus are suboptimal (Proposition \ref{prop:rand_feat}). Overall, our results lend theoretical support to the benefits of overparametrization in deep learning and shed light on the currently debated double descent phenomenon.

Intuitively, the number of parameters or the network width $m$ is not an appropriate measure of model complexity for the network \eqref{eq:net} in the overparametrized regime, and one must seek alternatives. The idea of our approach to achieving model complexity control while allowing $m$ to grow unbounded is to exploit the \emph{ridge--lasso duality} of the scaled variation regularizer. On the one hand, by the positive homogeneity of the ReLU function, a reparametrization yields the equivalence of scaled variation regularization to ridge regression, which is known as (standard) \emph{weight decay} in deep learning \citep{krogh1991simple} and, in general, does not induce sparsity. On the other hand, a linearization of the ReLU function by parameter space partitioning transforms the regularized problem into a group lasso. This gives a key insight into the geometry of the global minima: the estimated network weights residing in the same region must be parallel to each other. Such collinearity greatly reduces the effective number of parameters and enables us to measure the model complexity in terms of the number of nonparallel directions. This implicit (within-group) formation and (between-group) breaking of symmetry lies at the heart of our theoretical analysis.

\subsection{Related work}
Although not the focus of this paper, approximation theory is often an integral part and first step of establishing statistical guarantees for neural networks. Sharp approximation bounds can be obtained for target functions that are well represented by two-layer neural networks, for which purpose various function spaces have been proposed. For sigmoidal networks, the seminal work of \cite{barron1993universal} considered a class of functions that have an integral representation involving the Fourier transform. The idea was further developed by, for example, \citet{bach2017breaking} and \citet{siegel2022sharp} to define variation spaces and norms for positively homogeneous activation functions, including ReLU. Other recent work \citep{ongie2020function,parhi2021banach} has introduced an equivalent characterization of the variation space for two-layer ReLU networks via the Radon transform and has related it to more classical function spaces \citep{parhi2022near}. Our choice of the target function space and its associated norm is similar to but slightly extends those of \citet{ongie2020function} and \citet{parhi2022near} to allow the identification of affine functions.

Generalization bounds have been derived for two-layer neural networks in certain variation spaces. Most of the existing work focuses on variational formulations of the empirical risk minimization problem. For example, \citet{bach2017breaking} and \citet{parhi2022near} considered a variational problem by constraining the network estimator to within a ball in the function space; \citet{parhi2022near} showed that such network estimators are nearly minimax optimal. A representer theorem of \citet{parhi2021banach} ensures the existence of a solution to the variational problem in the form of a finite-width network with a skip connection. However, the network width of the solution is required to be smaller than the sample size, thus providing no clue about the effect of overparametrization. One exception is the work of \citet{e2019priori}, which obtained generalization bounds for finite-width two-layer networks that allow the network width to grow unbounded. The $\ell_1$ path norm regularization that they adopted, however, induces sparsity in the network parameters, casting doubt on the implication of their results for intrinsically overparametrized networks.

Mean-field and neural tangent kernel theories are two popular frameworks for analyzing the training dynamics of two-layer neural networks in the infinite-width limit. The mean-field theory shows that the stochastic gradient descent dynamics of two-layer networks is asymptotically described by a nonlinear partial differential equation (PDE), and approximation results such as laws of large numbers and central limit theorems can be derived \citep{mei2018mean,sirignano2020mean,rotskoff2022trainability}. The generalization behavior of the PDE model, however, is difficult to study except in some specific examples. Under a different scaling, overparametrized two-layer networks are shown to behave as their linearizations at random initialization, and optimization and generalization properties can be investigated by exploiting the neural tangent kernel \citep{jacot2018neural} and the associated kernel methods. This ``lazy training'' regime \citep{chizat2019lazy} entails a large performance gap between realistic and linearized networks and hence does not explain the power of fully trained neural networks \citep{e2020comparative,ghorbani2021linearized}. \citet{dou2021training} went a step further and developed an adaptive theory for neural network training with data-adaptive kernels. Nevertheless, the impact of adaptivity on generalization remains unclear.

Since the conceptualization of the double descent curve by \citet{belkin2019reconciling}, several theoretical models and explanations have been developed for the phenomenon. A majority of the effort has focused on linear regression and, in particular, minimum norm interpolators and ridge estimators, and has recovered the phenomenon under specific generative models for the random predictors \citep[e.g.,][]{belkin2020two,hastie2022surprises,muthukumar2020harmless}. Random matrix theory is the backbone of most of these results, which concerns the high-dimensional asymptotic regime where $n,d\to\infty$ with $n\asymp d$. Similar asymptotics have been derived for random feature models \citep{mei2022generalization} and classification problems \citep{deng2022model,liang2022precise}. \citet{li2021multi} and \citet{liang2020multiple} demonstrated a ``multiple descent'' phenomenon in infinite-dimensional linear regression and kernel ridgeless regression. Despite these important developments, it still seems difficult to isolate a general mechanism for the emergence of double descent from the oversimplified model assumptions and asymptotic regimes. Also, it remains elusive how these theories extend to realistic neural networks and fit in with our current understanding of the bias--variance trade-off \citep{geman1992neural,derumigny2023lower}.

\subsection{Organization of the paper}
Section \ref{sec:prelim} introduces the definitions of two-layer ReLU networks and the target function class. Theoretical assumptions and approximation properties are also described. Section \ref{sec:dual} presents the regularized estimation framework and formalizes the ridge--lasso duality. Our main results, including nonasymptotic generalization guarantees and minimax optimality, are developed in Section \ref{sec:main}. Section \ref{sec:rand_feat} discusses the random feature model and points out its suboptimality. Section \ref{sec:disc} provides some further discussion. Proofs are deferred to the Appendix and Supplemental Material.

\section{Preliminaries}\label{sec:prelim}

\subsection{Notation}
For $1\le q<\infty$, let $\|\cdot\|_q$ denote the $\ell_q$-norm of a vector. Let $\B^d$ and $\S^{d-1}$ be the $\ell_2$ unit ball and unit sphere, respectively, in $\R^d$. For a matrix $\bB=(\bbeta_1,\dots,\bbeta_J)$, define the $\ell_{2,1}$-norm $\|\bB\|_{2,1}=\sum_{j=1}^J\|\bbeta_j\|_2$. Denote by $\cM(D)$ the set of signed measures $\alpha$ on $D$ with finite total variation $|\alpha|(D)$. In particular, the Dirac measure $\delta_\bx\in\cM(D)$ if $\bx\in D$. For a function $f$, let $\|f\|_{L_\infty(D)}$ denote the $L_\infty$-norm on $D$, and $\|f\|_2$ and $\|f\|_n$ the $L_2$-norm under the distribution $\mu$ and its empirical counterpart, respectively.

\subsection{Neural networks and the target function class}\label{ssec:net}
We consider the two-layer neural network $g(\cdot;\btheta)$ with ReLU activation function and width $m$ given by \eqref{eq:net}. Let $\bTheta_m$ denote the parameter space. Define the \emph{scaled variation norm} of the finite-width network $g(\cdot;\btheta)$ by
\begin{equation}\label{eq:svnorm}
\nu(\btheta)=\sum_{k=1}^m|a_k|\|\bw_k\|_2,
\end{equation}
where $\bw_k=(\bv_k^T,b_k)^T$. This regularizer was also considered by \citet{parhi2021banach} and \citet{parhi2022near} for two-layer ReLU networks. It coincides with the $\ell_1$ path norm proposed by \citet{neyshabur2015norm} when $d$ degenerates to zero. For any $d\ge1$, however, it is not separable in the first-layer weights and hence not a path norm. We will show in Section~\ref{sec:dual} that the scaled variation regularizer \eqref{eq:svnorm} has some desirable properties that are key to our theoretical analysis.

The network \eqref{eq:net} has an integral representation with respect to a discrete signed measure. Specifically, if we define $\alpha_m=\sum_{k=1}^ma_k\delta_{\bw_k}$, then
\[
g(\bx;\btheta)=\int_{\R^{d+1}}\bigl(\sigma(\bv^T\bx+b)-\sigma(b)\bigr)\,d\alpha_m(\bw)+g(\bzero;\btheta).
\]
Motivated by this observation, we can naturally represent an infinite-width two-layer ReLU network associated with a signed measure $\alpha\in\cM(\R^{d+1})$ as
\[
g_\alpha(\bx)=\int_{\R^{d+1}}\bigl(\sigma(\bv^T\bx+b)-\sigma(b)\bigr)\,d\alpha(\bw)+g_\alpha(\bzero).
\]
For $g_\alpha(\cdot)$ to be well defined, a sufficient condition is
\[
\int_{\R^{d+1}}\|\bv\|_2\,d|\alpha|(\bw)<\infty,
\]
since by the Lipschitz continuity of the ReLU function, $|\sigma(\bv^T\bx+b)-\sigma(b)|\le|\bv^T\bx|\le\|\bv\|_2\|\bx\|_2$. Treating functions that differ by a constant as identical, we consider the space of functions modulo constants
\begin{equation}\label{eq:def_G}
\cG=\biggl\{\bx\mapsto\int_{\R^{d+1}}\bigl(\sigma(\bv^T\bx+b)-\sigma(b)\bigr)\,d\alpha(\bw):\int_{\R^{d+1}}\|\bv\|_2\,d|\alpha|(\bw)<\infty\biggr\}.
\end{equation}
Interestingly, there is a one-to-one correspondence between $\cG$ and $\cM_2(\R^{d+1})\equiv\bigl\{\alpha\in\cM(\R^{d+1}):\int_{\R^{d+1}}\|\bv\|_2\,d|\alpha|(\bw)<\infty\bigr\}$. Moreover, functions in $\cG$ are exactly those that can be approximated by two-layer ReLU networks with finite scaled variation norm. A formal statement is given by Proposition \ref{prop:functionspace2} in the Supplementary Material. To equip the function space $\cG$ with a norm, we introduce the following definition.

\begin{definition}\label{def:Snorm}
The $\cS$-norm of $f\in\cG$ is defined as $\|f\|_\cS=\int_{\R^{d+1}}\|\bv\|_2\,d|\alpha_f|(\bw)$, where the signed measure $\alpha_f\in\cM_2(\R^{d+1})$ is uniquely determined by
\[
f(\bx)=\int_{\R^{d+1}}\bigl(\sigma(\bv^T\bx+b)-\sigma(b)\bigr)\,d\alpha_f(\bw)+f(\bzero).
\]
\end{definition}

Clearly, the $\cS$-norm is a functional version of the scaled variation norm except for the omission of the bias term owing to the centering by $\sigma(b)$. In fact, the $\cS$-norm of a finite-width two-layer ReLU network $g(\cdot;\btheta)$ is $\|g(\cdot;\btheta)\|_\cS=\sum_{k=1}^m|a_k|\|\bv_k\|_2$, which is bounded above by the scaled variation norm \eqref{eq:svnorm}.

Our definition of the target function space is inspired by and related to several previously studied spaces for two-layer neural networks. In particular, $\cG$ is equivalent to the bounded variation spaces in the Radon domain considered by \citet{ongie2020function} and \citet{parhi2021banach} and the variation spaces considered by \citet{bach2017breaking} and \citet{siegel2022sharp}, which in turn contain the spectral Barron spaces and Sobolev spaces \citep{klusowski2018approximation,parhi2022near}. Our definition of the $\cS$-norm is more transparent in that it is defined explicitly as a functional of $\alpha_f$, a uniquely determined signed measure. Moreover, it slightly improves on previous proposals in several respects. Notably, for an affine function $f_{\bbeta}(\bx)=\bbeta^T\bx+c$, $\|f_{\bbeta}\|_\cS=2\|\bbeta\|_2$ instead of being zero. This has two important consequences: (i) the $\cS$-norm is a norm rather than a seminorm; and (ii) there is no need to introduce a skip connection in a representer theorem \citep[cf.][]{ongie2020function,parhi2021banach}. The latter is compatible with deep learning practice since skip connections are only necessary in deep neural networks such as residual networks \citep{he2016deep}. More mathematical details can be found in Section \ref{sec:target_function} of the Supplementary Material.

\subsection{Assumptions}\label{ssec:assumpt}
We consider the nonparametric regression model \eqref{eq:model} and impose the following conditions:
\begin{longlist}
\item $f^*\in\cG_M\equiv\{f\in\cG:\|f\|_\cS\le M\}$ for some constant $M>0$;\label{cond:f}
\item $\bx_i\sim\mu$ independently, where $\mu$ is supported in $\B^d$;\label{cond:x}
\item $\ve_i\sim N(0,\sigma_\ve^2)$ independently and are independent of $\bx_i$.\label{cond:eps}
\end{longlist}

Condition \eqref{cond:x} is mild and standard in the machine learning literature since the predictors are usually bounded and can be normalized. Under Condition \eqref{cond:x}, it suffices to consider the restrictions of functions in $\cG$ to $\B^d$; denote the space of such restrictions by $\cG(\B^d)$. An important consequence from Corollary \ref{cor:restriction} in the Supplementary Material is that, for any $f\in\cG(\B^d)$, there exists a signed measure $\widetilde\alpha_f\in\cM(\S^{d-1}\times[-1,1])$ such that
\begin{equation}\label{eq:int_rep}
f(\bx)=\int_{\S^{d-1}\times[-1,1]}\sigma(\bv^T\bx+b)\,d\widetilde\alpha_f(\bw)+c,\quad\bx\in\B^d.
\end{equation}
Compared with a similar integral representation in \citet[Remark 3]{parhi2022near}, note that no skip connection appears in \eqref{eq:int_rep}. Thus, functions in $\cG(\B^d)$ have a simpler integral representation
\[
\cG(\B^d)=\biggl\{\bx\mapsto\int_{\S^{d-1}\times[-1,1]}\sigma(\bv^T\bx+b)\,d\alpha(\bw):|\alpha|(\S^{d-1}\times[-1,1])<\infty\biggr\},
\]
which will allow us to obtain a sharp approximation bound.

\subsection{Approximation properties}
Approximation rates for two-layer neural networks of width $m$ have been derived in various function spaces. A classical probabilistic argument, first applied to neural networks by \citet{barron1993universal}, yields an approximation rate of $O(1/\sqrt{m})$ in the $L_2$-norm; see also \citet{jones1992simple} and \citet{siegel2020approximation}. The approximation rate has been improved by \citet{makovoz1996random}, \citet{bach2017breaking}, and \citet{klusowski2018approximation}, among others. In particular, \citet{bach2017breaking} obtained an $O(m^{-(d+3)/(2d)})$ rate in the $L_\infty$-norm by using a result from geometric discrepancy theory \citep{matousek1996improved}; \citet{siegel2022sharp} showed that this rate is sharp and not improvable. We have the following approximation result for functions in $\cG_M$, which is a direct consequence of \citet[Proposition 1]{bach2017breaking} and the integral representation \eqref{eq:int_rep}.

\begin{theorem}\label{thm:approx}
For any $f\in\cG_M$, there exists a network $g(\cdot;\btheta)$ of width $m$ in the form of \eqref{eq:net} such that $\nu(\btheta)\le6\|f\|_\cS$ and
\[
\|f-g(\cdot;\btheta)\|_{L_\infty(\B^d)}\le C\|f\|_\cS m^{-(d+3)/(2d)}
\]
for some constant $C>0$ depending only on $d$.
\end{theorem}

The construction in Theorem \ref{thm:approx} has a tight control on the scaled variation norm of the network parameter. This suggests using the scaled variation norm as a regularizer for the network estimation problem, as we will discuss in the next section.

\section{Methodology and the ridge--lasso duality}\label{sec:dual}
In this section we introduce our regularized estimation problem and formalize the notion of the ridge--lasso duality through two different reparametrizations.

\subsection{Regularized estimation}
In order to learn $f^*$ from the training sample, we adopt the penalized empirical risk minimization (ERM) framework and seek to minimize
\[
J_n(\btheta;\lambda)=\frac{1}{2n}\sum_{i=1}^n\bigl(y_i-g(\bx_i;\btheta)\bigr)^2+\lambda\nu(\btheta),
\]
where $g(\cdot;\btheta)$ is the two-layer ReLU network of width $m$ in \eqref{eq:net}, $\nu(\btheta)$ is the scaled variation norm in \eqref{eq:svnorm}, and $\lambda>0$ is a regularization parameter. The regularized network estimator is given by $g(\cdot;\widehat\btheta)$, where
\begin{equation}\label{eq:opt}
\widehat\btheta=\argmin_{\btheta\in\bTheta_m}J_n(\btheta;\lambda).
\end{equation}

In a related work, \citet{parhi2022near} studied a variational problem in the variation space associated with two-layer ReLU networks, where regularization is imposed as a constraint on the variation norm of the network function. A representer theorem guarantees the existence of a finitely supported solution of width $m\le n-(d+1)$ to the variational problem. However, the finite-dimensional network learning problem is equivalent to the variational problem only when $m\ge n-(d+1)$. See their Theorem 5 and Section III.B. Therefore, their results still fall within the underparametrized regime and do not fully characterize the influence of the network width. By contrast, we provide a direct attack on the finite-dimensional network learning problem \eqref{eq:opt} and allow the network width $m$ to vary freely.

\subsection{Equivalence to ridge regression}
In this and the next subsections, we explore some useful reformulations of the optimization problem \eqref{eq:opt}, which allow the scaled variation regularizer, when coupled with the ReLU function, to inherit some crucial properties from ridge regression \citep{hoerl2020ridge} and the group lasso \citep{yuan2006model}, two familiar forms of regularization in statistics. We start by recasting \eqref{eq:opt} as the $\ell_2$-regularized ERM problem
\begin{equation}\label{eq:ridge}
\widehat\btheta_{\ell_2}=\argmin_{\btheta\in\bTheta_m}\Biggl\{\frac{1}{2n}\sum_{i=1}^n\bigl(y_i-g(\bx_i;\btheta)\bigr)^2+\frac{\lambda}{2}\sum_{k=1}^m(a_k^2+\|\bw_k\|_2^2)\Biggr\}.
\end{equation}

To see this, consider the reparametrization $\widetilde\btheta=\cT_1(\btheta)$ defined by
\[
\widetilde{a}_k=a_k\sqrt{\frac{\|\bw_k\|_2}{|a_k|}},\qquad\widetilde\bw_k=\bw_k\sqrt{\frac{|a_k|}{\|\bw_k\|_2}}
\]
if $|a_k|\|\bw_k\|_2\ne0$, and $(\widetilde{a}_k,\widetilde\bw_k^T)=\bzero$ otherwise. After the reparametrization, we have $|\widetilde{a}_k|=\|\widetilde\bw_k\|_2$ and the regularizer becomes
\[
\sum_{k=1}^m|\widetilde{a}_k|\|\widetilde\bw_k\|_2=\frac{1}{2}\sum_{k=1}^m(\widetilde{a}_k^2+\|\widetilde\bw_k\|_2^2).
\]
Meanwhile, the positive homogeneity of the ReLU function implies that $a_k\sigma((\bx_k^T,1)\bw_k)=\widetilde{a}_k\sigma((\bx_k^T,1)\widetilde\bw_k)$, so that the network function is invariant under the reparametrization. Note further that any solution $\widehat\btheta_{\ell_2}$ to the problem \eqref{eq:ridge} must satisfy $\widehat\btheta_{\ell_2}=\cT_1(\widehat\btheta_{\ell_2})$, because otherwise it could be improved by a rescaling. Using these facts, we obtain the following equivalence result.

\begin{proposition}\label{prop:ridge}
Any solution $\widehat\btheta_{\ell_2}$ to the optimization problem \eqref{eq:ridge} is a solution to the problem \eqref{eq:opt}. Conversely, if $\widehat\btheta$ is a solution to the optimization problem \eqref{eq:opt}, then $\cT_1(\widehat\btheta)$ is a solution to the problem \eqref{eq:ridge}.
\end{proposition}

Proposition \ref{prop:ridge} says that the solutions to the $\ell_2$-regularized problem lie on a submanifold of the solution manifold of the original problem that is invariant under the reparametrization $\cT_1$. What is the implication of this equivalence for neural network training dynamics with, for example, gradient descent? The following result assures us that the gradient flow trajectories for the two problems are indeed identical when initialized with a reparametrization $\cT_1$.

\begin{proposition}\label{prop:grad_flow}
Consider the gradient flow for the optimization problem \eqref{eq:opt} defined by
\[
\frac{d}{dt}\btheta(t)=-\nabla_{\btheta}J_n(\btheta(t);\lambda)
\]
and for the problem \eqref{eq:ridge} defined similarly, both initialized at $\btheta(0)=\cT_1(\btheta_0)$ for an arbitrary $\btheta_0\in\bTheta_m$. Then the trajectories of the two gradient flows coincide.
\end{proposition}

Observations similar to Proposition \ref{prop:ridge} have been noted in slightly different forms by, for example, \citet[Theorem 1]{neyshabur2015search} and \citet[Theorem 8]{parhi2021banach}. Initialization with the reparametrization $\cT_1$ and its stationarity along the gradient flow have been exploited by \citet{dou2021training} for studying the $\ell_2$-regularized ERM problem, where it is referred to as a ``balanced condition.'' The messages of the above results are twofold. First, since $\ell_2$ regularization does not induce entrywise sparsity in the parameters (but see \citet{srebro2004maximum} for an unusual example where it does induce sparsity in spectral structures), we are assured that a sufficiently wide two-layer network can be intrinsically overparametrized. Second, several implicit regularization strategies for deep learning, such as noise injection and early stopping, have been shown to be equivalent to $\ell_2$ regularization \citep{bishop1995training,sjoberg1995overtraining}, which may help bridge the gap between our method and implicit regularization.

\subsection{Connection to the group lasso}\label{ssec:glasso}
One major obstacle in analyzing the generalization performance of neural networks is the excessive redundancy and nonidentifiability of the network parameters under the usual nonconvex formulation. The ReLU activation function, on the other hand, is simple enough in that it reduces to a linear function once the sign of $\bv_k^T\bx+b_k$ is fixed. This naturally suggests a partitioning of the parameter space $\R^{d+1}$ for $\bw$ by certain hyperplanes into regions within which the signs of $\bx_i^T\bv+b$ are all determined. The partition will then allow us to reveal a strong symmetry in the estimated network parameters and recast the optimization problem \eqref{eq:opt} in a group lasso form, which will be convenient for the derivation of generalization properties in the next section.

Specifically, denote by $\bX=((\bx_1^T,1)^T,\dots,(\bx_n^T,1)^T)^T$ the $n\times(d+1)$ design matrix, and $\bD=\diag(I(\bX\bw\ge0))$ the diagonal indicator matrix for the positivity of $\bX\bw$. Consider the hyperplanes in $\R^{d+1}$ passing through the origin and orthogonal to $\bx_i$, defined by $\bx_i^T\bv+b=0$. These $n$ hyperplanes divide the parameter space $\R^{d+1}$ into finitely many regions, denoted by $R_1,\dots,R_p$, such that $\bD$ stays constant over (the interior of) each $R_j$. It is well known \citep[Theorem 2]{cover1965geometrical} that the number of these regions
\begin{equation}\label{eq:p_bound}
p\le2\sum_{j=0}^d\binom{n-1}{j}\le2n^d,
\end{equation}
where the first upper bound is sharp when $\bX$ has full rank. Taking into account the sign of $a$, we thus partition the parameter space $\R^{d+2}$ for $(a,\bw^T)^T$ into $2p$ regions
\[
Q_j=[0,\infty)\times R_j,\quad Q_{p+j}=(-\infty,0)\times R_j,\quad j=1,\dots,p,
\]
and define $\bD_{p+j}=-\bD_j$. Clearly, $R_j$ and $Q_j$ are convex cones. The linearity of the ReLU function over each $Q_j$ and the optimality of $\widehat\btheta$ entail the following collinearity property.

\begin{proposition}\label{prop:sym}
For any solution $\widehat\btheta=(\widehat{a}_1,\dots,\widehat{a}_m,\widehat\bw_1^T,\dots,\widehat\bw_m^T)^T$ to the optimization problem \eqref{eq:opt}, if $(\widehat{a}_k,\widehat\bw_k^T)^T$ and $(\widehat{a}_\ell,\widehat\bw_\ell^T)^T$ lie in the interior of the same cone $Q_j$, then $\widehat\bw_k$ and $\widehat\bw_\ell$ must be collinear, that is, $\widehat\bw_k=c_0\widehat\bw_\ell$ for some constant $c_0>0$.
\end{proposition}

Define $S_j=\{1\le k\le m:(a_k,\bw_k^T)^T\in Q_j\}$, $s_j=1$, and $s_{p+j}=-1$ for $j=1,\dots,p$. To understand why the collinearity must hold, note that the ``conewise collinearization'' $\widetilde\btheta=\cT_2(\btheta)$ defined by
\begin{equation}\label{eq:collin}
\widetilde{a}_k=s_j,\quad\widetilde\bw_k=\frac{1}{|S_j|}\sum_{\ell\in S_j}|a_\ell|\bw_\ell,\quad k\in S_j
\end{equation}
does not change the value of the network function on the training sample, but would yield a smaller scaled variation norm by the triangle inequality if the network weights in $Q_j$ were not all collinear. Proposition \ref{prop:sym} provides a useful geometric insight into the regularization effect of scaled variation norm: it favors the most symmetric (yet not parsimonious) representation among many equivalent parametrizations within the same cone.

The parameter redundancy suggested by Proposition \ref{prop:sym} motivates us to collect the network weights falling within the same cone and define the aggregated parameters $\bB(\btheta)=(\bbeta_1(\btheta),\dots,\bbeta_{2p}(\btheta))$ with
\[
\bbeta_j(\btheta)=\sum_{k\in S_j}|a_k|\bw_k.
\]
With this new set of parameters, the network function on the training sample can be written in the linear form
\begin{equation}\label{eq:linear_form}
\sum_{k=1}^ma_k\sigma(\bX\bw_k)=\sum_{j=1}^{2p}\bD_j\bX\bbeta_j(\btheta),
\end{equation}
where $\sigma(\cdot)$ applies componentwise. For any $\btheta\in\bTheta_m$, the triangle inequality implies that
\[
\|\bB(\btheta)\|_{2,1}=\sum_{j=1}^{2p}\|\bbeta_j(\btheta)\|_2\le\sum_{j=1}^{2p}\sum_{k\in S_j}|a_k|\|\bw_k\|_2=\nu(\btheta),
\]
where the equality holds under the reparametrization $\cT_2$. In particular, since the estimator $\widehat\btheta$ satisfies the collinearity property, we can replace $\nu(\widehat\btheta)$ by $\|\bB(\widehat\btheta)\|_{2,1}$ and reformulate \eqref{eq:opt} as a group lasso problem. Denote by $\by=(y_1,\dots,y_n)^T$ the response vector. We summarize the above discussion in the following proposition.

\begin{proposition}\label{prop:glasso}
For any $\btheta\in\bTheta_m$, the reparametrization $\widetilde\btheta=\cT_2(\btheta)$ defined in \eqref{eq:collin} satisfies $g(\bx_i;\widetilde\btheta)=g(\bx_i;\btheta)$ for $i=1,\dots,n$ and $\|\bB(\widetilde\btheta)\|_{2,1}=\nu(\widetilde\btheta)\le\nu(\btheta)$. Furthermore, the solution $\widehat\btheta$ to the optimization problem \eqref{eq:opt} satisfies
\[
J_n(\widehat\btheta;\lambda)=\frac{1}{2n}\biggl\|\by-\sum_{j=1}^{2p}\bD_j\bX\bbeta_j(\widehat\btheta)\biggr\|_2^2+\lambda\|\bB(\widehat\btheta)\|_{2,1}.
\]
\end{proposition}

The group lasso formulation allows for borrowing ideas from high-dimensional statistics to derive generalization bounds. We emphasize, however, that the parameter space partition and the resulting group structure are data-adaptive and not known a priori. Hence, despite the connection to the group lasso, two-layer ReLU networks are radically different from linear models and hold the potential for better generalization.

Similar connections between $\ell_2$-regularized two-layer ReLU networks and the group lasso have been explored by \citet{pilanci2020neural} and \citet{wang2022hidden} from a purely optimization standpoint. A complete equivalence result, however, requires $m$ to be sufficiently large; see Theorem 1 of \citet{pilanci2020neural}. Our key observation is that for our purposes it suffices to have the weaker result of Proposition \ref{prop:glasso}, which places no restriction on the minimum network width.

\section{Main results}\label{sec:main}
In this section we establish statistical guarantees for two-layer ReLU networks. In Section \ref{ssec:upper} we present nonasymptotic bounds on the prediction error of the regularized network estimator, and in Section \ref{ssec:lower} show that they are nearly minimax optimal.

\subsection{Nonasymptotic generalization bounds}\label{ssec:upper}
For the nonparametric regression model \eqref{eq:model} and the regularized network estimator $g(\cdot;\widehat\btheta)$ defined by \eqref{eq:opt}, we are interested in bounding the empirical error
\[
\|g(\cdot;\widehat\btheta)-f^*\|_n^2=\frac{1}{n}\sum_{i=1}^n\bigl(g(\bx_i;\widehat\btheta)-f^*(\bx_i)\bigr)^2
\]
in the fixed design case, and the prediction (or generalization) error
\[
\|g(\cdot;\widehat\btheta)-f^*\|_2^2=\E_{\bx\sim\mu}\bigl(g(\bx;\widehat\btheta)-f^*(\bx)\bigr)^2
\]
in the random design case. Our main techniques for proving the nonasymptotic bounds are inspired by and synthesize those in previous work on high-dimensional linear models and two-layer neural networks. We first note that the technical arguments best suited to the underparametrized and overparametrized regimes may be rather different. For underparametrized networks, complexity control via metric entropy \citep[e.g.,][]{barron1994approximation,parhi2022near} can be effective and give sharp bounds. Moving into the overparametrized regime, however, entropy-based bounds tend to be too loose and pessimistic since they do not take into account the parameter redundancy growing with the network width. We therefore turn to the group lasso formulation outlined in Section \ref{ssec:glasso} and borrow ideas from (group) $\ell_1$-regularized linear regression and norm-based complexity control. Our first result concerns the empirical error of the regularized network estimator.

\begin{theorem}\label{thm:emp_err}
Under Conditions \eqref{cond:f} and \eqref{cond:eps} and the assumption that $\max_i\|\bx_i\|_2\le1$, the regularized network estimator $g(\cdot;\widehat\btheta)$ with $\lambda=C_1\sigma_\ve\sqrt{d\log n/n}$ satisfies
\begin{equation}\label{eq:emp_err1}
\|g(\cdot;\widehat\btheta)-f^*\|_n^2\le C\biggl\{\|f^*\|_\cS^2m^{-(d+3)/d}+(\sigma_\ve^2+\|f^*\|_\cS^2)\sqrt{\frac{d\log n}{n}}\biggr\}
\end{equation}
with probability at least $1-O(n^{-C_2})$, and
\begin{equation}\label{eq:emp_err2}
\E\|g(\cdot;\widehat\btheta)-f^*\|_n^2\le C\biggl\{\|f^*\|_\cS^2m^{-(d+3)/d}+(\sigma_\ve^2+\|f^*\|_\cS^2)\sqrt{\frac{d\log n}{n}}\biggr\},
\end{equation}
for some constants $C_1,C_2,C>0$.
\end{theorem}

Throughout this section, the constants are independent of $m$ and $n$, but may depend on $d$, $M$, and $\sigma_\ve$; we have suppressed the dependence for simplicity, which can be made explicit by inspecting our proofs. Our technique for proving Theorem \ref{thm:emp_err} differs from the standard group lasso theory for sparse linear regression in two aspects. First, it requires an extension of the theory to the case where the linear model is only approximate, as discussed in, for example, \citet[Section 6.2.3]{buhlmann2011statistics}. Here the linear model represents the reparametrized two-layer neural network, whose approximation error has been given by Theorem \ref{thm:approx}. Second, and more importantly, there is no guarantee that this linear model will be sparse or its design will satisfy a compatibility or restricted eigenvalue condition that is often imposed in high-dimensional linear regression. As a result, we can only prove a prediction bound at a slower rate, which is analogous to those for the lasso obtained by \citet[Corollary 6.1]{buhlmann2011statistics} and \citet{bartlett2012regularized}.

The error bounds in Theorem \ref{thm:emp_err} decompose into a bias term or approximation error that arises from using a finite-width neural network to approximate the nonparametric model \eqref{eq:model}, and a variance term or estimation error that accounts for the variability in estimating the finite-width network. The most surprising fact about this decomposition is that there is \emph{no} trade-off between the two terms: as the network width $m$ increases, the bias always decreases while the variance remains constant. To appreciate why this is possible, note first that the variance scales as $O(\sqrt{\log p/n})$ as a consequence of the lack of parameter identifiability. Moreover, the \emph{effective} dimension $p$ is bounded by $O(n^d)$ from \eqref{eq:p_bound}, which does not depend on the network width $m$. In fact, when the design matrix $\bX$ is of rank $r<n$, one can further replace $d$ by $r$ \citep{cover1965geometrical}. In other words, no matter how large $m$ grows, the number of distinct (nonparallel) features extracted by the first layer of the network is finite, leading to an upper bound for the variance. This result extends beyond the classical bias--variance trade-off and demonstrates the virtues of overparametrization in two-layer neural networks.

Combining Theorem \ref{thm:emp_err} with a maximal inequality, we obtain similar bounds on the prediction error of the regularized network estimator.

\begin{theorem}\label{thm:pred_err}
Under Conditions \eqref{cond:f}--\eqref{cond:eps}, if $m\ge C_1\{n/(d\log n)\}^{d/(2(d+3))}$, then the regularized network estimator $g(\cdot;\widehat\btheta)$ with $\lambda=\lambda_1\equiv C_2\sigma_\ve\sqrt{d\log n/n}$ satisfies
\begin{equation}\label{eq:pred_err1}
\|g(\cdot;\widehat\btheta)-f^*\|_2^2\le C\biggl\{\|f^*\|_\cS^2m^{-(d+3)/d}+(\sigma_\ve^2+\|f^*\|_\cS^2)\sqrt{\frac{d\log n}{n}}\biggr\}
\end{equation}
with probability at least $1-O(n^{-C_3})$, and
\begin{equation}\label{eq:pred_err2}
\E\|g(\cdot;\widehat\btheta)-f^*\|_2^2\le C\biggl\{\|f^*\|_\cS^2m^{-(d+3)/d}+(\sigma_\ve^2+\|f^*\|_\cS^2)\sqrt{\frac{d\log n}{n}}\biggr\},
\end{equation}
for some constants $C_1,C_2,C_3,C>0$ and sufficiently large $n$.
\end{theorem}

It is worthwhile to compare our results with those in the literature on overparametrized two-layer ReLU networks. Recent research has focused on the neural tangent kernel regime and showed that sufficiently wide two-layer ReLU networks trained by gradient descent with random initialization achieve a generalization error of $O(n^{-1/2})$ up to logarithmic factors; see, for example, \citet{arora2019fine}, \citet{e2020comparative}, and \citet{ji2020polylogarithmic}. While these results deliver roughly the same rates as ours, the target functions they considered fall in a certain reproducing kernel Hilbert space, which constitutes only a small subset of our target function space. In addition, our analysis is algorithm-independent and is valid for any global optimum.

\citet{e2019priori} considered explicit regularization for two-layer ReLU networks and obtained generalization bounds of $O(m^{-1}+n^{-1/2})$ up to logarithmic factors, which are of a similar nature to ours. However, several differences are notable. First, they employed the $\ell_1$ path norm, which penalizes on the $\ell_1$-norm of the first-layer weights and promotes sparsity. Accordingly, they considered the so-called Barron space
\[
\cB_2=\biggl\{\bx\mapsto\int_{\S_1^{d-1}}a(\bv)\sigma(\bv^T\bx)\,d\rho(\bv):\int_{\S_1^{d-1}}a(\bv)^2\,d\rho(\bv)<\infty\biggr\},
\]
where $\S_1^{d-1}$ is the $\ell_1$ unit sphere in $\R^d$. To compare with our definition of $\cG$ in \eqref{eq:def_G}, let $d\alpha_\rho(\bw)=a(\bv)I(\bv\in\S_1^{d-1},b=0)\,d\rho(\bv)$, where $I(\cdot)$ is the indicator function. Then
\[
\int_{\R^{d+1}}\|\bv\|_2\,d|\alpha_\rho|(\bw)\le\sqrt{d}\int_{\S_1^{d-1}}|a(\bv)|\,d\rho(\bv)\le\sqrt{d}\biggl(\int_{\S_1^{d-1}}a(\bv)^2\,d\rho(\bv)\biggr)^{1/2}<\infty
\]
by the Cauchy--Schwarz inequality. Thus, we see that $\cB_2$ is a subset of our target function space $\cG$. Furthermore, they resorted to a truncated risk to deal with the noisy case, which introduces some technicalities that seem unnecessary.

The group lasso approach and the size-independent upper bound \eqref{eq:p_bound} for $p$, albeit effective in the overparametrized regime, tend to overestimate the variance for sufficiently narrow networks. In this case, a standard metric entropy argument may be more appropriate and give a sharper estimate of the variance that increases with the network width. Adapting this argument to our regularization problem yields the following result, which demonstrates a classical bias--variance trade-off.

\begin{theorem}\label{thm:underparam}
Under Conditions \eqref{cond:f}--\eqref{cond:eps}, if $m<n/(d\log n)$, then the regularized network estimator $g(\cdot;\widehat\btheta)$ with $\lambda=\lambda_2\equiv C_1\sigma_\ve\max(m^{-(d+3)/d},md\log n/n)$ satisfies
\[
\|g(\cdot;\widehat\btheta)-f^*\|_2^2\le C\biggl\{\|f^*\|_\cS^2m^{-(d+3)/d}+(\sigma_\ve^2+\|f^*\|_\cS^2)\frac{md\log n}{n}\biggr\}
\]
with probability at least $1-O(n^{-C_2})$ for some constants $C_1,C_2,C>0$.
\end{theorem}

The proof technique used for Theorem \ref{thm:underparam} differs substantially from those in previous work, since we are analyzing a penalized rather than constrained problem and do not impose any boundedness constraints on the network function or parameters; cf.\ \citet{schmidt-hieber2020nonparametric}, \citet{farrell2021deep}, and \citet{parhi2022near}.

Finally, noting that the ranges of allowable $m$ in Theorems \ref{thm:pred_err} and \ref{thm:underparam} partially overlap, we put them together to obtain a complete picture of the generalization behavior of two-layer ReLU networks, as stated in the following encompassing result.

\begin{theorem}\label{thm:all_m}
Under Conditions \eqref{cond:f}--\eqref{cond:eps}, the regularized network estimator $g(\cdot;\widehat\btheta)$ with $\lambda=\min(\lambda_1,\lambda_2)$, where $\lambda_1$ and $\lambda_2$ are defined in Theorems \ref{thm:pred_err} and \ref{thm:underparam}, respectively, satisfies
\[
\|g(\cdot;\widehat\btheta)-f^*\|_2^2\le C\biggl\{\|f^*\|_\cS^2m^{-(d+3)/d}+(\sigma_\ve^2+\|f^*\|_\cS^2)\min\biggl(\sqrt{\frac{d\log n}{n}},\frac{md\log n}{n}\biggr)\biggr\}
\]
with probability at least $1-O(n^{-C_1})$ for some constants $C_1,C>0$ and sufficiently large $n$.
\end{theorem}

The implications of Theorem \ref{thm:all_m} have been discussed in the Introduction. In particular, it gives rise to the double descent risk curve illustrated in Figure \ref{fig:curve} and provides a simple yet appealing explanation for the curious phenomenon. In the underparametrized regime, the network estimator behaves as the usual nonparametric methods, with the network width $m$ controlling the trade-off between bias and variance. A too small or too large $m$ will result in an inferior performance, and a narrow valley around $m_0\asymp(n/(d\log n))^{d/(2d+3)}$ lies in between. As $m$ continues to increase and exceeds some threshold $m_1\asymp\sqrt{n/(d\log n)}$, the intrinsic model complexity and hence the variance of the network estimator become saturated and remain constant, while the bias diminishes consistently. This leads to a second, flat valley extending toward infinity.

Comparisons between the two valleys yield further insights. Asymptotically, the convergence rate of the first valley or underparametrized minimum risk, $O((d\log n/n)^{(d+3)/(2d+3)})$, is slightly smaller than that of the second valley or overparametrized minimum risk, $O(\sqrt{d\log n/n})$. In finite samples, however, this comparison can be reversed. A little algebra shows that the second valley is lower than the first whenever
\begin{equation}\label{eq:kappa}
\kappa\equiv\frac{\|f^*\|_\cS^2}{\sigma_\ve^2+\|f^*\|_\cS^2}>\biggl(\frac{1}{2}\biggr)^{(2d+3)/d}\biggl(\frac{n}{d\log n}\biggr)^{3/(2d)}.
\end{equation}
When $d\gg\log n$, the above condition approximately becomes $\kappa>1/4$, or the signal-to-noise ratio $\|f^*\|_\cS^2/\sigma_\ve^2=\kappa/(1-\kappa)>1/3$. An example with $\kappa=1$, $d=6$, and $n=1000$ was given in Figure \ref{fig:curve}. This makes intuitive sense since the reduction in approximation error plays a more important role when the signal is stronger. From the practitioner's perspective, the overparametrized regime is also more attractive in that it provides an infinitely wide sweet spot that avoids the choice of an optimal network width.

\subsection{Minimax lower bounds}\label{ssec:lower}
We have revealed that the risk curve of our estimator has two valleys. The convergence rate of the first valley is known to be minimax optimal over the function class $\cG_M$ \citep{parhi2022near}. In fact, the underparametrized result (Theorem \ref{thm:underparam}) relies crucially on the assumption that $M$ is finite; otherwise, the entropy calculations may be affected. In this subsection, we investigate the optimality of the second valley. Although it cannot be optimal over $\cG_M$, we will show that it is nearly minimax optimal over the larger function class $\cG$.

To gain intuition for the optimal rate, for any probability measure $\rho$ on $\S^{d-1}\times[-1,1]$, we consider the reproducing kernel Hilbert space (RKHS)
\[
\cH_\rho=\biggl\{\bx\mapsto\int_{\S^{d-1}\times[-1,1]}a(\bw)\sigma(\bv^T\bx+b)\,d\rho(\bw):\int_{\S^{d-1}\times[-1,1]}a(\bw)^2\,d\rho(\bw)<\infty\biggr\}
\]
associated with the kernel
\begin{equation}\label{eq:kernel}
H_\rho(\bx,\bz)=\E_{\bw\sim\rho}\bigl(\sigma(\bv^T\bx+b)\sigma(\bv^T\bz+b)\bigr).
\end{equation}
If the target function $f^*\in\cH_{\rho^*}$ for some known $\rho^*$, then the problem of recovering $f^*$ reduces to kernel ridge regression. It was shown by \citet{caponnetto2007optimal} that the minimax optimal rate for learning functions in an RKHS is $n^{-\gamma/(\gamma+1)}$ when the $j$th eigenvalue of the kernel decays at the rate of $j^{-\gamma}$ for $\gamma>1$. Noting that $\cH_\rho\subset\cG$ for all $\rho$ and letting $\gamma\to1$, we see that the desired minimax optimal rate should be $n^{-1/2}$. This heuristic argument is formalized in the following minimax result.

\begin{theorem}\label{thm:minimax}
Assume that $\bx_i\sim\mathrm{Uniform}(\B^d)$ and $\ve_i\sim N(0,1)$. Then there exists a constant $C>0$ such that
\[
\inf_{\widehat{f}}\sup_{f^*\in\cG}\E\|\widehat{f}-f^*\|_2^2\ge\frac{C}{\sqrt{n\log n}},
\]
where the infimum is taken over all estimators.
\end{theorem}

This result says that the upper bounds on the overparametrized minimum risk in Theorems \ref{thm:pred_err} and \ref{thm:all_m} are sharp up to logarithmic factors. Without requiring the existence of a finite $M$, these bounds are essentially unimprovable, which corroborates the effectiveness of overparametrized two-layer ReLU networks.

\section{Suboptimality of random feature models}\label{sec:rand_feat}
Random feature models \citep{rahimi2007random} provide a stochastic approximation to kernel methods by first mapping the input into a randomized feature space and then applying standard linear methods. Alternatively, they can be interpreted as two-layer neural networks with random first-layer weights and, as such, often serve as a prototype for studying the generalization behavior of realistic neural networks. For example, \citet{mei2022generalization} computed the prediction error of random feature regression that recovers the double descent curve in the asymptotic regime where $m,n,d\to\infty$ with $m\asymp n\asymp d$. We now show, however, that random feature models are not sufficient to explain the generalization power of fully trained two-layer networks by proving that they are suboptimal over our target function space.

Specifically, we consider the random feature model
\[
h_{\rho_0}(\bx;\ba)=\frac{1}{\sqrt{m}}\sum_{k=1}^ma_k\sigma(\bv_k^T\bx+b_k),
\]
where $\bw_k=(\bv_k^T,b_k)^T\sim\rho_0$ independently for some fixed $\rho_0$ and $\ba=(a_1,\dots,a_m)^T$ is the vector of parameters to be estimated. Minimizing the $\ell_2$-regularized empirical risk
\[
\frac{1}{2n}\sum_{k=1}^m\bigl(y_i-h_{\rho_0}(\bx_i;\ba)\bigr)^2+\frac{\lambda}{2}\|\ba\|_2^2
\]
gives the ridge estimator $\widehat\ba(\lambda)=(\bK+n\lambda\bI_n)^{-1}\by$, where $\bK=(K_{ij})\in\R^{n\times n}$ is the kernel matrix with entries
\[
K_{ij}=\frac{1}{m}\sum_{k=1}^m\sigma(\bv_k^T\bx_i+b_k)\sigma(\bv_k^T\bx_j+b_k).
\]
In fact, $K_{ij}\to H_{\rho_0}(\bx_i,\bx_j)$ as $m\to\infty$ for the kernel $H_\rho$ defined in \eqref{eq:kernel}. The following result establishes a lower bound on the worst-case performance of optimally tuned ridge estimators in random feature models.

\begin{proposition}\label{prop:rand_feat}
Under Conditions \eqref{cond:x} and \eqref{cond:eps}, there exists a universal constant $C>0$ such that
\[
\sup_{f^*\in\cG_M}\inf_{\lambda>0}\E\|h_{\rho_0}(\cdot;\widehat\ba(\lambda))-f^*\|_2^2\ge\frac{CM}{d\{\min(m,n)\}^{1/d}}.
\]
\end{proposition}

The proof of Proposition \ref{prop:rand_feat} builds on an approximation result of \citet[Theorem 6]{barron1993universal} for linear subspaces with fixed basis functions. Similar lower bounds have been obtained by \citet{e2020comparative} for random feature models trained by gradient descent with noiseless data. The exponential dependence of the rate on $d$ manifests the curse of dimensionality in random feature models for learning functions beyond an RKHS.

\section{Discussion}\label{sec:disc}
The ongoing debate over the double descent phenomenon and the virtues of overparametrization casts a cloud on the trustworthiness of modern deep learning methods and undermines the foundations of large machine learning models. We have developed a nonasymptotic generalization theory for finite-width two-layer neural networks without resorting to mean-field or neural tangent kernel approximations. As far as we are aware, this provides the first complete explanation for the double descent phenomenon beyond linear and kernel-type (e.g., random feature) methods. Compared with the existing literature, our theoretical framework has the following advantages: (i) we take a nonparametric viewpoint and consider target functions in a large function space, which allows us to define approximation and estimation errors in an appropriate manner and directly tackle the problem of bias--variance trade-off; (ii) unlike previous asymptotic studies, our nonasymptotic approach helps separate the effects of diverging dimensionality and overparametrization on generalization performance; (iii) the explicit regularization strategy we have adopted naturally extends classical and kernel ridge regression, making our results independent of the algorithmic specifics of nonconvex optimization.

Our theory yields insights that have not been previously obtained under simpler models or asymptotic regimes. We highlight some important ones as follows:

\paragraph*{Impact of dimensionality} In linear regression, the number of parameters coincides with the dimensionality, and hence it is impossible to decouple their effects. For kernel methods, \citet{liang2020multiple} and \citet{montanari2022interpolation} relaxed the proportional asymptotics on $n$ and $d$, but still required $d$ to be polynomially growing with $n$. Our results show that for two-layer neural networks the double descent curve persists even when $d$ is fixed and, therefore, the phenomenon is not tied to high dimensionality. Nevertheless, the dimensionality does play a role in determining the superiority of the overparametrized regime. Specifically, as seen from \eqref{eq:kappa}, a moderately large $d$ suffices to ensure the global optimality of infinite overparametrization over a wide range of signal-to-noise ratio.

\paragraph*{Double descent with optimal regularization} In linear and random feature models, it has been shown that optimal ridge regularization eliminates double descent \citep{hastie2022surprises,nakkiran2021optimal,mei2022generalization}. This raises the concern of whether double descent should be treated as a pathological behavior due to insufficient regularization and hence should be avoided or mitigated in practice. Contrary to this view, our theory, which has been derived under optimal choices of the regularization parameter, provides a radically different framework in which double descent is an intrinsic feature rather than an artifact and cannot be eliminated by optimal regularization.

\paragraph*{Complexity control} As pointed out by \citet{belkin2020reply}, the most interesting aspects of double descent is not the peaking phenomenon itself but its connection to classical ideas of the bias--variance trade-off and complexity control. Unfortunately, previous work offers little insight in this regard and does not clarify the mechanism behind the superiority of overparametrization. By contrast, our theory gives a clear explanation of what drives double descent: the partition of the parameter space into finitely many regions and the emergence of collinearity within each region reduce the effective dimensionality, thereby achieving adaptive complexity control in the overparametrized regime.

\paragraph*{Bias--variance trade-off} The literature presents a mixed picture of bias and variance in the overparametrized regime \citep{hastie2022surprises,mei2022generalization}: while the variance always decreases, the bias may increase (for well-specified linear models), decrease (for random feature models), or first decrease and then increase (for misspecified linear models). These somewhat peculiar behaviors are partly due to the fact that the ground truth in these settings is parametric and varying with the dimensionality. Neural networks, however, are intrinsically nonparametric, and the bias--variance trade-off should be discussed within this framework \citep{geman1992neural}. Embracing this viewpoint, our results show that the bias always decreases, while the variance remains constant after the saturation threshold. Although there is no more trade-off in the overparametrized regime, the general principle of bias--variance trade-off in the sense of \citet{derumigny2023lower} still seems to hold: the variance is lower bounded if the bias is small.

Our framework may be extended in several directions. The most important would be the development for deep neural networks, by following the idea of finding a convex reformulation and analyzing the symmetric structures arising from overparametrization. Such a formulation does not seem to be readily available, but see \citet{ergen2021revealing} for useful results in some special cases. For simplicity, we have considered only explicit regularization and the theoretical optimal solution to the regularized problem. An interesting direction is to take into account practical algorithms and implicit regularization, possibly by exploring the connection of our problem to $\ell_2$ regularization. Finally, it would be worthwhile to extend our theory to classification problems and more network architectures such as convolutional and recurrent neural networks.

\begin{appendix}
\section{Proofs for Section \ref{sec:dual}}
In this appendix we provide the proofs of Propositions \ref{prop:ridge}--\ref{prop:glasso}. To simplify the notation, we write $\widetilde\bx_i=(\bx_i^T,1)^T$.

\begin{proof}[Proof of Proposition \ref{prop:ridge}]
Let $\widehat\btheta_{\ell_2}$ be an arbitrary solution to problem \eqref{eq:ridge} and $\widehat\btheta$ an arbitrary solution to problem \eqref{eq:opt}. By the optimality of $\widehat\btheta_{\ell_2}$ and $\widehat\btheta$, we have
\begin{equation}\label{eq:opt_cond}
J_n^{\ell_2}(\widehat\btheta_{\ell_2};\lambda)\le J_n^{\ell_2}(\cT_1(\widehat\btheta);\lambda),\qquad J_n(\widehat\btheta;\lambda)\le J_n(\widehat\btheta_{\ell_2};\lambda),
\end{equation}
where $J_n^{\ell_2}(\btheta;\lambda)$ is the objective function of \eqref{eq:ridge}. By the definition of $\cT_1$, $J_n^{\ell_2}(\cT_1(\btheta);\lambda)=J_n(\btheta;\lambda)$ for any $\btheta\in\bTheta_m$. Moreover, $\widehat\btheta_{\ell_2}=\cT_1(\widehat\btheta_{\ell_2})$. Combining these facts with \eqref{eq:opt_cond} gives
\[
J_n^{\ell_2}(\widehat\btheta_{\ell_2};\lambda)\le J_n^{\ell_2}(\cT_1(\widehat\btheta);\lambda)=J_n(\widehat\btheta;\lambda)\le J_n(\widehat\btheta_{\ell_2};\lambda)=J_n^{\ell_2}(\widehat\btheta_{\ell_2};\lambda),
\]
which means that $\widehat\btheta_{\ell_2}$ is a minimizer of $J_n(\btheta;\lambda)$ and that $\cT_1(\widehat\btheta)$ a minimizer of $J_n^{\ell_2}(\btheta;\lambda)$, completing the proof.
\end{proof}

\begin{proof}[Proof of Proposition \ref{prop:grad_flow}]
By direct differentiation, the gradient flow $d\btheta(t)/dt=-\nabla_{\btheta}J_n(\btheta(t);\lambda)$ for problem \eqref{eq:opt} can be written as
\begin{align}
\frac{d}{dt}a_j(t)&=\frac{1}{n}\sum_{i=1}^n\bigl(y_i-g(\bx_i;\btheta(t))\bigr)^2\sigma(\widetilde\bx_i^T\bw_j(t))-\lambda\|\bw_j(t)\|_2\partial|a_j(t)|,\label{eq:grad_a}\\
\frac{d}{dt}\bw_j(t)&=\frac{1}{n}\sum_{i=1}^n\bigl(y_i-g(\bx_i;\btheta(t))\bigr)^2a_j(t)\partial\sigma(\widetilde\bx_i^T\bw_j(t))\widetilde\bx_i-\lambda|a_j(t)|\partial\|w_j(t)\|_2,\label{eq:grad_w}
\end{align}
for $j=1,\dots,m$, where $\partial$ denotes the subgradient. Using the identities $a\partial|a|=|a|$, $\bw^T\partial\|\bw\|_2=\|\bw\|_2^2$, and $z\partial\sigma(z)=\sigma(z)$, left multiplying \eqref{eq:grad_a} by $a_j(t)$ and \eqref{eq:grad_w} by $\bw_j(t)^T$ gives
\[
\frac{d}{dt}|a_j(t)|^2=\frac{d}{dt}\|\bw_j(t)\|_2^2,\quad j=1,\dots,m.
\]
If the initialization is reparametrized by $\mathcal{T}_1$, that is, $|a_j(0)|^2=\|\bw_j(0)\|_2^2$ for all $j$, then we have, for all $t\ge0$,
\begin{equation}\label{eq:a=w}
|a_j(t)|^2=\|\bw_j(t)\|_2^2,\quad j=1,\dots,m.
\end{equation}

Similarly, the gradient flow for problem \eqref{eq:ridge} can be written as
\begin{align}
\frac{d}{d t}a_j^{\ell_2}(t)&=\frac{1}{n}\sum_{i=1}^n\bigl(y_i-g(\bx_i;\btheta_{\ell_2}(t))\bigr)^2\sigma(\widetilde\bx_i^T\bw_j^{\ell_2}(t))-\lambda a_j^{\ell_2}(t),\label{eq:grad2_a}\\
\frac{d}{d t}\bw_j^{\ell_2}(t)&=\frac{1}{n}\sum_{i=1}^n\bigl(y_i-g(\bx_i;\btheta_{\ell_2}(t))\bigr)^2a_j^{\ell_2}(t)\partial\sigma(\widetilde\bx_i^T\bw_j^{\ell_2}(t))\widetilde\bx_i-\lambda\bw_j^{\ell_2}(t),\label{eq:grad2_w}
\end{align}
for $j=1,\dots,m$. Using \eqref{eq:a=w} we have $\|\bw_j(t)\|_2\partial|a_j(t)|=|a_j(t)|\partial|a_j(t)|=a_j(t)$ and $|a_j(t)|\partial\|\bw_j(t)\|_2=\|\bw_j(t)\|_2\partial\|\bw_j(t)\|_2=\bw_j(t)$, in which case the gradient flows \eqref{eq:grad_a}--\eqref{eq:grad_w} and \eqref{eq:grad2_a}--\eqref{eq:grad2_w} are identical. Initialized at the same point, their trajectories must coincide.
\end{proof}

To prove Propositions \ref{prop:sym} and \ref{prop:glasso}, we first introduce the following lemma, which says that the ReLU function is linear over each cone $Q_j$.
\begin{lemma}\label{lem:linear_relu}
If $(a_k,\bw_k^T)^T$ and $(a_\ell,\bw_\ell^T)^T$ lie in the interior of the same cone $Q_j$, then
\[
a_k\sigma(\bw_k^T\widetilde\bx_i)+a_\ell\sigma(\bw_\ell^T\widetilde\bx_i)=s_j\sigma(|a_k|\bw_k^T\widetilde\bx_i+|a_\ell|\bw_\ell^T\widetilde\bx_i)
\]
for $i=1,\dots,n$.
\end{lemma}

\begin{proof}
By definition, all points $(a,\bw^T)^T$ in the interior of $Q_j$ satisfy $\sgn(a)=s_j$ and $I(\bw^T\widetilde\bx_i\ge0)=(\bD_j)_{ii}$, where $(\bD_j)_{ii}$ is the $i$th diagonal entry of $\bD_j$. Then
\begin{align*}
&a_k\sigma(\bw_k^T\widetilde\bx_i)+a_\ell\sigma(\bw_\ell^T\widetilde\bx_i)\\
&\quad=a_k\bw_k^T\widetilde\bx_iI(\bw_k^T\widetilde\bx_i\ge0)+a_\ell\bw_\ell^T\widetilde\bx_iI(\bw_\ell^T\widetilde\bx_i\ge0)\\
&\quad=s_j(\bD_j)_{ii}(|a_k|\bw_k^T\widetilde\bx_i+|a_\ell|\bw_\ell^T\widetilde\bx_i)=s_j\sigma(|a_k|\bw_k^T\widetilde\bx_i+|a_\ell|\bw_\ell^T\widetilde\bx_i).\mbox{\qedhere}
\end{align*}
\end{proof}

\begin{proof}[Proof of Proposition \ref{prop:sym}]
Suppose that $(\widehat{a}_k,\widehat\bw_k^T)^T$ and $(\widehat{a}_\ell,\widehat\bw_\ell^T)^T$ lie in the interior of $Q_j$ but are not collinear. Define the new parameter $\widetilde\btheta=(\widetilde{a}_1,\dots,\widetilde{a}_m,\widetilde\bw_1^T,\dots,\widetilde\bw_m^T)^T$ with
\[
\widetilde{a}_k=\widetilde{a}_\ell=s_j,\qquad\widetilde\bw_k=\widetilde\bw_\ell=\frac{1}{2}(|a_k|\bw_k+|a_\ell|\bw_\ell),
\]
while keeping the other components unchanged. Then by Lemma \ref{lem:linear_relu} we have
\[
\widehat{a}_k\sigma(\widehat\bw_k^T\widetilde\bx_i)+\widehat{a}_\ell\sigma(\widehat\bw_\ell^T\widetilde\bx_i) =\widetilde{a}_k\sigma(\widetilde\bx_i^T\widetilde\bw_\ell)+\widetilde{a}_\ell\sigma(\widetilde\bx_i^T\widetilde\bw_\ell),
\]
and by the triangle inequality,
\[
|\widetilde{a}_k|\|\widetilde\bw_k\|_2+|\widetilde{a}_\ell|\|\widetilde\bw_\ell\|_2=\bigl\||\widehat{a}_k|\widehat\bw_k+|\widehat{a}_\ell|\widehat\bw_\ell\bigr\|_2 <|\widehat{a}_k|\|\widehat\bw_k\|_2+|\widehat{a}_\ell|\|\widehat\bw_\ell\|_2.
\]
This entails that $J_n(\widetilde\btheta;\lambda)<J_n(\widehat\btheta;\lambda)$, which contradicts the optimality of $\widehat\btheta$.
\end{proof}

\begin{proof}[Proof of Proposition \ref{prop:glasso}]
By Lemma \ref{lem:linear_relu} and the definition of $\widetilde\btheta=\cT_2(\btheta)$ in \eqref{eq:collin}, we have
\begin{align*}
g(\bx_i;\btheta)&=\sum_{j=1}^{2p}\sum_{k\in S_j}a_k\sigma(\bw_k^T\widetilde\bx_i)=\sum_{j=1}^{2p}s_j\sigma\biggl(\sum_{k\in S_j}|a_k|\bw_k^T\widetilde\bx_i\biggr)\\
&=\sum_{j=1}^{2p}\sum_{k\in S_j}\widetilde{a}_k\sigma(\widetilde\bw_k^T\widetilde\bx_i)=g(\bx_i;\widetilde\btheta)
\end{align*}
and
\begin{align*}
\|\bB(\widetilde\btheta)\|_{2,1}&=\sum_{j=1}^{2p}\|\bbeta_j(\widetilde\btheta)\|_2=\sum_{j=1}^{2p}\biggl\|\sum_{k\in S_j}|\widetilde{a}_k|\widetilde\bw_k\biggr\|_2=\sum_{k=1}^m|\widetilde{a}_k|\|\widetilde\bw_k\|_2 =\nu(\widetilde\btheta)\\
&=\sum_{j=1}^{2p}\biggl\|\sum_{k\in S_j}|a_k|\bw_k\biggr\|_2\le\sum_{k=1}^m|a_k|\|\bw_k\|_2=\nu(\btheta).
\end{align*}

To prove the second assertion, from \eqref{eq:linear_form} we have, for any $\btheta\in\bTheta_m$,
\begin{equation}\label{eq:J_loss}
\frac{1}{2n}\biggl\|\by-\sum_{k=1}^ma_k\sigma(\bX\bw_k)\biggr\|_2^2=\frac{1}{2n}\biggl\|\by-\sum_{j=1}^{2p}\bD_j\bX\bbeta_j(\btheta)\biggr\|_2^2.
\end{equation}
Also, by the collinearity property of $\widehat\btheta$ from Proposition \ref{prop:sym},
\begin{equation}\label{eq:J_pen}
\|\bB(\widehat\btheta)\|_{2,1}=\sum_{j=1}^{2p}\biggl\|\sum_{k\in S_j}|\widehat{a}_k|\widehat\bw_k\biggr\|_2=\sum_{k=1}^m|\widehat{a}_k|\|\widehat\bw_k\|_2=\nu(\widehat\btheta).
\end{equation}
Combining \eqref{eq:J_loss} and \eqref{eq:J_pen} yields the expression for $J_n(\widehat\btheta;\lambda)$.
\end{proof}

\section{Proofs of results in the overparametrized regime}\label{sec:proof_over}
In this appendix we provide the proofs of high-probability bounds \eqref{eq:emp_err1} and \eqref{eq:pred_err1} for the regularized network estimator under the overparametrized regime. The proofs of risk bounds \eqref{eq:emp_err2} and \eqref{eq:pred_err2} in expectation are deferred to Section \ref{sec:exp_bound} of the Supplementary Material.

We first introduce some notation. Define the class of two-layer ReLU networks with bounded scaled variation norm by
$\cF(m,F)=\{g(\cdot;\btheta):\nu(\btheta)\le F\}$. For any $f^*\in\cG_M$, let $g(\cdot;\btheta^*)$ denote the $L_\infty(\B^d)$-approximation of $f^*$ in Theorem \ref{thm:approx}, where $\btheta_m^*=(a_1^*,\dots,a_m^*,\linebreak[0]\bw_1^{*T},\dots,\bw_m^{*T})^T$.

\begin{proof}[Proof of \eqref{eq:emp_err1} in Theorem \ref{thm:emp_err}]
By the optimality of $\widehat\btheta$, we have
\[
\frac{1}{2n}\sum_{i=1}^n\bigl(g(\bx_i;\widehat\btheta)-y_i\bigr)^2+\lambda\nu(\widehat\btheta)\le\frac{1}{2n}\sum_{i=1}^n\bigl(g(\bx_i;\btheta^*)-y_i\bigr)^2+\lambda\nu(\btheta^*).
\]
Rearranging terms gives
\begin{equation}\label{eq:small_m_basic}
\begin{split}
&\frac{1}{2}\|g(\cdot;\widehat\btheta)-f^*(\cdot)\|_n^2\\
&\quad\le\lambda\bigl(\nu(\btheta^*)-\nu(\widehat{\btheta})\bigr)+\frac{1}{2}\|g(\cdot;\btheta^*)-f^*(\cdot)\|_n^2+\frac{1}{n}\biggl|\sum_{i=1}^n\ve_i\bigl(g(\bx_i;\widehat\btheta)-g(\bx_i;\btheta^*)\bigr)\biggr|\\
&\quad\equiv T_1+T_2+T_3.
\end{split}
\end{equation}
For brevity, we write $\bB^*=\bB(\btheta^*)=(\bbeta_1^*,\dots,\bbeta_{2p}^*)$ and $\widehat\bB=\bB(\widehat\btheta)=(\widehat\bbeta_1,\dots,\widehat\bbeta_{2p})$. Since we need only evaluate $g(\cdot;\btheta^*)$ on the training sample, by Proposition \ref{prop:glasso} we can assume without loss of generality that $\nu(\btheta^*)=\|\bB^*\|_{2,1}$. It also follows from Proposition \ref{prop:glasso} that $\nu(\widehat\btheta)=\|\widehat\bB\|_{2,1}$. These facts, together with the triangle inequality, imply that
\begin{equation}\label{eq:large_diff_norm}
T_1=\lambda(\|\bB^*\|_{2,1}-\|\widehat\bB\|_{2,1})\le2\lambda\|\bB^*\|_{2,1}-\lambda\|\widehat\bB-\bB^*\|_{2,1}.
\end{equation}
To bound $T_2$, applying Theorem \ref{thm:approx} yields
\begin{equation}\label{eq:large_approx}
T_2=\frac{1}{2n}\sum_{i=1}^n\bigl(g(\bx_i;\btheta^*)-f^*(\bx_i)\bigr)^2\le C_1\|f^*\|_\cS^2 m^{-(d+3)/d}
\end{equation}
for some constant $C_1>0$. Also, by H\"{o}lder's inequality,
\begin{equation}\label{eq:large_noise}
T_3=\frac{1}{n}\biggl|\bve^T\sum_{j=1}^{2p}\bD_j\bX(\widehat\bbeta_j-\bbeta_j^*)\biggr|\le\frac{1}{\sqrt{n}}\max_{1\le j\le2p}\|\bv_j\|_2\sum_{j=1}^{2p}\|\widehat\bbeta_j-\bbeta_j^*\|_2,
\end{equation}
where $\bv_j=\bX^T\bD_j^T\bve/\sqrt{n}$ and $\bve=(\ve_1,\dots,\ve_n)^T$. Combining \eqref{eq:large_diff_norm}--\eqref{eq:large_noise}, choosing $\lambda\ge2n^{-1/2}\max_{j}\|\bv_j\|_2$, and noting that $\nu(\btheta^*)\le6\|f^*\|_\cS$ by Theorem \ref{thm:approx}, we obtain
\begin{equation}\label{eq:large_overall}
\begin{split}
&\|g(\cdot;\widehat\btheta)-f^*(\cdot)\|_n^2\\
&\quad\le2C_1\|f^*\|_\cS^2m^{-(d+3)/d}+4\lambda\nu(\btheta^*)+2\biggl(\frac{1}{\sqrt{n}}\max_{1\le j=1\le2p}\|\bv_j\|_2-\lambda\biggr)\|\widehat{\bB}-\bB^*\|_{2,1}\\
&\quad\le2C_1\|f^*\|_\cS^2m^{-(d+3)/d}+24\lambda \|f^*\|_\cS.
\end{split}
\end{equation}

It remains to bound $n^{-1/2}\max_{j}\|\bv_j\|_2$. Let $\bH_j=\bD_j\bX\bX^T\bD_j/n$, so that $\bv_j^T\bv_j=\bve^T\bH_j\bve$. By the definition of $\bD_j$ and the assumption that $\max_i\|\bx_i\|_2\le 1$, we have
\[
\|\bH_j\|_2\le\tr(\bH_j)=n^{-1}\tr(\bX^{T}\bX)\le2.
\]
Applying a tail bound for quadratic forms of sub-Gaussian vectors \citep{hsu2012tail} gives
\[
P(\|\bv_j\|_2^2\ge2\sigma_\ve^2+4\sigma_\ve^2\sqrt{t}+4\sigma_\ve^2t)\le e^{-t}.
\]
Hence, by the union bound, $P(\max_j\|\bv_j\|_2^2\ge2\sigma_\ve^2+4\sigma_\ve^2\sqrt{t}+4\sigma_\ve^2t)\le2pe^{-t}$. Recall from \eqref{eq:p_bound} that $p\le2n^d$. Choosing $t=(4+d)\log n>1$ for $n\ge2$ yields
\begin{equation}\label{eq:concentration_xi}
\max_{1\le j\le2p}\|\bv_j\|_2^2<2\sigma_\ve^2+4\sigma_\ve^2\sqrt{t}+4\sigma_\ve^2t<10\sigma_\ve^2t<16\sigma_\ve^2(4+d)\log n
\end{equation}
with probability at least $1-4n^{-4}$. Thus, for $\lambda\ge2n^{-1/2}\max_j\|\bv_j\|_2$ to hold with the same probability, it suffices to set $\lambda=8\sigma_\ve\sqrt{(4+d)\log n/n}$. To complete the proof, substituting the value of $\lambda$ into \eqref{eq:large_overall} gives
\[
\|g(\cdot;\widehat\btheta)-f^*(\cdot)\|_n^2\le2C_1\|f^*\|_\cS^2m^{-(d+3)/d}+96(\sigma_\ve^2+\|f^*\|_\cS^2)\sqrt{\frac{(4+d)\log n}{n}},
\]
where we have used the inequality $2\sigma_\ve\|f^*\|_\cS\le\sigma_\ve^2+\|f^*\|_\cS^2$.
\end{proof}

To prove \eqref{eq:pred_err1} in Theorem \ref{thm:pred_err}, we need the following maximal inequality, whose proof can be found in Section \ref{sec:upper_lemma_proof} of the Supplementary Material.

\begin{lemma}\label{lem:talagrand}
Assume that Condition \eqref{cond:x} holds. Let $\cF^*(m,1)=\{f-f^*\colon f\in\cF(m,1),\linebreak[0]f^*\text{ is fixed with }\|f^*\|_\cS\le1\}$ and
$Z_n=\sup_{f\in\cF^*(m,1)}\bigl|\|f\|_n^2-\|f\|_2^2\bigr|$. Then $\E Z_n\le C_\cF n^{-1/2}$ for some constant $C_\cF>0$ depending only on $d$. Furthermore, if $n\ge C_\cF^2$, then
\begin{equation}\label{eq:talagrand}
P\biggl(Z_n\ge\frac{C_\cF}{\sqrt{n}}+t\biggr)\le\exp\biggl\{-\frac{n}{32}\min\biggl(\frac{t^2}{12e},t\biggr)\biggr\}.
\end{equation}
\end{lemma}

\begin{proof}[Proof of \eqref{eq:pred_err1} in Theorem \ref{thm:pred_err}]
Let $\widehat{f}(\cdot)=g(\cdot;\widehat\btheta)$ and $\widehat\Delta=\widehat{f}-f^*$. By the proof of \eqref{eq:emp_err1} in Theorem \ref{thm:emp_err} and, in particular, \eqref{eq:large_overall}, if we choose $\lambda=8\sigma_\ve\sqrt{(4+d)\log n/n}$, then, with probability at least $1-4n^{-4}$,
\[
0\le\|\widehat{f}-f^*\|_n^2\le2C_1\|f^*\|_\cS^2m^{-(d+3)/d}+4\lambda\nu(\btheta^*)-\lambda(\|\widehat\bB\|_{2,1}-\|\bB^*\|_{2,1})
\]
for some constant $C_1>0$. Since $\nu(\btheta^*)=\|\bB^*\|_{2,1}\le6\|f^*\|_\cS$ and $\|\widehat{\bB}\|_{2,1}=\nu(\widehat{\btheta})$, we further obtain
\[
\lambda \nu(\widehat{\btheta})\le5\lambda\nu(\btheta^*)+2C_1\|f^*\|_\cS^2m^{-(d+3)/d}.
\]
If $m$ satisfies
\begin{equation}\label{eq:critic_m}
m\ge\biggl(\frac{C_1\|f^*\|_\cS}{8\sigma_\ve}\biggr)^{d/(d+3)}\biggl(\frac{n}{d\log n}\biggr)^{d/(2(d+3))},
\end{equation}
then
\begin{equation}\label{eq:large_norm_bound}
\nu(\widehat{\btheta})\le5\nu(\btheta^*)+2\|f^*\|_\cS\le32\|f^*\|_\cS.
\end{equation}
By \eqref{eq:svnorm} and the homogeneity of ReLU, the scaled variation norm of $\widehat{f}/\nu(\widehat{\btheta})$ is exactly 1. Also, by definition, the $\cS$-norm of $f^*/(32\|f^*\|_\cS)$ is smaller than 1. Thus, the event
\begin{equation}\label{eq:Delta_in_class}
\frac{\widehat{\Delta}}{32\|f^*\|_\cS}=\frac{\widehat{f}}{32\|f^*\|_\cS}-\frac{f^*}{32\|f^*\|_\cS}\in\cF^*(m,1)
\end{equation}
holds with probability at least $1-4n^{-4}$.

Now, conditioning on the event $\{\widehat\Delta/(32\|f^*\|_\cS)\in\cF^*(m,1)\}$, applying Lemma \ref{lem:talagrand} with $t=8\sqrt{6e\log n/n}<12e$ yields
\[
\|\widehat\Delta\|_2^2\le\|\widehat\Delta\|_n^2+\frac{1024}{\sqrt{n}}C_\cF\|f^*\|_\cS^2+8192\|f^*\|_\cS^2\sqrt{\frac{6e\log n}{n}}
\]
with probability at least $1-n^{-1}$. By Theorem \ref{thm:emp_err}, with probability at least $1-4n^{-4}$ we have
\[
\|\widehat\Delta\|_n^2=\|\widehat{f}-f^*\|_n^2\le C_3\biggl\{\|f^*\|_\cS^2m^{-(d+3)/d}+(\sigma_\ve^2+\|f^*\|_\cS^2)\sqrt{\frac{d\log n}{n}}\biggr\}
\]
for some constant $C_3>0$. Combining these pieces, we conclude that
\[
\|\widehat{f}-f^*\|_2^2\le C_4\biggl\{\|f^*\|_\cS^2m^{-(d+3)/d}+(\sigma_\ve^2+\|f^*\|_\cS^2)\sqrt{\frac{d\log n}{n}}\biggr\}
\]
with probability at least $1-O(n^{-1})$ for some constant $C_4>0$.
\end{proof}
\end{appendix}

\begin{funding}
This research was supported by Beijing Natural Science Foundation grant Z190001 and National Natural Science Foundation of China grants 12171012 and 12292981.
\end{funding}

\begin{supplement}
\stitle{Supplement to ``Nonasymptotic theory for two-layer neural networks: Beyond the bias--variance trade-off''}
\sdatatype{.pdf}
\sdescription{The Supplementary Material contains the remaining proofs and technical details.}
\end{supplement}

\bibliographystyle{imsart-nameyear}
\bibliography{tradeoff}

\begin{thebibliography}{76}

\bibitem[\protect\citeauthoryear{Arora, Cohen and
  Hazan}{2018}]{arora2018optimization}
\begin{binproceedings}[author]
\bauthor{\bsnm{Arora},~\bfnm{Sanjeev}\binits{S.}},
  \bauthor{\bsnm{Cohen},~\bfnm{Nadav}\binits{N.}} \AND
  \bauthor{\bsnm{Hazan},~\bfnm{Elad}\binits{E.}}
(\byear{2018}).
\btitle{On the optimization of deep networks: Implicit acceleration by
  overparameterization}.
In \bbooktitle{Proceedings of the 35th International Conference on Machine
  Learning}
\bpages{244--253}.
\end{binproceedings}
\endbibitem

\bibitem[\protect\citeauthoryear{Arora et~al.}{2019}]{arora2019fine}
\begin{binproceedings}[author]
\bauthor{\bsnm{Arora},~\bfnm{Sanjeev}\binits{S.}},
  \bauthor{\bsnm{Du},~\bfnm{Simon~S.}\binits{S.~S.}},
  \bauthor{\bsnm{Hu},~\bfnm{Wei}\binits{W.}},
  \bauthor{\bsnm{Li},~\bfnm{Zhiyuan}\binits{Z.}} \AND
  \bauthor{\bsnm{Wang},~\bfnm{Ruosong}\binits{R.}}
(\byear{2019}).
\btitle{Fine-grained analysis of optimization and generalization for
  overparameterized two-layer neural networks}.
In \bbooktitle{Proceedings of the 36th International Conference on Machine
  Learning}
\bpages{322--332}.
\end{binproceedings}
\endbibitem

\bibitem[\protect\citeauthoryear{Bach}{2017}]{bach2017breaking}
\begin{barticle}[author]
\bauthor{\bsnm{Bach},~\bfnm{Francis}\binits{F.}}
(\byear{2017}).
\btitle{Breaking the curse of dimensionality with convex neural networks}.
\bjournal{J. Mach. Learn. Res.}
\bvolume{18}
\bpages{1--53}.
\end{barticle}
\endbibitem

\bibitem[\protect\citeauthoryear{Barron}{1993}]{barron1993universal}
\begin{barticle}[author]
\bauthor{\bsnm{Barron},~\bfnm{Andrew~R.}\binits{A.~R.}}
(\byear{1993}).
\btitle{Universal approximation bounds for superpositions of a sigmoidal
  function}.
\bjournal{IEEE Trans. Inf. Theory}
\bvolume{39}
\bpages{930--945}.
\end{barticle}
\endbibitem

\bibitem[\protect\citeauthoryear{Barron}{1994}]{barron1994approximation}
\begin{barticle}[author]
\bauthor{\bsnm{Barron},~\bfnm{Andrew~R.}\binits{A.~R.}}
(\byear{1994}).
\btitle{Approximation and estimation bounds for artificial neural networks}.
\bjournal{Mach. Learn.}
\bvolume{14}
\bpages{115--133}.
\end{barticle}
\endbibitem

\bibitem[\protect\citeauthoryear{Bartlett, Mendelson and
  Neeman}{2012}]{bartlett2012regularized}
\begin{barticle}[author]
\bauthor{\bsnm{Bartlett},~\bfnm{Peter~L.}\binits{P.~L.}},
  \bauthor{\bsnm{Mendelson},~\bfnm{Shahar}\binits{S.}} \AND
  \bauthor{\bsnm{Neeman},~\bfnm{Joseph}\binits{J.}}
(\byear{2012}).
\btitle{$\ell_1$-regularized linear regression: Persistence and oracle
  inequalities}.
\bjournal{Probab. Theory Related Fields}
\bvolume{154}
\bpages{193--224}.
\end{barticle}
\endbibitem

\bibitem[\protect\citeauthoryear{Bartlett et~al.}{2020}]{bartlett2020benign}
\begin{barticle}[author]
\bauthor{\bsnm{Bartlett},~\bfnm{P.~L.}\binits{P.~L.}},
  \bauthor{\bsnm{Long},~\bfnm{P.~M.}\binits{P.~M.}},
  \bauthor{\bsnm{Lugosi},~\bfnm{G.}\binits{G.}} \AND
  \bauthor{\bsnm{Tsigler},~\bfnm{A.}\binits{A.}}
(\byear{2020}).
\btitle{Benign overfitting in linear regression}.
\bjournal{Proc. Natl. Acad. Sci. USA}
\bvolume{117}
\bpages{30063--30070}.
\end{barticle}
\endbibitem

\bibitem[\protect\citeauthoryear{Belkin, Hsu and Xu}{2020}]{belkin2020two}
\begin{barticle}[author]
\bauthor{\bsnm{Belkin},~\bfnm{Mikhail}\binits{M.}},
  \bauthor{\bsnm{Hsu},~\bfnm{Daniel}\binits{D.}} \AND
  \bauthor{\bsnm{Xu},~\bfnm{Ji}\binits{J.}}
(\byear{2020}).
\btitle{Two models of double descent for weak features}.
\bjournal{SIAM J. Math. Data Sci.}
\bvolume{2}
\bpages{1167--1180}.
\end{barticle}
\endbibitem

\bibitem[\protect\citeauthoryear{Belkin et~al.}{2019}]{belkin2019reconciling}
\begin{barticle}[author]
\bauthor{\bsnm{Belkin},~\bfnm{Mikhail}\binits{M.}},
  \bauthor{\bsnm{Hsu},~\bfnm{Daniel}\binits{D.}},
  \bauthor{\bsnm{Ma},~\bfnm{Siyuan}\binits{S.}} \AND
  \bauthor{\bsnm{Mandal},~\bfnm{Soumik}\binits{S.}}
(\byear{2019}).
\btitle{Reconciling modern machine-learning practice and the classical
  bias--variance trade-off}.
\bjournal{Proc. Natl. Acad. Sci. USA}
\bvolume{116}
\bpages{15849--15854}.
\end{barticle}
\endbibitem

\bibitem[\protect\citeauthoryear{Belkin et~al.}{2020}]{belkin2020reply}
\begin{barticle}[author]
\bauthor{\bsnm{Belkin},~\bfnm{Mikhail}\binits{M.}},
  \bauthor{\bsnm{Hsu},~\bfnm{Daniel}\binits{D.}},
  \bauthor{\bsnm{Ma},~\bfnm{Siyuan}\binits{S.}} \AND
  \bauthor{\bsnm{Mandal},~\bfnm{Soumik}\binits{S.}}
(\byear{2020}).
\btitle{Reply to loog et al.: Looking beyond the peaking phenomenon}.
\bjournal{Proc. Natl. Acad. Sci. USA}
\bvolume{117}
\bpages{10627}.
\end{barticle}
\endbibitem

\bibitem[\protect\citeauthoryear{Bishop}{1995}]{bishop1995training}
\begin{barticle}[author]
\bauthor{\bsnm{Bishop},~\bfnm{Chris~M.}\binits{C.~M.}}
(\byear{1995}).
\btitle{Training with noise is equivalent to {T}ikhonov regularization}.
\bjournal{Neural Comput.}
\bvolume{7}
\bpages{108--116}.
\end{barticle}
\endbibitem

\bibitem[\protect\citeauthoryear{Brown et~al.}{2020}]{brown2020language}
\begin{binproceedings}[author]
\bauthor{\bsnm{Brown},~\bfnm{T.~B.}\binits{T.~B.}},
  \bauthor{\bsnm{Mann},~\bfnm{B.}\binits{B.}},
  \bauthor{\bsnm{Ryder},~\bfnm{N.}\binits{N.}},
  \bauthor{\bsnm{Subbiah},~\bfnm{M.}\binits{M.}},
  \bauthor{\bsnm{Kaplan},~\bfnm{J.}\binits{J.}},
  \bauthor{\bsnm{Dhariwal},~\bfnm{P.}\binits{P.}},
  \bauthor{\bsnm{Neelakantan},~\bfnm{A.}\binits{A.}},
  \bauthor{\bsnm{Shyam},~\bfnm{P.}\binits{P.}},
  \bauthor{\bsnm{Sastry},~\bfnm{G.}\binits{G.}},
  \bauthor{\bsnm{Askell},~\bfnm{A.}\binits{A.}},
  \bauthor{\bsnm{Agarwal},~\bfnm{S.}\binits{S.}},
  \bauthor{\bsnm{Herbert-Voss},~\bfnm{A.}\binits{A.}},
  \bauthor{\bsnm{Krueger},~\bfnm{G.}\binits{G.}},
  \bauthor{\bsnm{Henighan},~\bfnm{T.}\binits{T.}},
  \bauthor{\bsnm{Child},~\bfnm{R.}\binits{R.}},
  \bauthor{\bsnm{Ramesh},~\bfnm{A.}\binits{A.}},
  \bauthor{\bsnm{Ziegler},~\bfnm{D.~M.}\binits{D.~M.}},
  \bauthor{\bsnm{Wu},~\bfnm{J.}\binits{J.}},
  \bauthor{\bsnm{Winter},~\bfnm{C.}\binits{C.}},
  \bauthor{\bsnm{Hesse},~\bfnm{C.}\binits{C.}},
  \bauthor{\bsnm{Chen},~\bfnm{M.}\binits{M.}},
  \bauthor{\bsnm{Sigler},~\bfnm{E.}\binits{E.}},
  \bauthor{\bsnm{Litwin},~\bfnm{M.}\binits{M.}},
  \bauthor{\bsnm{Gray},~\bfnm{S.}\binits{S.}},
  \bauthor{\bsnm{Chess},~\bfnm{B.}\binits{B.}},
  \bauthor{\bsnm{Clark},~\bfnm{J.}\binits{J.}},
  \bauthor{\bsnm{Berner},~\bfnm{C.}\binits{C.}},
  \bauthor{\bsnm{McCandlish},~\bfnm{S.}\binits{S.}},
  \bauthor{\bsnm{Radford},~\bfnm{A.}\binits{A.}},
  \bauthor{\bsnm{Sutskever},~\bfnm{I.}\binits{I.}} \AND
  \bauthor{\bsnm{Amodei},~\bfnm{D.}\binits{D.}}
(\byear{2020}).
\btitle{Language models are few-shot learners}.
In \bbooktitle{Advances in Neural Information Processing Systems}
\bvolume{33}
\bpages{1877--1901}.
\end{binproceedings}
\endbibitem

\bibitem[\protect\citeauthoryear{B{\"u}hlmann and van~de
  Geer}{2011}]{buhlmann2011statistics}
\begin{bbook}[author]
\bauthor{\bsnm{B{\"u}hlmann},~\bfnm{Peter}\binits{P.}} \AND
  \bauthor{\bparticle{van~de} \bsnm{Geer},~\bfnm{Sara}\binits{S.}}
(\byear{2011}).
\btitle{Statistics for High-Dimensional Data: Methods, Theory and
  Applications}.
\bpublisher{Springer}, \baddress{Berlin}.
\end{bbook}
\endbibitem

\bibitem[\protect\citeauthoryear{Caponnetto and
  De~Vito}{2007}]{caponnetto2007optimal}
\begin{barticle}[author]
\bauthor{\bsnm{Caponnetto},~\bfnm{A.}\binits{A.}} \AND
  \bauthor{\bsnm{De~Vito},~\bfnm{E.}\binits{E.}}
(\byear{2007}).
\btitle{Optimal rates for the regularized least-squares algorithm}.
\bjournal{Found. Comput. Math.}
\bvolume{7}
\bpages{331--368}.
\end{barticle}
\endbibitem

\bibitem[\protect\citeauthoryear{Chinot, L{\"o}ffler and van~de
  Geer}{2022}]{chinot2022robustness}
\begin{barticle}[author]
\bauthor{\bsnm{Chinot},~\bfnm{Geoffrey}\binits{G.}},
  \bauthor{\bsnm{L{\"o}ffler},~\bfnm{Matthias}\binits{M.}} \AND
  \bauthor{\bparticle{van~de} \bsnm{Geer},~\bfnm{Sara}\binits{S.}}
(\byear{2022}).
\btitle{On the robustness of minimum norm interpolators and regularized
  empirical risk minimizers}.
\bjournal{Ann. Statist.}
\bvolume{50}
\bpages{2306--2333}.
\end{barticle}
\endbibitem

\bibitem[\protect\citeauthoryear{Chizat, Oyallon and
  Bach}{2019}]{chizat2019lazy}
\begin{barticle}[author]
\bauthor{\bsnm{Chizat},~\bfnm{Lenaic}\binits{L.}},
  \bauthor{\bsnm{Oyallon},~\bfnm{Edouard}\binits{E.}} \AND
  \bauthor{\bsnm{Bach},~\bfnm{Francis}\binits{F.}}
(\byear{2019}).
\btitle{On lazy training in differentiable programming}.
\bjournal{Advances in Neural Information Processing Systems}
\bvolume{32}
\bpages{2937--2947}.
\end{barticle}
\endbibitem

\bibitem[\protect\citeauthoryear{Cover}{1965}]{cover1965geometrical}
\begin{barticle}[author]
\bauthor{\bsnm{Cover},~\bfnm{Thomas~M.}\binits{T.~M.}}
(\byear{1965}).
\btitle{Geometrical and statistical properties of systems of linear
  inequalities with applications in pattern recognition}.
\bjournal{IEEE Trans. Electron. Comput.}
\bvolume{EC-14}
\bpages{326--334}.
\end{barticle}
\endbibitem

\bibitem[\protect\citeauthoryear{Deng, Kammoun and
  Thrampoulidis}{2022}]{deng2022model}
\begin{barticle}[author]
\bauthor{\bsnm{Deng},~\bfnm{Zeyu}\binits{Z.}},
  \bauthor{\bsnm{Kammoun},~\bfnm{Abla}\binits{A.}} \AND
  \bauthor{\bsnm{Thrampoulidis},~\bfnm{Christos}\binits{C.}}
(\byear{2022}).
\btitle{A model of double descent for high-dimensional binary linear
  classification}.
\bjournal{Inf. Inference}
\bvolume{11}
\bpages{435--495}.
\end{barticle}
\endbibitem

\bibitem[\protect\citeauthoryear{Derumigny and
  Schmidt-Hieber}{2023}]{derumigny2023lower}
\begin{barticle}[author]
\bauthor{\bsnm{Derumigny},~\bfnm{Alexis}\binits{A.}} \AND
  \bauthor{\bsnm{Schmidt-Hieber},~\bfnm{Johannes}\binits{J.}}
(\byear{2023}).
\btitle{On lower bounds for the bias-variance trade-off}.
\bjournal{Ann. Statist.}
\end{barticle}
\endbibitem

\bibitem[\protect\citeauthoryear{Dou and Liang}{2021}]{dou2021training}
\begin{barticle}[author]
\bauthor{\bsnm{Dou},~\bfnm{Xialiang}\binits{X.}} \AND
  \bauthor{\bsnm{Liang},~\bfnm{Tengyuan}\binits{T.}}
(\byear{2021}).
\btitle{Training neural networks as learning data-adaptive kernels: Provable
  representation and approximation benefits}.
\bjournal{J. Amer. Statist. Assoc.}
\bvolume{116}
\bpages{1507--1520}.
\end{barticle}
\endbibitem

\bibitem[\protect\citeauthoryear{E, Ma and Wu}{2019}]{e2019priori}
\begin{barticle}[author]
\bauthor{\bsnm{E},~\bfnm{Weinan}\binits{W.}},
  \bauthor{\bsnm{Ma},~\bfnm{Chao}\binits{C.}} \AND
  \bauthor{\bsnm{Wu},~\bfnm{Lei}\binits{L.}}
(\byear{2019}).
\btitle{A priori estimates of the population risk for two-layer neural
  networks}.
\bjournal{Commun. Math. Sci.}
\bvolume{17}
\bpages{1407--1425}.
\end{barticle}
\endbibitem

\bibitem[\protect\citeauthoryear{E, Ma and Wu}{2020}]{e2020comparative}
\begin{barticle}[author]
\bauthor{\bsnm{E},~\bfnm{Weinan}\binits{W.}},
  \bauthor{\bsnm{Ma},~\bfnm{Chao}\binits{C.}} \AND
  \bauthor{\bsnm{Wu},~\bfnm{Lei}\binits{L.}}
(\byear{2020}).
\btitle{A comparative analysis of optimization and generalization properties of
  two-layer neural network and random feature models under gradient descent
  dynamics}.
\bjournal{Sci. China Math.}
\bvolume{63}
\bpages{1235--1258}.
\end{barticle}
\endbibitem

\bibitem[\protect\citeauthoryear{Efron}{2020}]{efron2020prediction}
\begin{barticle}[author]
\bauthor{\bsnm{Efron},~\bfnm{Bradley}\binits{B.}}
(\byear{2020}).
\btitle{Prediction, estimation, and attribution}.
\bjournal{J. Amer. Statist. Assoc.}
\bvolume{115}
\bpages{636--655}.
\end{barticle}
\endbibitem

\bibitem[\protect\citeauthoryear{Ergen and Pilanci}{2021}]{ergen2021revealing}
\begin{binproceedings}[author]
\bauthor{\bsnm{Ergen},~\bfnm{Tolga}\binits{T.}} \AND
  \bauthor{\bsnm{Pilanci},~\bfnm{Mert}\binits{M.}}
(\byear{2021}).
\btitle{Revealing the structure of deep neural networks via convex duality}.
In \bbooktitle{Proceedings of the 38th International Conference on Machine
  Learning}
\bpages{3004--3014}.
\end{binproceedings}
\endbibitem

\bibitem[\protect\citeauthoryear{Esteva et~al.}{2017}]{esteva2017dermatologist}
\begin{barticle}[author]
\bauthor{\bsnm{Esteva},~\bfnm{Andre}\binits{A.}},
  \bauthor{\bsnm{Kuprel},~\bfnm{Brett}\binits{B.}},
  \bauthor{\bsnm{Novoa},~\bfnm{Roberto~A.}\binits{R.~A.}},
  \bauthor{\bsnm{Ko},~\bfnm{Justin}\binits{J.}},
  \bauthor{\bsnm{Swetter},~\bfnm{Susan~M.}\binits{S.~M.}},
  \bauthor{\bsnm{Blau},~\bfnm{Helen~M.}\binits{H.~M.}} \AND
  \bauthor{\bsnm{Thrun},~\bfnm{Sebastian}\binits{S.}}
(\byear{2017}).
\btitle{Dermatologist-level classification of skin cancer with deep neural
  networks}.
\bjournal{Nature}
\bvolume{542}
\bpages{115--118}.
\end{barticle}
\endbibitem

\bibitem[\protect\citeauthoryear{Farrell, Liang and
  Misra}{2021}]{farrell2021deep}
\begin{barticle}[author]
\bauthor{\bsnm{Farrell},~\bfnm{Max~H.}\binits{M.~H.}},
  \bauthor{\bsnm{Liang},~\bfnm{Tengyuan}\binits{T.}} \AND
  \bauthor{\bsnm{Misra},~\bfnm{Sanjog}\binits{S.}}
(\byear{2021}).
\btitle{Deep neural networks for estimation and inference}.
\bjournal{Econometrica}
\bvolume{89}
\bpages{181--213}.
\end{barticle}
\endbibitem

\bibitem[\protect\citeauthoryear{Geman, Bienenstock and
  Doursat}{1992}]{geman1992neural}
\begin{barticle}[author]
\bauthor{\bsnm{Geman},~\bfnm{Stuart}\binits{S.}},
  \bauthor{\bsnm{Bienenstock},~\bfnm{Elie}\binits{E.}} \AND
  \bauthor{\bsnm{Doursat},~\bfnm{Ren{\'e}}\binits{R.}}
(\byear{1992}).
\btitle{Neural networks and the bias/variance dilemma}.
\bjournal{Neural Comput.}
\bvolume{4}
\bpages{1--58}.
\end{barticle}
\endbibitem

\bibitem[\protect\citeauthoryear{Ghorbani
  et~al.}{2021}]{ghorbani2021linearized}
\begin{barticle}[author]
\bauthor{\bsnm{Ghorbani},~\bfnm{Behrooz}\binits{B.}},
  \bauthor{\bsnm{Mei},~\bfnm{Song}\binits{S.}},
  \bauthor{\bsnm{Misiakiewicz},~\bfnm{Theodor}\binits{T.}} \AND
  \bauthor{\bsnm{Montanari},~\bfnm{Andrea}\binits{A.}}
(\byear{2021}).
\btitle{Linearized two-layers neural networks in high dimension}.
\bjournal{Ann. Statist.}
\bvolume{49}
\bpages{1029--1054}.
\end{barticle}
\endbibitem

\bibitem[\protect\citeauthoryear{Golowich, Rakhlin and
  Shamir}{2020}]{golowich2020size}
\begin{barticle}[author]
\bauthor{\bsnm{Golowich},~\bfnm{Noah}\binits{N.}},
  \bauthor{\bsnm{Rakhlin},~\bfnm{Alexander}\binits{A.}} \AND
  \bauthor{\bsnm{Shamir},~\bfnm{Ohad}\binits{O.}}
(\byear{2020}).
\btitle{Size-independent sample complexity of neural networks}.
\bjournal{Inf. Inference}
\bvolume{9}
\bpages{473--504}.
\end{barticle}
\endbibitem

\bibitem[\protect\citeauthoryear{Goodfellow, Bengio and
  Courville}{2016}]{goodfellow2016deep}
\begin{bbook}[author]
\bauthor{\bsnm{Goodfellow},~\bfnm{Ian}\binits{I.}},
  \bauthor{\bsnm{Bengio},~\bfnm{Yoshua}\binits{Y.}} \AND
  \bauthor{\bsnm{Courville},~\bfnm{Aaron}\binits{A.}}
(\byear{2016}).
\btitle{Deep Learning}.
\bpublisher{MIT Press}, \baddress{Cambridge, MA}.
\end{bbook}
\endbibitem

\bibitem[\protect\citeauthoryear{Hastie et~al.}{2022}]{hastie2022surprises}
\begin{barticle}[author]
\bauthor{\bsnm{Hastie},~\bfnm{Trevor}\binits{T.}},
  \bauthor{\bsnm{Montanari},~\bfnm{Andrea}\binits{A.}},
  \bauthor{\bsnm{Rosset},~\bfnm{Saharon}\binits{S.}} \AND
  \bauthor{\bsnm{Tibshirani},~\bfnm{Ryan~J.}\binits{R.~J.}}
(\byear{2022}).
\btitle{Surprises in high-dimensional ridgeless least squares interpolation}.
\bjournal{Ann. Statist.}
\bvolume{50}
\bpages{949--986}.
\end{barticle}
\endbibitem

\bibitem[\protect\citeauthoryear{Hayakawa and
  Suzuki}{2020}]{hayakawa2020minimax}
\begin{barticle}[author]
\bauthor{\bsnm{Hayakawa},~\bfnm{Satoshi}\binits{S.}} \AND
  \bauthor{\bsnm{Suzuki},~\bfnm{Taiji}\binits{T.}}
(\byear{2020}).
\btitle{On the minimax optimality and superiority of deep neural network
  learning over sparse parameter spaces}.
\bjournal{Neural Netw.}
\bvolume{123}
\bpages{343--361}.
\end{barticle}
\endbibitem

\bibitem[\protect\citeauthoryear{He et~al.}{2016}]{he2016deep}
\begin{binproceedings}[author]
\bauthor{\bsnm{He},~\bfnm{Kaiming}\binits{K.}},
  \bauthor{\bsnm{Zhang},~\bfnm{Xiangyu}\binits{X.}},
  \bauthor{\bsnm{Ren},~\bfnm{Shaoqing}\binits{S.}} \AND
  \bauthor{\bsnm{Sun},~\bfnm{Jian}\binits{J.}}
(\byear{2016}).
\btitle{Deep residual learning for image recognition}.
In \bbooktitle{Proceedings of the IEEE Conference on Computer Vision and
  Pattern Recognition}
\bpages{770--778}.
\end{binproceedings}
\endbibitem

\bibitem[\protect\citeauthoryear{Hoerl}{2020}]{hoerl2020ridge}
\begin{barticle}[author]
\bauthor{\bsnm{Hoerl},~\bfnm{Roger~W.}\binits{R.~W.}}
(\byear{2020}).
\btitle{Ridge regression: A historical context}.
\bjournal{Technometrics}
\bvolume{62}
\bpages{420--425}.
\end{barticle}
\endbibitem

\bibitem[\protect\citeauthoryear{Hsu, Kakade and Zhang}{2012}]{hsu2012tail}
\begin{barticle}[author]
\bauthor{\bsnm{Hsu},~\bfnm{Daniel}\binits{D.}},
  \bauthor{\bsnm{Kakade},~\bfnm{Sham~M.}\binits{S.~M.}} \AND
  \bauthor{\bsnm{Zhang},~\bfnm{Tong}\binits{T.}}
(\byear{2012}).
\btitle{A tail inequality for quadratic forms of subgaussian random vectors}.
\bjournal{Electron. Commun. Probab.}
\bvolume{17}
\bpages{1--6}.
\end{barticle}
\endbibitem

\bibitem[\protect\citeauthoryear{Jacot, Gabriel and
  Hongler}{2018}]{jacot2018neural}
\begin{binproceedings}[author]
\bauthor{\bsnm{Jacot},~\bfnm{Arthur}\binits{A.}},
  \bauthor{\bsnm{Gabriel},~\bfnm{Franck}\binits{F.}} \AND
  \bauthor{\bsnm{Hongler},~\bfnm{Cl{\'e}ment}\binits{C.}}
(\byear{2018}).
\btitle{Neural tangent kernel: Convergence and generalization in neural
  networks}.
In \bbooktitle{Advances in Neural Information Processing Systems}
\bvolume{31}
\bpages{8580--8589}.
\end{binproceedings}
\endbibitem

\bibitem[\protect\citeauthoryear{Jarrett et~al.}{2009}]{jarrett2009best}
\begin{binproceedings}[author]
\bauthor{\bsnm{Jarrett},~\bfnm{Kevin}\binits{K.}},
  \bauthor{\bsnm{Kavukcuoglu},~\bfnm{Koray}\binits{K.}},
  \bauthor{\bsnm{Ranzato},~\bfnm{Marc'Aurelio}\binits{M.}} \AND
  \bauthor{\bsnm{LeCun},~\bfnm{Yann}\binits{Y.}}
(\byear{2009}).
\btitle{What is the best multi-stage architecture for object recognition?}
In \bbooktitle{IEEE 12th International Conference on Computer Vision}
\bpages{2146--2153}.
\end{binproceedings}
\endbibitem

\bibitem[\protect\citeauthoryear{Ji and
  Telgarsky}{2020}]{ji2020polylogarithmic}
\begin{binproceedings}[author]
\bauthor{\bsnm{Ji},~\bfnm{Ziwei}\binits{Z.}} \AND
  \bauthor{\bsnm{Telgarsky},~\bfnm{Matus}\binits{M.}}
(\byear{2020}).
\btitle{Polylogarithmic width suffices for gradient descent to achieve
  arbitrarily small test error with shallow {ReLU} networks}.
In \bbooktitle{International Conference on Learning Representations}.
\end{binproceedings}
\endbibitem

\bibitem[\protect\citeauthoryear{Jones}{1992}]{jones1992simple}
\begin{barticle}[author]
\bauthor{\bsnm{Jones},~\bfnm{Lee~K.}\binits{L.~K.}}
(\byear{1992}).
\btitle{A simple lemma on greedy approximation in {H}ilbert space and
  convergence rates for projection pursuit regression and neural network
  training}.
\bjournal{Ann. Statist.}
\bvolume{20}
\bpages{608--613}.
\end{barticle}
\endbibitem

\bibitem[\protect\citeauthoryear{Klusowski and
  Barron}{2018}]{klusowski2018approximation}
\begin{barticle}[author]
\bauthor{\bsnm{Klusowski},~\bfnm{Jason~M.}\binits{J.~M.}} \AND
  \bauthor{\bsnm{Barron},~\bfnm{Andrew~R.}\binits{A.~R.}}
(\byear{2018}).
\btitle{Approximation by combinations of {ReLU} and squared {ReLU} ridge
  functions with $\ell^1$ and $\ell^0$ controls}.
\bjournal{IEEE Trans. Inf. Theory}
\bvolume{64}
\bpages{7649--7656}.
\end{barticle}
\endbibitem

\bibitem[\protect\citeauthoryear{Kohler and Langer}{2021}]{kohler2021rate}
\begin{barticle}[author]
\bauthor{\bsnm{Kohler},~\bfnm{Michael}\binits{M.}} \AND
  \bauthor{\bsnm{Langer},~\bfnm{Sophie}\binits{S.}}
(\byear{2021}).
\btitle{On the rate of convergence of fully connected deep neural network
  regression estimates}.
\bjournal{Ann. Statist.}
\bvolume{49}
\bpages{2231--2249}.
\end{barticle}
\endbibitem

\bibitem[\protect\citeauthoryear{Krizhevsky, Sutskever and
  Hinton}{2012}]{krizhevsky2012imagenet}
\begin{binproceedings}[author]
\bauthor{\bsnm{Krizhevsky},~\bfnm{Alex}\binits{A.}},
  \bauthor{\bsnm{Sutskever},~\bfnm{Ilya}\binits{I.}} \AND
  \bauthor{\bsnm{Hinton},~\bfnm{Geoffrey~E.}\binits{G.~E.}}
(\byear{2012}).
\btitle{{ImageNet} classification with deep convolutional neural networks}.
In \bbooktitle{Advances in Neural Information Processing Systems}
\bvolume{25}
\bpages{1097--1105}.
\end{binproceedings}
\endbibitem

\bibitem[\protect\citeauthoryear{Krogh and Hertz}{1991}]{krogh1991simple}
\begin{barticle}[author]
\bauthor{\bsnm{Krogh},~\bfnm{A.}\binits{A.}} \AND
  \bauthor{\bsnm{Hertz},~\bfnm{J.~A.}\binits{J.~A.}}
(\byear{1991}).
\btitle{A simple weight decay can improve generalization}.
\bjournal{Advances in Neural Information Processing Systems}
\bvolume{4}
\bpages{950--957}.
\end{barticle}
\endbibitem

\bibitem[\protect\citeauthoryear{Li and Meng}{2021}]{li2021multi}
\begin{barticle}[author]
\bauthor{\bsnm{Li},~\bfnm{Xinran}\binits{X.}} \AND
  \bauthor{\bsnm{Meng},~\bfnm{Xiao-Li}\binits{X.-L.}}
(\byear{2021}).
\btitle{A multi-resolution theory for approximating infinite-$p$-zero-$n$:
  Transitional inference, individualized predictions, and a world without
  bias-variance tradeoff}.
\bjournal{J. Amer. Statist. Assoc.}
\bvolume{116}
\bpages{353--367}.
\end{barticle}
\endbibitem

\bibitem[\protect\citeauthoryear{Liang, Rakhlin and
  Zhai}{2020}]{liang2020multiple}
\begin{binproceedings}[author]
\bauthor{\bsnm{Liang},~\bfnm{Tengyuan}\binits{T.}},
  \bauthor{\bsnm{Rakhlin},~\bfnm{Alexander}\binits{A.}} \AND
  \bauthor{\bsnm{Zhai},~\bfnm{Xiyu}\binits{X.}}
(\byear{2020}).
\btitle{On the multiple descent of minimum-norm interpolants and restricted
  lower isometry of kernels}.
In \bbooktitle{Proceedings of the 33rd Conference on Learning Theory}
\bpages{2683--2711}.
\end{binproceedings}
\endbibitem

\bibitem[\protect\citeauthoryear{Liang and Sur}{2022}]{liang2022precise}
\begin{barticle}[author]
\bauthor{\bsnm{Liang},~\bfnm{Tengyuan}\binits{T.}} \AND
  \bauthor{\bsnm{Sur},~\bfnm{Pragya}\binits{P.}}
(\byear{2022}).
\btitle{A precise high-dimensional asymptotic theory for boosting and
  minimum-$\ell_1$-norm interpolated classifiers}.
\bjournal{Ann. Statist.}
\bvolume{50}
\bpages{1669--1695}.
\end{barticle}
\endbibitem

\bibitem[\protect\citeauthoryear{Makovoz}{1996}]{makovoz1996random}
\begin{barticle}[author]
\bauthor{\bsnm{Makovoz},~\bfnm{Y.}\binits{Y.}}
(\byear{1996}).
\btitle{Random approximants and neural networks}.
\bjournal{J. Approx. Theory}
\bvolume{85}
\bpages{98--109}.
\end{barticle}
\endbibitem

\bibitem[\protect\citeauthoryear{Matou\v{s}ek}{1996}]{matousek1996improved}
\begin{barticle}[author]
\bauthor{\bsnm{Matou\v{s}ek},~\bfnm{J.}\binits{J.}}
(\byear{1996}).
\btitle{Improved upper bounds for approximation by zonotopes}.
\bjournal{Acta Math.}
\bvolume{177}
\bpages{55--73}.
\end{barticle}
\endbibitem

\bibitem[\protect\citeauthoryear{Mei, Montanari and Nguyen}{2018}]{mei2018mean}
\begin{barticle}[author]
\bauthor{\bsnm{Mei},~\bfnm{Song}\binits{S.}},
  \bauthor{\bsnm{Montanari},~\bfnm{Andrea}\binits{A.}} \AND
  \bauthor{\bsnm{Nguyen},~\bfnm{Phan-Minh}\binits{P.-M.}}
(\byear{2018}).
\btitle{A mean field view of the landscape of two-layer neural networks}.
\bjournal{Proc. Natl. Acad. Sci. USA}
\bvolume{115}
\bpages{E7665--E7671}.
\end{barticle}
\endbibitem

\bibitem[\protect\citeauthoryear{Mei and
  Montanari}{2022}]{mei2022generalization}
\begin{barticle}[author]
\bauthor{\bsnm{Mei},~\bfnm{Song}\binits{S.}} \AND
  \bauthor{\bsnm{Montanari},~\bfnm{Andrea}\binits{A.}}
(\byear{2022}).
\btitle{The generalization error of random features regression: Precise
  asymptotics and the double descent curve}.
\bjournal{Comm. Pure Appl. Math.}
\bvolume{75}
\bpages{667--766}.
\end{barticle}
\endbibitem

\bibitem[\protect\citeauthoryear{Montanari and
  Zhong}{2022}]{montanari2022interpolation}
\begin{barticle}[author]
\bauthor{\bsnm{Montanari},~\bfnm{Andrea}\binits{A.}} \AND
  \bauthor{\bsnm{Zhong},~\bfnm{Yiqiao}\binits{Y.}}
(\byear{2022}).
\btitle{The interpolation phase transition in neural networks: Memorization and
  generalization under lazy training}.
\bjournal{Ann. Statist.}
\bvolume{50}
\bpages{2816--2847}.
\end{barticle}
\endbibitem

\bibitem[\protect\citeauthoryear{Muthukumar
  et~al.}{2020}]{muthukumar2020harmless}
\begin{barticle}[author]
\bauthor{\bsnm{Muthukumar},~\bfnm{Vidya}\binits{V.}},
  \bauthor{\bsnm{Vodrahalli},~\bfnm{Kailas}\binits{K.}},
  \bauthor{\bsnm{Subramanian},~\bfnm{Vignesh}\binits{V.}} \AND
  \bauthor{\bsnm{Sahai},~\bfnm{Anant}\binits{A.}}
(\byear{2020}).
\btitle{Harmless interpolation of moisy data in regression}.
\bjournal{IEEE J. Sel. Areas Inform. Theory}
\bvolume{1}
\bpages{67-83}.
\end{barticle}
\endbibitem

\bibitem[\protect\citeauthoryear{Nakkiran et~al.}{2021}]{nakkiran2021optimal}
\begin{binproceedings}[author]
\bauthor{\bsnm{Nakkiran},~\bfnm{Preetum}\binits{P.}},
  \bauthor{\bsnm{Venkat},~\bfnm{Prayaag}\binits{P.}},
  \bauthor{\bsnm{Kakade},~\bfnm{Sham}\binits{S.}} \AND
  \bauthor{\bsnm{Ma},~\bfnm{Tengyu}\binits{T.}}
(\byear{2021}).
\btitle{Optimal regularization can mitigate double descent}.
In \bbooktitle{International Conference on Learning Representations}.
\end{binproceedings}
\endbibitem

\bibitem[\protect\citeauthoryear{Neyshabur, Tomioka and
  Srebro}{2015a}]{neyshabur2015norm}
\begin{binproceedings}[author]
\bauthor{\bsnm{Neyshabur},~\bfnm{Behnam}\binits{B.}},
  \bauthor{\bsnm{Tomioka},~\bfnm{Ryota}\binits{R.}} \AND
  \bauthor{\bsnm{Srebro},~\bfnm{Nathan}\binits{N.}}
(\byear{2015}a).
\btitle{Norm-based capacity control in neural networks}.
In \bbooktitle{Proceedings of the 28th Conference on Learning Theory}
\bpages{1376--1401}.
\end{binproceedings}
\endbibitem

\bibitem[\protect\citeauthoryear{Neyshabur, Tomioka and
  Srebro}{2015b}]{neyshabur2015search}
\begin{binproceedings}[author]
\bauthor{\bsnm{Neyshabur},~\bfnm{Behnam}\binits{B.}},
  \bauthor{\bsnm{Tomioka},~\bfnm{Ryota}\binits{R.}} \AND
  \bauthor{\bsnm{Srebro},~\bfnm{Nathan}\binits{N.}}
(\byear{2015}b).
\btitle{In search of the real inductive bias: On the role of implicit
  regularization in deep learning}.
In \bbooktitle{International Conference on Learning Representations}.
\end{binproceedings}
\endbibitem

\bibitem[\protect\citeauthoryear{Neyshabur et~al.}{2019}]{neyshabur2019role}
\begin{binproceedings}[author]
\bauthor{\bsnm{Neyshabur},~\bfnm{Behnam}\binits{B.}},
  \bauthor{\bsnm{Li},~\bfnm{Zhiyuan}\binits{Z.}},
  \bauthor{\bsnm{Bhojanapalli},~\bfnm{Srinadh}\binits{S.}},
  \bauthor{\bsnm{LeCun},~\bfnm{Yann}\binits{Y.}} \AND
  \bauthor{\bsnm{Srebro},~\bfnm{Nathan}\binits{N.}}
(\byear{2019}).
\btitle{The role of over-parametrization in generalization of neural networks}.
In \bbooktitle{International Conference on Learning Representations}.
\end{binproceedings}
\endbibitem

\bibitem[\protect\citeauthoryear{Ongie et~al.}{2020}]{ongie2020function}
\begin{binproceedings}[author]
\bauthor{\bsnm{Ongie},~\bfnm{Greg}\binits{G.}},
  \bauthor{\bsnm{Willett},~\bfnm{Rebecca}\binits{R.}},
  \bauthor{\bsnm{Soudry},~\bfnm{Daniel}\binits{D.}} \AND
  \bauthor{\bsnm{Srebro},~\bfnm{Nathan}\binits{N.}}
(\byear{2020}).
\btitle{A function space view of bounded norm infinite width {ReLU} nets: The
  multivariate case}.
In \bbooktitle{International Conference on Learning Representations}.
\end{binproceedings}
\endbibitem

\bibitem[\protect\citeauthoryear{Parhi and Nowak}{2021}]{parhi2021banach}
\begin{barticle}[author]
\bauthor{\bsnm{Parhi},~\bfnm{Rahul}\binits{R.}} \AND
  \bauthor{\bsnm{Nowak},~\bfnm{Robert~D.}\binits{R.~D.}}
(\byear{2021}).
\btitle{Banach space representer theorems for neural networks and ridge
  splines}.
\bjournal{J. Mach. Learn. Res.}
\bvolume{22}
\bpages{1--40}.
\end{barticle}
\endbibitem

\bibitem[\protect\citeauthoryear{Parhi and Nowak}{2022}]{parhi2022near}
\begin{barticle}[author]
\bauthor{\bsnm{Parhi},~\bfnm{Rahul}\binits{R.}} \AND
  \bauthor{\bsnm{Nowak},~\bfnm{Robert~D.}\binits{R.~D.}}
(\byear{2022}).
\btitle{Near-minimax optimal estimation with shallow {ReLU} neural networks}.
\bjournal{IEEE Trans. Inf. Theory}.
\end{barticle}
\endbibitem

\bibitem[\protect\citeauthoryear{Pilanci and Ergen}{2020}]{pilanci2020neural}
\begin{binproceedings}[author]
\bauthor{\bsnm{Pilanci},~\bfnm{Mert}\binits{M.}} \AND
  \bauthor{\bsnm{Ergen},~\bfnm{Tolga}\binits{T.}}
(\byear{2020}).
\btitle{Neural networks are convex regularizers: Exact polynomial-time convex
  optimization formulations for two-layer networks}.
In \bbooktitle{Proceedings of the 37th International Conference on Machine
  Learning}
\bpages{7695--7705}.
\end{binproceedings}
\endbibitem

\bibitem[\protect\citeauthoryear{Rahimi and Recht}{2007}]{rahimi2007random}
\begin{binproceedings}[author]
\bauthor{\bsnm{Rahimi},~\bfnm{Ali}\binits{A.}} \AND
  \bauthor{\bsnm{Recht},~\bfnm{Benjamin}\binits{B.}}
(\byear{2007}).
\btitle{Random features for large-scale kernel machines}.
In \bbooktitle{Advances in Neural Information Processing Systems}
\bvolume{20}
\bpages{1177--1184}.
\end{binproceedings}
\endbibitem

\bibitem[\protect\citeauthoryear{Rotskoff and
  Vanden-Eijnden}{2022}]{rotskoff2022trainability}
\begin{barticle}[author]
\bauthor{\bsnm{Rotskoff},~\bfnm{G.~M.}\binits{G.~M.}} \AND
  \bauthor{\bsnm{Vanden-Eijnden},~\bfnm{E.}\binits{E.}}
(\byear{2022}).
\btitle{Trainability and accuracy of artificial neural networks: An interacting
  particle system approach}.
\bjournal{Comm. Pure Appl. Math.}
\bvolume{75}
\bpages{1889--1935}.
\end{barticle}
\endbibitem

\bibitem[\protect\citeauthoryear{Schmidt-Hieber}{2020}]{schmidt-hieber2020nonparametric}
\begin{barticle}[author]
\bauthor{\bsnm{Schmidt-Hieber},~\bfnm{Johannes}\binits{J.}}
(\byear{2020}).
\btitle{Nonparametric regression using deep neural networks with {ReLU}
  activation function}.
\bjournal{Ann. Statist.}
\bvolume{48}
\bpages{1875--1897}.
\end{barticle}
\endbibitem

\bibitem[\protect\citeauthoryear{Schrittwieser
  et~al.}{2020}]{schrittwieser2020mastering}
\begin{barticle}[author]
\bauthor{\bsnm{Schrittwieser},~\bfnm{Julian}\binits{J.}},
  \bauthor{\bsnm{Antonoglou},~\bfnm{Ioannis}\binits{I.}},
  \bauthor{\bsnm{Hubert},~\bfnm{Thomas}\binits{T.}},
  \bauthor{\bsnm{Simonyan},~\bfnm{Karen}\binits{K.}},
  \bauthor{\bsnm{Sifre},~\bfnm{Laurent}\binits{L.}},
  \bauthor{\bsnm{Schmitt},~\bfnm{Simon}\binits{S.}},
  \bauthor{\bsnm{Guez},~\bfnm{Arthur}\binits{A.}},
  \bauthor{\bsnm{Lockhart},~\bfnm{Edward}\binits{E.}},
  \bauthor{\bsnm{Hassabis},~\bfnm{Demis}\binits{D.}},
  \bauthor{\bsnm{Graepel},~\bfnm{Thore}\binits{T.}},
  \bauthor{\bsnm{Lillicrap},~\bfnm{Timothy}\binits{T.}} \AND
  \bauthor{\bsnm{Silver},~\bfnm{David}\binits{D.}}
(\byear{2020}).
\btitle{Mastering {A}tari, {G}o, chess and shogi by planning with a learned
  model}.
\bjournal{Nature}
\bvolume{588}
\bpages{604--609}.
\end{barticle}
\endbibitem

\bibitem[\protect\citeauthoryear{Siegel and Xu}{2020}]{siegel2020approximation}
\begin{barticle}[author]
\bauthor{\bsnm{Siegel},~\bfnm{Jonathan~W.}\binits{J.~W.}} \AND
  \bauthor{\bsnm{Xu},~\bfnm{Jinchao}\binits{J.}}
(\byear{2020}).
\btitle{Approximation rates for neural networks with general activation
  functions}.
\bjournal{Neural Netw.}
\bvolume{128}
\bpages{313--321}.
\end{barticle}
\endbibitem

\bibitem[\protect\citeauthoryear{Siegel and Xu}{2022}]{siegel2022sharp}
\begin{barticle}[author]
\bauthor{\bsnm{Siegel},~\bfnm{Jonathan~W.}\binits{J.~W.}} \AND
  \bauthor{\bsnm{Xu},~\bfnm{Jinchao}\binits{J.}}
(\byear{2022}).
\btitle{Sharp bounds on the approximation rates, metric entropy, and $n$-widths
  of shallow neural networks}.
\bjournal{Found. Comput. Math.}
\end{barticle}
\endbibitem

\bibitem[\protect\citeauthoryear{Sirignano and
  Spiliopoulos}{2020}]{sirignano2020mean}
\begin{barticle}[author]
\bauthor{\bsnm{Sirignano},~\bfnm{Justin}\binits{J.}} \AND
  \bauthor{\bsnm{Spiliopoulos},~\bfnm{Konstantinos}\binits{K.}}
(\byear{2020}).
\btitle{Mean field analysis of neural networks: A law of large numbers}.
\bjournal{SIAM J. Appl. Math.}
\bvolume{80}
\bpages{725--752}.
\end{barticle}
\endbibitem

\bibitem[\protect\citeauthoryear{Sj{\"o}berg and
  Ljung}{1995}]{sjoberg1995overtraining}
\begin{barticle}[author]
\bauthor{\bsnm{Sj{\"o}berg},~\bfnm{J.}\binits{J.}} \AND
  \bauthor{\bsnm{Ljung},~\bfnm{L.}\binits{L.}}
(\byear{1995}).
\btitle{Overtraining, regularization and searching for a minimum, with
  application to neural networks}.
\bjournal{Internat. J. Control}
\bvolume{62}
\bpages{1391--1407}.
\end{barticle}
\endbibitem

\bibitem[\protect\citeauthoryear{Soltanolkotabi, Javanmard and
  Lee}{2019}]{soltanolkotabi2019theoretical}
\begin{barticle}[author]
\bauthor{\bsnm{Soltanolkotabi},~\bfnm{M.}\binits{M.}},
  \bauthor{\bsnm{Javanmard},~\bfnm{A.}\binits{A.}} \AND
  \bauthor{\bsnm{Lee},~\bfnm{J.~D.}\binits{J.~D.}}
(\byear{2019}).
\btitle{Theoretical insights into the optimization landscape of
  over-parameterized shallow neural networks}.
\bjournal{IEEE Trans. Inf. Theory}
\bvolume{65}
\bpages{742--769}.
\end{barticle}
\endbibitem

\bibitem[\protect\citeauthoryear{Srebro, Rennie and
  Jaakkola}{2004}]{srebro2004maximum}
\begin{binproceedings}[author]
\bauthor{\bsnm{Srebro},~\bfnm{Nathan}\binits{N.}},
  \bauthor{\bsnm{Rennie},~\bfnm{Jason D.~M.}\binits{J.~D.~M.}} \AND
  \bauthor{\bsnm{Jaakkola},~\bfnm{Tommi~S.}\binits{T.~S.}}
(\byear{2004}).
\btitle{Maximum-margin matrix factorization}.
In \bbooktitle{Advances in Neural Information Processing Systems}
\bvolume{17}
\bpages{1329--1336}.
\end{binproceedings}
\endbibitem

\bibitem[\protect\citeauthoryear{Srivastava
  et~al.}{2014}]{srivastava2014dropout}
\begin{barticle}[author]
\bauthor{\bsnm{Srivastava},~\bfnm{Nitish}\binits{N.}},
  \bauthor{\bsnm{Hinton},~\bfnm{Geoffrey}\binits{G.}},
  \bauthor{\bsnm{Krizhevsky},~\bfnm{Alex}\binits{A.}},
  \bauthor{\bsnm{Sutskever},~\bfnm{Ilya}\binits{I.}} \AND
  \bauthor{\bsnm{Salakhutdinov},~\bfnm{Ruslan}\binits{R.}}
(\byear{2014}).
\btitle{Dropout: A simple way to prevent neural networks from overfitting}.
\bjournal{J. Mach. Learn. Res.}
\bvolume{15}
\bpages{1929--1958}.
\end{barticle}
\endbibitem

\bibitem[\protect\citeauthoryear{Sutskever, Vinyals and
  Le}{2014}]{sutskever2014sequence}
\begin{binproceedings}[author]
\bauthor{\bsnm{Sutskever},~\bfnm{Ilya}\binits{I.}},
  \bauthor{\bsnm{Vinyals},~\bfnm{Oriol}\binits{O.}} \AND
  \bauthor{\bsnm{Le},~\bfnm{Quoc~V.}\binits{Q.~V.}}
(\byear{2014}).
\btitle{Sequence to sequence learning with neural networks}.
In \bbooktitle{Advances in Neural Information Processing Systems}
\bvolume{27}
\bpages{3104--3112}.
\end{binproceedings}
\endbibitem

\bibitem[\protect\citeauthoryear{Wainwright}{2019}]{wainwright2019high}
\begin{bbook}[author]
\bauthor{\bsnm{Wainwright},~\bfnm{M.~J.}\binits{M.~J.}}
(\byear{2019}).
\btitle{High-Dimensional Statistics: A Non-Asymptotic Viewpoint}.
\bpublisher{Cambridge Univ. Press}, \baddress{Cambridge}.
\end{bbook}
\endbibitem

\bibitem[\protect\citeauthoryear{Wang, Lacotte and
  Pilanci}{2022}]{wang2022hidden}
\begin{binproceedings}[author]
\bauthor{\bsnm{Wang},~\bfnm{Yifei}\binits{Y.}},
  \bauthor{\bsnm{Lacotte},~\bfnm{Jonathan}\binits{J.}} \AND
  \bauthor{\bsnm{Pilanci},~\bfnm{Mert}\binits{M.}}
(\byear{2022}).
\btitle{The hidden convex optimization landscape of regularized two-layer
  {ReLU} networks: An exact characterization of optimal solutions}.
In \bbooktitle{International Conference on Learning Representations}.
\end{binproceedings}
\endbibitem

\bibitem[\protect\citeauthoryear{Yuan and Lin}{2006}]{yuan2006model}
\begin{barticle}[author]
\bauthor{\bsnm{Yuan},~\bfnm{Ming}\binits{M.}} \AND
  \bauthor{\bsnm{Lin},~\bfnm{Yi}\binits{Y.}}
(\byear{2006}).
\btitle{Model selection and estimation in regression with grouped variables}.
\bjournal{J. R. Stat. Soc. Ser. B. Stat. Methodol.}
\bvolume{68}
\bpages{49--67}.
\end{barticle}
\endbibitem

\bibitem[\protect\citeauthoryear{Zhang et~al.}{2021}]{zhang2021understanding}
\begin{barticle}[author]
\bauthor{\bsnm{Zhang},~\bfnm{Chiyuan}\binits{C.}},
  \bauthor{\bsnm{Bengio},~\bfnm{Samy}\binits{S.}},
  \bauthor{\bsnm{Hardt},~\bfnm{Moritz}\binits{M.}},
  \bauthor{\bsnm{Recht},~\bfnm{Benjamin}\binits{B.}} \AND
  \bauthor{\bsnm{Vinyals},~\bfnm{Oriol}\binits{O.}}
(\byear{2021}).
\btitle{Understanding deep learning (still) requires rethinking
  generalization}.
\bjournal{Commun. ACM}
\bvolume{64}
\bpages{107--115}.
\end{barticle}
\endbibitem

\end{thebibliography}


\begin{thebibliography}{12}

\bibitem[\protect\citeauthoryear{Bach}{2017}]{bach2017breaking}
\begin{barticle}[author]
\bauthor{\bsnm{Bach},~\bfnm{Francis}\binits{F.}}
(\byear{2017}).
\btitle{Breaking the curse of dimensionality with convex neural networks}.
\bjournal{J. Mach. Learn. Res.}
\bvolume{18}
\bpages{1--53}.
\end{barticle}
\endbibitem

\bibitem[\protect\citeauthoryear{Barron}{1993}]{barron1993universal}
\begin{barticle}[author]
\bauthor{\bsnm{Barron},~\bfnm{Andrew~R.}\binits{A.~R.}}
(\byear{1993}).
\btitle{Universal approximation bounds for superpositions of a sigmoidal
  function}.
\bjournal{IEEE Trans. Inf. Theory}
\bvolume{39}
\bpages{930--945}.
\end{barticle}
\endbibitem

\bibitem[\protect\citeauthoryear{Donoho}{1993}]{donoho1993unconditional}
\begin{barticle}[author]
\bauthor{\bsnm{Donoho},~\bfnm{David~L.}\binits{D.~L.}}
(\byear{1993}).
\btitle{Unconditional bases are optimal bases for data compression and for
  statistical estimation}.
\bjournal{Appl. Comput. Harmon. Anal.}
\bvolume{1}
\bpages{100--115}.
\end{barticle}
\endbibitem

\bibitem[\protect\citeauthoryear{Dunkl and Xu}{2014}]{dunkl2014orthogonal}
\begin{bbook}[author]
\bauthor{\bsnm{Dunkl},~\bfnm{Charles~F.}\binits{C.~F.}} \AND
  \bauthor{\bsnm{Xu},~\bfnm{Yuan}\binits{Y.}}
(\byear{2014}).
\btitle{Orthogonal Polynomials of Several Variables},
\bedition{2nd} ed.
\bpublisher{Cambridge Univ. Press}, \baddress{Cambridge}.
\end{bbook}
\endbibitem

\bibitem[\protect\citeauthoryear{E, Ma and Wu}{2020}]{e2020comparative}
\begin{barticle}[author]
\bauthor{\bsnm{E},~\bfnm{Weinan}\binits{W.}},
  \bauthor{\bsnm{Ma},~\bfnm{Chao}\binits{C.}} \AND
  \bauthor{\bsnm{Wu},~\bfnm{Lei}\binits{L.}}
(\byear{2020}).
\btitle{A comparative analysis of optimization and generalization properties of
  two-layer neural network and random feature models under gradient descent
  dynamics}.
\bjournal{Sci. China Math.}
\bvolume{63}
\bpages{1235--1258}.
\end{barticle}
\endbibitem

\bibitem[\protect\citeauthoryear{Hayakawa and
  Suzuki}{2020}]{hayakawa2020minimax}
\begin{barticle}[author]
\bauthor{\bsnm{Hayakawa},~\bfnm{Satoshi}\binits{S.}} \AND
  \bauthor{\bsnm{Suzuki},~\bfnm{Taiji}\binits{T.}}
(\byear{2020}).
\btitle{On the minimax optimality and superiority of deep neural network
  learning over sparse parameter spaces}.
\bjournal{Neural Netw.}
\bvolume{123}
\bpages{343--361}.
\end{barticle}
\endbibitem

\bibitem[\protect\citeauthoryear{Ongie et~al.}{2020}]{ongie2020function}
\begin{binproceedings}[author]
\bauthor{\bsnm{Ongie},~\bfnm{Greg}\binits{G.}},
  \bauthor{\bsnm{Willett},~\bfnm{Rebecca}\binits{R.}},
  \bauthor{\bsnm{Soudry},~\bfnm{Daniel}\binits{D.}} \AND
  \bauthor{\bsnm{Srebro},~\bfnm{Nathan}\binits{N.}}
(\byear{2020}).
\btitle{A function space view of bounded norm infinite width {ReLU} nets: The
  multivariate case}.
In \bbooktitle{International Conference on Learning Representations}.
\end{binproceedings}
\endbibitem

\bibitem[\protect\citeauthoryear{Shalev-Shwartz and
  Ben-David}{2014}]{shalev2014understanding}
\begin{bbook}[author]
\bauthor{\bsnm{Shalev-Shwartz},~\bfnm{Shai}\binits{S.}} \AND
  \bauthor{\bsnm{Ben-David},~\bfnm{Shai}\binits{S.}}
(\byear{2014}).
\btitle{Understanding Machine Learning: From Theory to Algorithms}.
\bpublisher{Cambridge Univ. Press}, \baddress{New York}.
\end{bbook}
\endbibitem

\bibitem[\protect\citeauthoryear{van~der Vaart and Wellner}{1996}]{van1996weak}
\begin{bbook}[author]
\bauthor{\bparticle{van~der} \bsnm{Vaart},~\bfnm{Aad~W.}\binits{A.~W.}} \AND
  \bauthor{\bsnm{Wellner},~\bfnm{Jon~A.}\binits{J.~A.}}
(\byear{1996}).
\btitle{Weak Convergence and Empirical Processes: With Applications to
  Statistics}.
\bpublisher{Springer}, \baddress{New York}.
\end{bbook}
\endbibitem

\bibitem[\protect\citeauthoryear{Wainwright}{2019}]{wainwright2019high}
\begin{bbook}[author]
\bauthor{\bsnm{Wainwright},~\bfnm{M.~J.}\binits{M.~J.}}
(\byear{2019}).
\btitle{High-Dimensional Statistics: A Non-Asymptotic Viewpoint}.
\bpublisher{Cambridge Univ. Press}, \baddress{Cambridge}.
\end{bbook}
\endbibitem

\bibitem[\protect\citeauthoryear{Xu}{1998}]{xu1998orthogonal}
\begin{barticle}[author]
\bauthor{\bsnm{Xu},~\bfnm{Yuan}\binits{Y.}}
(\byear{1998}).
\btitle{Orthogonal polynomials and cubature formulae on spheres and on balls}.
\bjournal{SIAM J. Math. Anal.}
\bvolume{29}
\bpages{779--793}.
\end{barticle}
\endbibitem

\bibitem[\protect\citeauthoryear{Yang and Barron}{1999}]{yang1999information}
\begin{barticle}[author]
\bauthor{\bsnm{Yang},~\bfnm{Yuhong}\binits{Y.}} \AND
  \bauthor{\bsnm{Barron},~\bfnm{Andrew}\binits{A.}}
(\byear{1999}).
\btitle{Information-theoretic determination of minimax rates of convergence}.
\bjournal{Ann. Statist.}
\bvolume{27}
\bpages{1564--1599}.
\end{barticle}
\endbibitem

\end{thebibliography}

\end{document}


\begin{frontmatter}
\title{Supplement to ``Nonasymptotic theory for two-layer neural networks: Beyond the bias--variance trade-off''}
\runtitle{Supplementary Material}

\begin{aug}
\author{\fnms{Huiyuan}~\snm{Wang}\ead[label=e1]{huiyuan.wang@pku.edu.cn}}
\and
\author{\fnms{Wei}~\snm{Lin}\ead[label=e2]{weilin@math.pku.edu.cn}}

\address{School of Mathematical Sciences and Center for Statistical Science, Peking University\printead[presep={,\ }]{e1,e2}}
\end{aug}

\end{frontmatter}
This Supplementary Material contains the remaining proofs and technical details. To complement Appendix \ref{sec:proof_over}, the proofs of risk bounds in expectation under the overparametrized regime are presented in Section \ref{sec:exp_bound}. Section \ref{sec:proof_under_regime} includes the proofs of results in the underparametrized regime. Section \ref{sec:target_function} contains mathematical details of the target function class and the proof of Theorem \ref{thm:approx}. Proofs of lower bounds are provided in Section \ref{sec:lowerboundsupplementray}. Section \ref{sec:tech_lemmas} collects some technical lemmas that are needed to prove the main results.

\section{Proofs of risk bounds in expectation}\label{sec:exp_bound}
In this section we present the proofs of \eqref{eq:emp_err2} in Theorem \ref{thm:emp_err} and \eqref{eq:pred_err2} in Theorem \ref{thm:pred_err}.

\begin{proof}[Proof of \eqref{eq:emp_err2} in Theorem \ref{thm:emp_err}]
Define the event $E_1=\{\lambda\ge2n^{-1/2}\max_{j}\|\bv_j\|_2\}$ with $\lambda=8\sigma_\ve\sqrt{(4+d)\log n/n}$. It follows from \eqref{eq:large_overall} that
\[
\E\{\|g(\cdot;\widehat\btheta)-f^*\|_n^2I(E_1)\}\le24\lambda\|f^*\|_\cS+2C\|f^*\|_\cS^2m^{-(d+3)/d}.
\]
It remains to bound $\E\{\|g(\cdot;\widehat\btheta)-f^*\|_n^2I(E_1^c)\}$. Recall the definition of $T_3$ in \eqref{eq:small_m_basic}. By the Cauchy--Schwarz inequality, we have
\begin{align*}
T_3&\le\biggl(\frac{1}{n}\sum_{i=1}^n\ve_i^2\biggr)^{1/2}\biggl\{\frac{1}{n}\sum_{i=1}^n\big(g(\bx_i;\widehat\btheta)-g(\bx_i;\btheta^*)\big)^2\biggr\}^{1/2}\\
&\le\biggl(\frac{1}{n}\sum_{i=1}^n\ve_i^2\biggr)^{1/2}(\|g(\cdot;\widehat\btheta)-f^*\|_n+\|g(\cdot;\btheta^*)-f^*\|_n)\\
&\le\frac{2}{n}\sum_{i=1}^n\ve_i^2+\frac{1}{4}\|g(\cdot;\widehat\btheta)-f^*\|_n^2+\frac{1}{4}\|g(\cdot;\btheta^*)-f^*\|_n^2.
\end{align*}
Rearranging terms, \eqref{eq:small_m_basic} becomes
\begin{equation*}
\frac{1}{4}\|g(\cdot;\widehat\btheta)-f^*(\cdot)\|_n^2\le\lambda\nu(\btheta^*)+\frac{3}{4}\|g(\cdot;\btheta^*)-f^*(\cdot)\|_n^2+\frac{2}{n}\sum_{i=1}^n\ve_i^2.
\end{equation*}
Taking expectation gives
\begin{align*}
&\E\{\|g(\cdot;\widehat\btheta)-f^*(\cdot)\|_n^2I(E_1^c)\}\\
&\quad\le\{4\lambda \nu(\btheta^*)+3\|g(\cdot;\btheta^*)-f^*(\cdot)\|_n^2\}P(E_1^c)+\frac{8}{n}\E\biggl\{\sum_{i=1}^n\ve_i^2I(E_1^c)\biggr\}\\
&\quad\equiv R_1+R_2.
\end{align*}
It follows from \eqref{eq:concentration_xi} that $P(E_1^c)\le4n^{-4}$. Setting $\lambda=8\sigma_\ve\sqrt{(4+d)\log n/n}$ and applying Theorem \ref{thm:approx} yields
\[
R_1\le C_1\biggl(\|f^*\|_\cS\sigma_\ve\sqrt{\frac{d\log n}{n}}+\|f^*\|_\cS^2m^{-(d+3)/d}\biggr)n^{-4}
\]
for some constant $C_1>0$. By the Cauchy--Schwarz inequality, we have $R_2\le8(\E\ve^4)^{1/2}n^{-2}$. We then conclude that $\E\{\|g(\cdot;\widehat\btheta)-f^*\|_n^2I(E_1^c)\}$ is of smaller order than $\E\{\|g(\cdot;\widehat\btheta)-f^*\|_n^2I(E_1)\}$, and consequently
\[
\E\|g(\cdot;\widehat\btheta)-f^*\|_n^2\le C_2\biggl\{\|f^*\|_\cS^2m^{-(d+3)/d}+(\sigma^2_\ve+\|f^*\|_\cS^2)\sqrt{\frac{d\log n}{n}}\biggr\}
\]
for some constant $C_2>0$.
\end{proof}

\begin{proof}[Proof of \eqref{eq:pred_err1} in Theorem \ref{thm:pred_err}]
Note that $\E\|\widehat{f}-f^*\|_2^2=R_1+R_2$, where
\begin{equation*}
R_1=\E\{\|\widehat{f}-f^*\|_2^2I(E_1)\},\quad R_2=\E\{\|\widehat{f}-f^*\|_2^2I(E_1^c)\}.
\end{equation*}
Under the event $E_1$, we have proved in \eqref{eq:large_norm_bound} that $\nu(\widehat\btheta)\le32\|f^*\|_\cS$; that is, $I(E_1)\le I(\nu(\widehat\btheta)\le32\|f^*\|_\cS)$. It follows from \eqref{eq:Delta_in_class}, Lemma \ref{lem:talagrand}, and Theorem \ref{thm:emp_err} that
\begin{align}
\begin{split}\label{eq:epi}
R_1&\le\E\{\|\widehat{f}-f^*\|_2^2I(\nu(\widehat\btheta)\le20\|f^*\|_\cS)\}\\
&\le\E\|\widehat{f}-f^*\|_n^2+1024\|f^*\|_\cS^2\E\sup_{f\in\cF^*(m,1)}\left|\|f\|_n^2-\|f\|_2^2\right|\\
&\le\E\|\widehat{f}-f^*\|_n^2+\frac{1024}{\sqrt{n}}C_\cF\|f^*\|_\cS^2\\
&\le C_2\biggl\{(\sigma_\ve^2+\|f^*\|_\cS^2)\sqrt{\frac{d\log n}{n}}+\|f^*\|_\cS^2m^{-(d+3)/d}\biggr\}
\end{split}
\end{align}
for some constant $C_2>0$.

It remains to bound $R_2$. By \eqref{eq:small_m_basic}, the Cauchy--Schwarz inequality, and the assumption on $m$ in \eqref{eq:critic_m}, we obtain
\begin{equation}\label{eq:bound_nu_esti}
\begin{split}
\nu(\widehat\btheta)&\le\|f^*\|_\cS+(2n\lambda)^{-1}\sum_{i=1}^n\bigl(g(\bx_i;\btheta^*)-y_i\bigr)^2\\
&\le\|f^*\|_\cS+\lambda^{-1}\|g(\cdot;\btheta^*)-f^*\|_n^2+(n\lambda)^{-1}\sum_{i=1}^n\ve_i^2\\
&\le2\|f^*\|_\cS+(n\lambda)^{-1}\sum_{i=1}^n\ve_i^2.
\end{split}
\end{equation}
Note that $|\widehat{f}(\bx)|\le\nu(\widehat\btheta)$ for any $\bx\in\B^d$. Thus,
\begin{equation}\label{eq:eps1}
R_2\le\E\biggl\{\sup_{\bx\in\B^d}2(|\widehat{f}(\bx)|^2+|f^*(\bx)|^2)I(E_1^c)\biggr\}\le\E\{(2\nu^2(\widehat\btheta)+2\|f^*\|_\cS^2)I(E_1^c)\}.
\end{equation}
By the Cauchy--Schwarz inequality and the fact that $P(E_1^c)\le4n^{-4}$, we further obtain
\begin{equation}\label{eq:eps2}
\begin{split}
\E\{(2\nu^2(\widehat\btheta)+2\|f^*\|_\cS^2)I(E_1^c)\}&\le2\E^{1/2}\{(\nu^2(\widehat\btheta)+|f^*\|_\cS^2)^2\}P^{1/2}(E_1^c)\\
&\le4n^{-2}\E^{1/2}\{(\nu^2(\widehat\btheta)+\|f^*\|_\cS^2)^2\}\\
&\le4n^{-2}\{2\E\nu^4(\widehat\btheta)+2\|f^*\|_\cS^4\}^{1/2}.
\end{split}
\end{equation}
Repeatedly using $(a+b)^2\le2a^2+2b^2$ in \eqref{eq:bound_nu_esti} gives
\begin{equation}\label{eq:nu4}
\nu^4(\widehat\btheta)\le\biggl\{8\|f^*\|_\cS^2+2(n\lambda)^{-2}\biggl(\sum_{i=1}^n\ve_i^2\biggr)^2\biggr\}^2\le128\|f^*\|_\cS^4+8(n\lambda)^{-4}\biggl(\sum_{i=1}^n\ve_i^2\biggr)^4.
\end{equation}
Since $\ve_i$ are independent Gaussian variables, we have $\E(n^{-1}\sum_{i=1}^n\ve_i^2)^4<\infty$. Substituting \eqref{eq:eps2} and \eqref{eq:nu4} into \eqref{eq:eps1} yields
\begin{equation}\label{expectation_second_part3}
R_2\le C_3n^{-2}\lambda^{-2}
\end{equation}
for some constant $C_3>0$. Note that $\lambda=O(\sqrt{d\log n/n})$. For $d\ge1$, $n^{-2}\lambda^{-3}=o(n^{-1/2})$. Combining \eqref{eq:epi} and \eqref{expectation_second_part3}, we conclude that
\[
\E\|g(\cdot;\widehat{\btheta})-f^*\|_2^2\le C_4\biggl\{\|f^*\|_\cS^2m^{-(d+3)/d}+(\sigma_\ve^2+\|f^*\|_\cS^2)\sqrt{\frac{d\log n}{n}}\biggr\}
\]
for some constant $C_4>0$.
\end{proof}

\section{Proofs of results in the underparametrized regime}\label{sec:proof_under_regime}
In this section we present the proof of Theorem \ref{thm:underparam}. Let $\cN_m(\delta)\equiv\cN(\delta,\cF(m,1),\|\cdot\|_n)$ be the $\delta$-covering number of $\cF(m,1)$ with respect to $\|\cdot\|_n$, and define $\cF^*(m,1)=\{f-f^*\colon \|f^*\|_\cS\le 1, f\in \cF(m,1)\}$.

We first bound the empirical error of the regularized network estimator in the underparametrized regime. To deal with scaled variation regularization, we need to analyze the supremum of an empirical process
\[
V_\delta(\bve)\equiv\sup_{f\in\cF(m,1)}\biggl\{\frac{n^{-1}|\sum_{i=1}^n\ve_if(\bx_i)|-\delta\sqrt{n^{-1}\sum_{i=1}^n\ve_i^2}}{\|f\|_n+\delta}\biggr\},
\]
where $\bve=(\ve_1,\dots,\ve_n)^T$.

\begin{lemma}\label{lem:variance_maxima}
The function $V_{\delta}(\cdot)\colon\R^n\to\R$ is $n^{-1/2}$-Lipschitz continuous, and
\[
\E V_\delta(\bve)\le2\sigma_\ve\sqrt{\frac{\log\cN_m(\delta)}{n}}.
\]
\end{lemma}

\begin{proof}
We first show that $V_{\delta}(\bve)$ is Lipschitz continuous with respect to the Euclidean norm. For any two vectors $\bve^{(1)}=(\ve_1^{(1)},\dots,\ve_n^{(1)})^T$ and $\bve^{(2)}=(\ve_1^{(2)},\dots,\ve_n^{(2)})^T$, the inequality $|\sup_{a\in\cA}F(a)-\sup_{a\in\cA}G(a)|\le\sup_{a\in\cA}|F(a)-G(a)|$ implies that
\begin{align*}
&|V_\delta(\bve^{(1)})-V_\delta(\bve^{(2)})|\\
&\quad\le\sup_{f\in\cF(m,1)}\biggl\{\frac{|n^{-1}\sum_{i=1}^n(\ve_i^{(1)}-\ve_i^{(2)})f(\bx_i)|+\delta n^{-1/2}\|\bve^{(1)}-\bve^{(2)}\|_2}{\|f\|_n+\delta}\biggr\}\\
&\quad\le\sup_{f\in\cF(m,1)}\biggl\{\frac{n^{-1/2}\|\bve^{(1)}-\bve^{(2)}\|_2\|f\|_n+\delta n^{-1/2}\|\bve^{(1)}-\bve^{(2)}\|_2}{\|f\|_n+\delta}\biggr\}=\frac{1}{\sqrt{n}}\|\bve^{(1)}-\bve^{(2)}\|_2,
\end{align*}
which gives the desired result.

Next, we bound $\E V_\delta(\bve)$. Let the minimal $\delta$-covering of $\cF(m,1)$ with respect to $\|\cdot\|_n$ be $\{f_j(\cdot):j=1,\dots,\cN_m(\delta)\}$. By definition, for any $f\in\cF(m,1)$, there exists some $j^*$ such that $\|f_{j^*}-f\|_n\le \delta$. By the triangle inequality, we obtain
\begin{align*}
\biggl|\sum_{i=1}^n\ve_if(\bx_i)\biggr|&\le\biggl|\sum_{i=1}^n\ve_if_{j^*}(\bx_i)\biggr|+\biggl|\sum_{i=1}^n\ve_i\bigl(f(\bx_i)-f_{j^*}(\bx_i)\bigr)\biggr|\\
&\le\biggl|\sum_{i=1}^n\ve_i\frac{f_{j^*}(\bx_i)}{\|f_{j^*}\|_n}\biggr|\|f_{j^*}\|_n+\delta\biggl(n\sum_{i=1}^n\ve_i^2\biggr)^{1/2}\\
&\le\max_{1\le j\le\cN_m(\delta)}\biggl|\sum_{i=1}^n\ve_i\frac{f_j(\bx_i)}{\|f_{j}\|_n}\biggr|(\|f\|_n+\delta)+\delta\biggl(n\sum_{i=1}^n\ve_i^2\biggr)^{1/2}.
\end{align*}
After some algebra, we have
\[
(\|f\|_n+\delta)^{-1}\biggl\{\biggl|\sum_{i=1}^n\ve_if(\bx_i)\biggr|-\delta\biggl(n\sum_{i=1}^n\ve_i^2\biggr)^{1/2}\biggr\}\le\max_{j}\biggl|\sum_{i=1}^n\ve_i\frac{f_j(\bx_i)}{\|f_j\|_n}\biggr|
\]
for all $f\in\cF(m,1)$, or equivalently
\[
V_\delta(\bve)\le\frac{1}{\sqrt{n}}\max_j\frac{1}{\sqrt{n}}\biggl|\sum_{i=1}^n\ve_i\frac{f_j(\bx_i)}{\|f_j\|_n}\biggr|.
\]
Since $\ve_1,\dots,\ve_n$ are normally distributed with variance $\sigma_\ve^2$, $n^{-1/2}\sum_{i=1}^n\ve_i f_{j}(\bx_i)/\|f_{j}\|_n$ are also normally distributed with $\sigma_\ve^2$. It follows from Lemma \ref{lem:maxima_Gauss} that
\[
\E V_{\delta}(\bve)\le2\sigma_\ve\sqrt{\frac{\log\cN_m(\delta)}{n}},
\]
completing the proof.
\end{proof}

Using Lemma \ref{lem:variance_maxima}, we can obtain bounds for $\|g(\cdot;\widehat{\btheta})-f^*\|_n^2$ in the underparametrized regime.

\begin{theorem}\label{thm:small_width}
Let $\delta_n=n^{-1}md\log n<1$. Under Conditions \eqref{cond:f}--\eqref{cond:eps}, the regularized network estimator $g(\cdot;\widehat\btheta)$ with $\lambda=C_1\sigma_\ve\max(\delta_n,m^{-(d+3)/d})$ satisfies
\begin{equation}\label{eq:emp_err_under1}
\|g(\cdot;\widehat{\btheta})-f^*\|_n^2\le C\biggl\{\|f^*\|_{{\cS}}^2m^{-(d+3)/d}+(\sigma_\ve^2+\|f^*\|_\cS^2)\frac{md\log n}{n}\biggr\}
\end{equation}
and $\nu(\widehat{\btheta})\le C_\kappa$ with probability at least $1-O(n^{-C_2})$ for some constants $C_1,C_2,C,C_\kappa>0$.
\end{theorem}

\begin{proof}
Similar to the proof of Theorem \ref{thm:emp_err}, we bound $T_1,T_2$, and $T_3$ in \eqref{eq:small_m_basic}. Denote $\Delta^*(\bx)=g(\bx;\widehat\btheta)-g(\bx;\btheta^*)$. Note that $\Delta^*(\cdot)$ is also a two-layer ReLU network with at most $2m$ hidden units. With a slight abuse of notation, we write $\nu(\Delta^*)=\|\widehat\bB-\bB^*\|_{2,1}$ for the scaled variation norm of $\Delta^*$. From \eqref{eq:large_diff_norm} and \eqref{eq:large_approx}, we have
\begin{equation}\label{eq:small_diff_norm}
T_1\le2\lambda\nu(\btheta^*)-\lambda\nu(\Delta^*),\quad T_2\le C_1\|f^*\|_\cS^2 m^{-(d+3)/d}
\end{equation}
for some constant $C_1>0$. Define $\widehat\sigma_\ve=\sqrt{n^{-1}\sum_{i=1}^n\ve_i^2}$. For $T_3$, since $\Delta^*/\nu(\Delta^*)\in\cF(2m,1)$, we obtain
\[
\frac{T_3-\delta_n\nu(\Delta^*)\widehat\sigma_\ve}{\|\Delta^*\|_n+\delta_n\nu(\Delta^*)}=\frac{n^{-1}|\sum_{i=1}^n\ve_i\Delta^*(\bx_i)/\nu(\Delta^*)|-\delta_n\widehat\sigma_\ve}{\|\Delta^*/\nu(\Delta^*)\|_n+\delta_n}\le V_{\delta_n}(\bve).
\]
Noting that $V_{\delta_n}(\bve)$ is a Lipschitz continuous function of independent Gaussian variables and applying Theorem 2.26 of \citet{wainwright2019high} yields
\[
P(|V_{\delta_n}(\bve)-\E V_{\delta_n}(\bve)|\ge t)\le2\exp\biggl(-\frac{nt^2}{2}\biggr).
\]
By Lemma \ref{lem:variance_maxima}, $\E V_{\delta_n}(\bve)\le2\sigma_\ve\sqrt{\log\cN_{2m}(\delta_n)/n}$. Choosing $t=2\sigma_\ve\sqrt{\log\widetilde{p}/n}$ for some $\widetilde{p}\ge\cN_{2m}(\delta_n)$ to be specified later, we have, with probability at least $1-2\widetilde{p}^{-2\sigma_\ve^2}$,
\[
V_{\delta_n}(\bve)<4\sigma_\ve\sqrt{\frac{\log\widetilde{p}}{n}}.
\]
Similarly, $P(\widehat\sigma_\ve\ge\sigma_\ve+t)\le\exp(-nt^2/2)$ since $n^{-1/2}\|\bve\|_2$ is also $n^{-1/2}$-Lipschitz continuous and $n^{-1/2}\E\|\bve\|_2\le\sqrt{n^{-1}\E\bve^T\bve}=\sigma_\ve$. Choosing $t=\sigma_\ve$, we have, with probability at least $1-\exp(-n\sigma_\ve^2/2)$,
\[
\widehat\sigma_\ve<2\sigma_\ve.
\]
Combining these pieces gives
\begin{equation}\label{eq:small_variance}
T_3\le4\sigma_\ve\sqrt{\frac{\log\widetilde{p}}{n}}\big(\|\Delta^*\|_n+\delta_n\nu(\Delta^*)\big)+2\sigma_\ve\delta_n\nu(\Delta^*)
\end{equation}
with probability at least $1-2\widetilde{p}^{-2\sigma_\ve^2}-\exp(-\sigma_\ve^2n/2)$. Further combining \eqref{eq:small_m_basic}, \eqref{eq:small_diff_norm}, and \eqref{eq:small_variance} yields
\begin{equation}\label{eq:control_small_norm}
\begin{split}
\frac{1}{2}\|g(\cdot;\widehat\btheta)-f^*\|_n^2&\le C_1\|f^*\|_\cS^2m^{-(d+3)/d}+4\sigma_\ve\sqrt{\frac{\log\widetilde{p}}{n}}\|\Delta^*\|_n\\
&\relph{}+\biggl\{\biggl(2\sqrt{\frac{\log\widetilde{p}}{n}}+1\biggr)2\sigma_\ve\delta_n-\lambda\biggr\}\nu(\Delta^*)+2\lambda\nu(\btheta^*).
\end{split}
\end{equation}
Choosing $\lambda\ge(2\sqrt{n^{-1}\log\widetilde{p}}+1)4\sigma_\ve\delta_n$, we have
\begin{equation}\label{eq:sbc}
\begin{split}
\frac{1}{2}\|g(\cdot;\widehat\btheta)-f^*\|_n^2&\le C_1\|f^*\|_\cS^2m^{-(d+3)/d}+2\lambda \nu(\btheta^*)\\
&\mathrel{\phantom{\le}}{}+4\sigma_\ve\sqrt{\frac{\log\widetilde{p}}{n}}(\|g(\cdot;\widehat\btheta)-f^*\|_n+\|g(\cdot;\btheta^*)-f^*\|_n),
\end{split}
\end{equation}
where we have used the triangle inequality to bound $\|\Delta^*\|_n$. Using the inequality $ab\le a^2+b^2/4$, we obtain
\begin{align}
4\sigma_\ve\sqrt{\frac{\log\widetilde{p}}{n}}\|g(\cdot;\widehat\btheta)-f^*\|_n&\le16\sigma_\ve^2\frac{\log\widetilde{p}}{n}+\frac{1}{4}\|g(\cdot;\widehat\btheta)-f^*\|_n^2,\label{eq:cf}\\
4\sigma_\ve\sqrt{\frac{\log\widetilde{p}}{n}}\|g(\cdot;\btheta^*)-f^*\|_n&\le16\sigma_\ve^2\frac{\log\widetilde{p}}{n}+\frac{1}{4}\|g(\cdot;\btheta^*)-f^*\|_n^2.\label{eq:cs}
\end{align}
Substituting \eqref{eq:cf} and \eqref{eq:cs} into \eqref{eq:sbc} and noting that $\nu(\btheta^*)\le6\|f^*\|_\cS$ yields
\[
\|g(\cdot;\widehat\btheta)-f^*\|_n^2\le6C_1\|f^*\|_\cS^2m^{-(d+3)/d}+128\sigma_\ve^2\frac{\log\widetilde{p}}{n}+48\|f^*\|_\cS\lambda
\]
with probability at least $1-2\widetilde{p}^{-2\sigma_\ve^2}-\exp(-\sigma_\ve^2n/2)$.

It remains to bound $\cN_{2m}(\delta_n)\equiv\cN(\delta_n,\cF(2m,1),\|\cdot\|_n)$. Since a $\delta_n$-covering of $\cF(2m,1)$ with respect to $\|\cdot\|_\infty$ is always a $\delta_n$-covering with respect to $\|\cdot\|_n$, by Lemma \ref{lem:metric_entropy} we have
\[
\log\cN_{2m}(\delta_n)\le\log\cN(\delta_n,\cF(2m,1),\|\cdot\|_\infty)\le4md\log(1+4\delta_n^{-1}).
\]
Recall that $\delta_n=n^{-1}md\log n$, and we have
\begin{equation}\label{eq:delta_bound}
\log(1+4\delta_n^{-1})\le\log\Bigl(1+\frac{4n}{md}\Bigr)\le2\log n.
\end{equation}
Now take $\widetilde{p}=n^{8md}\ge\cN_{2m}(\delta_n)$, and by the assumption that $\delta_n\le1$ we have
\begin{equation}\label{eq:lambda_ineq}
\sqrt{\frac{\log\widetilde{p}}{n}}\le\sqrt{\frac{9md\log n}{n}}\le3.
\end{equation}
In order for $\lambda\ge(2\sqrt{\log\widetilde{p}/n}+1)4\sigma_\ve\delta_n$ to hold, setting $\lambda=28\sigma_\ve\max(\delta_n,m^{-(d+3)/d})$ is sufficient. With this choice of $\lambda$, we conclude that \eqref{eq:emp_err_under1} holds with probability at least  $1-2\widetilde{p}^{-2\sigma_\ve^2}-\exp(-\sigma_\ve^2n/2)$. This completes the proof of \eqref{eq:emp_err_under1} by noting that
\[
-2\sigma_\ve^2\log\widetilde{p}=-16\sigma_\ve^2md\log n\le-16\sigma_\ve^2\log n
\]
and $\exp(-\sigma_\ve^2n/2)=o(n^{-C_2})$ for any constant $C_2>0$.

Next, we prove that $\nu(\widehat{\btheta})\le C_\kappa$. For $\lambda=28\sigma_\ve\max(\delta_n,m^{-(d+3)/d})$, it follows from \eqref{eq:lambda_ineq} that $\lambda-(4\sigma_\ve\sqrt{n^{-1}\log\widetilde{p}}+2\sigma_\ve)\delta_n\ge\lambda/2$. Then \eqref{eq:control_small_norm} becomes
\begin{equation}\label{eq:csn2}
\begin{split}
&\lambda\nu(\Delta^*)+\|g(\cdot;\widehat\btheta)-f^*\|_n^2\\
&\quad\le2C_1\|f^*\|_\cS^2m^{-(d+3)/d}+8\sigma_\ve\sqrt{\frac{\log\widetilde{p}}{n}}\|\Delta^*\|_n+4\lambda\nu(\btheta^*).
\end{split}
\end{equation}
Combining \eqref{eq:csn2}, \eqref{eq:cf}, and \eqref{eq:cs}, and using $\nu(\btheta^*)\le6\|f^*\|_\cS$, we obtain
\[
\lambda\nu(\Delta^*)\le2C_1\|f^*\|_\cS^2m^{-(d+3)/d}+8\sigma_\ve^2\frac{\log\widetilde{p}}{n}+24\|f^*\|_\cS\lambda.
\]
Dividing both sides by $\lambda$ and noting that $28\sigma_\ve m^{-(d+3)/d}\lambda^{-1}\le1$ yields
\[
\nu(\Delta^*)\le\biggl(2C_1\sqrt{\frac{\kappa}{1-\kappa}}+24\biggr)\|f^*\|_\cS+8\sqrt{\frac{1-\kappa}{\kappa}}\sigma_\ve,
\]
and hence
\[
\nu(\widehat{\btheta})\le\nu(\Delta^*)+6\|f^*\|_\cS\equiv C_\kappa,
\]
where $\kappa=\|f^*\|_\cS^2/(\sigma_\ve^2+\|f^*\|_\cS^2)$.
\end{proof}

We present a lemma useful for proving Theorem \ref{thm:underparam}, whose proof can be found in Section \ref{sec:upper_lemma_proof}.

\begin{lemma}\label{lem:maximal_small_width}
For any $0<\gamma<1$, define by $\cB_\cF(\gamma)=\{f\in\cF^*(m,1)\colon\|f\|_2<\gamma\}$ the $L_2(\mu)$-ball in $\cF^*(m,1)$ of radius smaller than $\gamma$. Let $Z_n(\gamma)=\sup_{f\in\cB_\cF(\gamma)}\bigl|\|f\|_n^2-\|f\|_2^2\bigr|$. Then, for any $\sqrt{6md\log n/n}\le\gamma\le1$,
\[
\E Z_n(\gamma)\le480\gamma\sqrt{\frac{md\log n}{n}}.
\]
\end{lemma}

We are now in a position to prove Theorem \ref{thm:underparam}.

\begin{proof}[Proof of Theorem \ref{thm:underparam}]
Let $\widehat\Delta(\cdot)=g(\cdot;\widehat\btheta)-f^*(\cdot)$, $\Delta^*(\cdot)=g(\cdot;\widehat\btheta)-g(\cdot;\btheta^*)$, and $\gamma_n=\sqrt{6md\log n/n}$. Define the events $E_2(\gamma)=\{Z_n(\gamma)\ge C_{E_2}\gamma^2\}$, $E_3=\{Z_n(\|f\|_2)\ge C_{E_3}\gamma_n\|f\|_2\text{ for some }f\in\cF^*(m,1)\text{ with }\|f\|_2\ge\gamma_n\}$, and
\[
E_4=\bigl\{\bigl|\|f\|_n-\|f\|_2\bigr|\le C_{E_4}\sqrt{\delta_n}\text{ for all }f\in\cF^*(m,1)\bigr\},
\]
where $C_{E_2},C_{E_3},C_{E_4}>0$ are universal constants to be specified later.

It follows from the proof of Theorem 14.1 in \citet{wainwright2019high} that $E_2^c(\gamma_n)\cap E_3^c\subset E_4$. In fact, if $f\in\cF^*(m,1)$ with $\|f\|_2\le\gamma_n$, then conditioning on $E_2^c(\gamma_n)$ we have $\|f\|_n\le\sqrt{C_{E_2}+1}\gamma_n$, in which case $E_4$ holds with $C_{E_4}\ge1+\sqrt{C_{E_2}+1}$. If $f\in\cF^*(m,1)$ with $\|f\|_2>\gamma_n$, then conditioning on $E_3^c$ we have
\[
\bigl|\|f\|_n-\|f\|_2\bigr|=\frac{\bigl|\|f\|_n^2-\|f\|_2^2\bigr|}{\|f\|_n+\|f\|_2}\le\frac{C_{E_3}\|f\|_2\gamma_n}{\|f\|_2}=C_{E_3}\gamma_n,
\]
and thus $E_4$ holds with $C_{E_4}\ge C_{E_3}$. We then choose $C_{E_4}=\max(1+\sqrt{C_{E_2}+1},C_{E_3})$.

By Theorem \ref{thm:small_width}, with probability at least $1-O(n^{-C_0})$ for some constant $C_0>0$ we have $\max(\nu(\widehat\btheta),\|f^*\|_\cS)\le C_\kappa$ and \eqref{eq:emp_err_under1} holds. Note that $\widehat\Delta/\max(\nu(\widehat\btheta),\|f^*\|_\cS)\in\cF^{*}(m,1)$. Further conditioning on $E_4$, we have
\[
\bigl|\|\widehat\Delta\|_n-\|\widehat\Delta\|_2\bigr|\le\max(\nu(\widehat\btheta),\|f^*\|_\cS)C_{E_4}\gamma_n\le C_\kappa C_{E_4}\gamma_n,
\]
which, together with \eqref{eq:emp_err_under1}, yields the desired error bound.

It remains to bound the probability of $E_4$, or those of $E_2(\gamma_n)$ and $E_3$. To this end, from Lemma \ref{lem:maximal_small_width} we have
\[
\E Z_n(\gamma)\le(480/\sqrt{6})\gamma_n\gamma<240\gamma_n\gamma
\]
for any $\gamma_n\le\gamma\le1$. Define
\[
U=\sup_{\bx\in\B^d}\sup_{f\in\cF^*(m,1)}|f(\bx)|^2,\quad\eta^2=\sup_{f\in\cB_\cF(\gamma)}\E|f(\bx)|^4,\quad K_n=2U\E Z_n(\gamma)+\eta^2,
\]
and note that
\[
U\le4,\quad\eta^2\le U\sup_{f\in\cB_\cF(\gamma)}\E|f(\bx)|^2\le4\gamma^2,\quad K_n\le1924\gamma^2.
\]
Applying Talagrand's inequality \citep[Theorem 3.27]{wainwright2019high} yields
\[
P\{Z_n(\gamma)-\E Z_n(\gamma)\ge t\}\le2\exp\biggl(-\frac{nt^2}{8eK_n+4Ut}\biggr).
\]
Choosing $t=\gamma_n^2$ and $\gamma=\gamma_n$, we have, for some universal constant $C_1>0$,
\[
P\{Z_n(\gamma_n)\le241\gamma_n^2\}\ge1-2\exp(-C_1n\delta_n).
\]
Thus, with probability at least $1-2n^{-C_1md}$, $E_2^c(\gamma_n)$ holds with $C_{E_2}=241$. Moreover, taking $t=s_n\gamma_n$ and $\gamma=s_n$ for any $\gamma_n\le s_n\le1$ gives
\begin{equation}\label{eq:any_gamma}
P\{Z_n(s_n)\le241s_n\gamma_n\}\ge1-2\exp(-C_2n\delta_n)
\end{equation}
for some universal constant $C_2>0$.

Owing to the randomness of $\widehat\Delta$, we cannot directly take a fixed $s_n=\|f\|_2$ to complete the proof. Instead, since we need only consider the case $\gamma_n<\|f\|_2\le1$ (otherwise the desired result follows immediately), we propose to cover the set $\{f\in\cF^*(m,1):\gamma_n\le\|f\|_2\le1\}$ using segments $\cL_q$, where $\cL_q=\{f\in\cF^*(m,1):2^{q-1}\gamma_n\le\|f\|_2\le2^q\gamma_n\}$. By \eqref{eq:delta_bound},
$\log(1/\gamma_n)\le\log n$, and hence we need $Q\equiv\log_2(1/\gamma_n)$ many segments with $Q\le2\log n$.

By the union bound, we have $P(E_3)\le\sum_qP(E_3\cap\cL_q)$. Conditioning on $E_3\cap\cL_q$, there exists a function $f$ with $2^{q-1}\gamma_m\le\|f\|_2\le2^q\gamma_m$ such that $\bigl|\|f\|_n-\|f\|_2\bigr|\ge482\gamma_n\|f\|_2\ge241\cdot2^q\gamma_n^2$. By \eqref{eq:any_gamma}, we have
\[
P(E_3)\le\sum_{q=1}^QP(E_3\cap\cL_q)\le\sum_{q=1}^QP\{Z_n(2^q\gamma_n)\ge241\cdot2^q\gamma_n^2\}\le2Q\exp(-C_2n\delta_n).
\]
Therefore, $E_3^c$ holds with $C_{E_3}=482$ and with probability at least $1-4(\log n)n^{-C_2md}$. Note that $E_5$ holds with probability $O(n^{-C_3})$. Combining these pieces, we conclude that
\[
\|\widehat\Delta\|_2^2\le C_4\biggl\{\|f^*\|_{{\cS}}^2m^{-(d+3)/d}+(\sigma_\ve^2+\|f^*\|_\cS^2)\frac{md\log n}{n}\biggr\}
\]
holds with probability at least $1-O(n^{-C_5})$ for some constants $C_4,C_5>0$.
\end{proof}

\section{Mathematical details of the target function space}\label{sec:target_function} In this section we provide the omitted proofs regarding our target function space and the proof of Theorem \ref{thm:approx}.
We first introduce the basic concepts of signed measures and total variation norm.

\subsection{Signed measures}
Given a measurable space $(D,\cB(D))$, that is, a set $D\subset\R^d$ and a $\sigma$-algebra $\cB(D)$ on $D$, a finite signed measure $\mu$ is a set function $\mu\colon\cB(D)\to\R$ such that $\mu(\emptyset)=0$ and $\mu$ is $\sigma$-additive. Denote by $\cM(D)$ the set of finite signed measures on $(D,\cB(D))$. The Jordan decomposition theorem states that any finite signed measure $\mu\in\cM(D)$ has a decomposition
\[
\mu=\mu_+-\mu_-,
\]
where $\mu_+$ and $\mu_-$ are mutually singular nonnegative measures. Then the total variation of $\mu$ is defined by $|\mu|=\mu_++\mu_-$, and the total variation norm defined by $|\mu|(D)=\int_Dd|\mu|$.

\subsection{Proofs for Section \ref{sec:prelim}}\label{sec:target_function_motiv}
In the subsection, we formally state and prove the properties of the target function space $\cG$ claimed in Section \ref{sec:prelim} and give the proof of Theorem \ref{thm:approx}.

For a two-layer ReLU network with parameter $\btheta$, denote by $C(\btheta)$ the squared $\ell_2$-norm of the network weights excluding the bias term, that is,
\[
C(\btheta)=\frac{1}{2}\sum_{k=1}^m(\|\bv_k\|_2^2+|a_k|^2).
\]
The \emph{universal approximation ability} of neural networks refers to the property that any continuous function $f\colon\R^d\to\R$ can be approximated arbitrarily well by neural networks $g(\cdot;\btheta)$ on a compact set $K\subset\R^d$ such that
\[
\sup_{\bx\in K}|f(\bx)-g(\bx;\btheta)|\le\ve
\]
for any $\ve>0$. Given any continuous function $f$, we are interested in the \emph{representational cost} $\overline{R}(f)$ for finite-width ReLU networks to approximate it, which, following \citet{ongie2020function}, is defined by
\[
\overline{R}(f)=\liminf_{\ve\to0}\biggl\{C(\btheta):\sup_{\|\bx\|\le\ve^{-1}}|g(\bx;\btheta)-f(\bx)|\le\ve\text{ and }g(\bzero;\btheta)=f(\bzero)\biggr\}.
\]
\citet{ongie2020function} proved that the representational cost $\overline{R}(f)$ is finite if and only if $f$ is an infinite-width two-layer ReLU network with skip connections, which is formally stated below.

\begin{lemma}[Lemma 10 and Theorem 2 of \cite{ongie2020function}]\label{lem:functionspace1}
Let $f$ be a Lipschitz function defined on $\R^d$. There exists a seminorm $\|\cdot\|_\cR$ such that $\|f\|_\cR<\infty$ if and only if
\[
f(\bx)=\int_{\mathbb{S}^{d-1}\times\R}\bigl(\sigma(\bv^T\bx+b)-\sigma(b)\bigr){d}\alpha_f(\bv,b)+\bbeta^T\bx+c
\]
for some unique even signed measure $\alpha_f\in\cM(\mathbb{S}^{d-1}\times\R)$, a unique vector $\bbeta\in\R^d$, and a unique constant $c$.
Moreover, $\|f\|_\cR$ is finite if and only if $\overline{R}(f)$ is finite, in which case $\|f\|_\cR=\int_{\mathbb{S}^{d-1}\times\R}{d}|\alpha_f|(\bv,b)$ and $\overline{R}(f)\ge\|f\|_\cR$.
\end{lemma}

\cite{ongie2020function} used $\|\cdot\|_\cR$ to characterize the representational cost $\overline{R}(\cdot)$, which is further linked to infinite-width two-layer ReLU networks. However, $\|\cdot\|_\cR$ is not a norm on the function class $\{f:\overline{R}(f)<\infty\}$. We then extend Lemma \ref{lem:functionspace1} to provide another equivalent characterization of the representational cost, under which the finiteness of $\overline{R}(\cdot)$ is related to a norm. We first define
\[
\cM_2(\R^{d+1})=\biggl\{\alpha\in\cM(\R^{d+1}):\int_{\R^{d+1}}\|\bv\|_2{d}|\alpha|(\bv,b)<\infty\biggr\}.
\]

\begin{proposition}\label{prop:functionspace2}
Let $f$ be a Lipschitz function defined on $\R^d$. Then $\|f\|_\cR<\infty$ if and only if
\begin{equation}\label{eq:Gf}
f(\bx)=\int_{\R^{d+1}}\bigl(\sigma(\bv^T\bx+b)-\sigma(b)\bigr)\,d\alpha_f(\bv,b)+c
\end{equation}
for some unique signed measure $\alpha_f\in\cM_2(\R^{d+1})$ and a unique constant $c$. Moreover, $\overline{R}(f)$ is finite if and only if $\int_{\R^{d+1}}\|\bv\|_2\,d|\alpha_f|(\bv,b)$ is finite.
\end{proposition}

\begin{proof}
We first show the necessity part. If $\|f\|_\cR<\infty$, then by Lemma \ref{lem:functionspace1}, there exists some unique even signed measure $\alpha_f\in\cM(\mathbb{S}^{d-1}\times\R)$, a unique $\bbeta\in\R^d$, and a unique constant $c$ such that
\[
f(\bx)=\int_{\mathbb{S}^{d-1}\times\R}\bigl(\sigma(\bv^T\bx+b)-\sigma(b)\bigr)\,d\alpha_f(\bv,b)+\bbeta_f^T\bx+c_f.
\]
Noting that $\bbeta^T\bx=\sigma(\bbeta^T\bx)-\sigma(-\bbeta^T\bx)$, there exists some signed measure
\[
\alpha_{\bbeta}(\bv,b)=I(\bbeta_f\ne\bzero_d)\|\bbeta_f\|_2\{I(\bv=\bbeta_f/\|\bbeta_f\|_2,b=0)-I(\bv=\bbeta_f/\|\bbeta_f\|_2,b=0)\}.
\]
In fact, one can verify that $\alpha_{\bbeta}\in\cM(\S^{d-1}\times\R)$, $\int_{\mathbb{S}^{d-1}\times\R}d|\alpha_{\bbeta}|=2\|\bbeta\|_2$, and $\bbeta^T\bx=\int_{\mathbb{S}^{d-1}\times\R}\bigl(\sigma(\bv^T\bx+b)-\sigma(b)\bigr)\,d\alpha_{\bbeta}(\bv,b)$. Noting that $\cM(\S^{d-1}\times\R)$ is a subspace of $\cM_2(\R^{d+1})$, we then have
\begin{align*}
f(\bx)&=\int_{\S^{d-1}\times\R}\bigl(\sigma(\bv^T\bx+b)-\sigma(b)\bigr)\,d(\alpha_f+\alpha_{\bbeta})(\bv,b)+c\\
&=\int_{\S^{d-1}\times\R}\bigl(\sigma(\bv^T\bx+b)-\sigma(b)\bigr)\,d\widetilde\alpha_f(\bv,b)+c\in\cM_2(\R^{d+1}).
\end{align*}
The uniqueness of $\widetilde\alpha_f$ is implied by the uniqueness of $\alpha_f$ and $\bbeta_f$, which completes the proof of the necessity part.

For the sufficiency part, consider the function $L(\bv,b)=(\bv^T/\|\bv\|_2,b/\|\bv\|_2)^TI(\bv\ne\bzero_d)$ that maps $\R^{d+1}\setminus\{\bv=\bzero_d\}$ to $\S^{d-1}\times\R$. The associated push-forward measure is defined by $\cL(\alpha)$ such that for all $A\in\cB(\S^{d-1}\times\R)$,
\[
\cL(\alpha)(A)=\int_{\R^{d+1}\setminus\{\bv=\bzero_d\}}I\{L(\bv,b)\in A\}\|\bv\|_2\,d\alpha(\bv,b),
\]
where $\cB(\S^{d-1}\times\R\})$ denotes the $\sigma$-algebra on $\S^{d-1}\times\R$. By the Hahn--Jordan decomposition, there exist a set $A^+$ on which $\cL(\alpha)$ is positive and $A^-$ on which $\cL(\alpha)$ is negative with $A^+\cup A^-=\S^{d-1}\times \R$. Then we obtain
\[
\int_{\S^{d-1}\times\R}d|\cL(\alpha)|(\bv,b)=\cL(\alpha)(A^+)-\cL(\alpha)(A^-)\le\int_{\R^{d+1}}\|\bv\|_2\,d|\alpha|(\bv,b).
\]
By the positive homogeneity of ReLU and the change-of-variable formula,
\begin{equation}\label{eq:lifting}
\begin{split}
&\int_{\R^{d+1}}\bigl(\sigma(\bv^T\bx+b)-\sigma(b)\bigr)\,d\alpha(\bv,b)\\
&\quad=\int_{\R^{d+1}\setminus\{\bv=\bzero_d\}}\bigl(\sigma(\bx^T\bv/\|\bv\|_2+b/\|\bv\|_2)-\sigma(b/\|\bv\|_2)\bigr)\|\bv\|_2\,d\alpha(\bv,b)\\
&\quad=\int_{\R^{d+1}\setminus\{\bv=\bzero_d\}}(\widetilde\sigma_\bx\circ L)(\bv,b)\|\bv\|_2\,d\alpha(\bv,b)\\
&\quad=\int_{\S^{d-1}\times\R}\widetilde\sigma_\bx(\bv,b)\,d\cL(\alpha)(\bv,b),
\end{split}
\end{equation}
where $\widetilde\sigma_\bx(\bv,b)=\sigma(\bv^T\bx+b)-\sigma(b)$. Note that we have the decomposition $\cL(\alpha)=\cL^+(\alpha)+\cL^-(\alpha)$, where $\cL^+(\alpha)$ is even and $\cL^-(\alpha)$ is odd. The identity $\widetilde\sigma_\bx-\widetilde\sigma_{-\bx}=\bx^T\bv$ gives
\begin{equation}\label{eq:linear_part}
\begin{split}
&\int_{\S^{d-1}\times\R}\widetilde\sigma_\bx(\bv,b)\,d\cL^-(\alpha)(\bv,b)\\
&\quad=\frac{1}{2}\int_{\S^{d-1}\times\R}\widetilde\sigma_\bx(\bv,b)\,d\cL^-(\alpha)(\bv,b)-\frac{1}{2}\int_{\S^{d-1}\times\R}\widetilde\sigma_\bx(-\bv,-b)\,d\cL^-(\alpha)(\bv,b)\\
&\quad=\frac{1}{2}\bx^T\int_{\S^{d-1}\times\R}\bv d\cL^-(\alpha)(\bv,b)\equiv\frac{1}{2}\bx^T\bbeta.
\end{split}
\end{equation}
Note that $\bbeta$ is well-defined, since
\begin{align*}
2\|\bbeta\|_2&\le\int_{\S^{d-1}\times\R}\|\bv\|_2d\cL^-(\alpha)(\bv,b)=\int_{\S^{d-1}\times\R}d\cL^-(\alpha)(\bv,b)\\
&\le\int_{\R^{d+1}}\|\bv\|_2\,d|\alpha|(\bv,b)<\infty.
\end{align*}
If there exists some $\alpha_f\in \cM_2(\R^{d+1})$ such that \eqref{eq:Gf} holds, then combining \eqref{eq:lifting} and \eqref{eq:linear_part} gives
\begin{align*}
f(\bx)&=\int_{\S^{d-1}\times\R}\widetilde\sigma_\bx(\bv,b)\,d\cL\{\alpha_f\}(\bv,b)+c\\
&=\int_{\S^{d-1}\times\R}\bigl(\sigma(\bx^T\bv+b)-\sigma(b)\bigr)\,d\cL^+\{\alpha_f\}(\bv,b)+\bbeta^T\bx+c
\end{align*}
with $\int_{\S^{d-1}\times\R}d|\cL^+\{\alpha_f\}|\le\int_{\R^{d+1}}\|\bv\|_2\,d|\alpha_f|(\bv,b)<\infty$. By Lemma \ref{lem:functionspace1}, we conclude that $\|f\|_\cR\le\overline{R}(f)<\infty$.

It is worth noting that $\overline{R}(f)<\infty$ implies that the function $f$ has a unique variational representation \eqref{eq:Gf} with $\int_{\R^{d+1}}\|\bv\|_2\,d|\alpha_f|(\bv,b)<\infty$. To complete the proof, it remains to prove $\overline{R}(f)\le\int_{\R^{d+1}}\|\bv\|_2\,d|\alpha_f|(\bv,b)$. By Lemma 11 of \citet{ongie2020function}, we have
\[
\nabla f(\infty)=\frac{1}{2}\int_{\S^{d-1}\times\R}\bv d\cL^-\{\alpha_f\}(\bv,b).
\]
Then by the definition of $\overline{R}(f)$, it follows from the proof of Theorem 2 in \citet{ongie2020function} that
\begin{align*}
\overline{R}(f)&=\min_{\widetilde{\alpha}^-\text{odd}}\|\cL^+\{\alpha_f\}+\widetilde{\alpha}^-\|_1\\
&\relph\text{such that }\int_{\S^{d-1}\times\R}\bv d\widetilde{\alpha}^-(\bv,b)=\int_{\S^{d-1}\times\R}\bv d\cL^-\{\alpha_f\}(\bv,b).
\end{align*}
Since $\cL^-\{\alpha_f\}$ is a feasible solution, we conclude that
\[
\overline{R}(f)\le\int d|\cL^+\{\alpha_f\}|+\int d|\cL^-\{\alpha_f\}|\le\int_{\R^{d+1}}\|\bv\|_2\,d|\alpha_f|(\bv,b).\qedhere
\]
\end{proof}

In the following proposition, we show that functions in $\cG(\B^d)$ have a cleaner integral representation.

\begin{corollary}\label{cor:restriction}
For any $f\in \cG(\B^d)$, there exists some signed measure $\widetilde\alpha$ on $\S^{d-1}\times[-1,1]$ such that
\[
f(\bx)=\int_{\S^{d-1}\times[-1,1]}\sigma(\bv^T\bx+b)\,d\widetilde{\alpha}(\bv,b)+c,\quad\bx\in\B^d
\]
with $\int_{\S^{d-1}\times[-1,1]}d|\widetilde{\alpha}|(\bv,b)\le3\|f\|_\cS<\infty$. Thus,
\[
\cG(\B^d)=\biggl\{\int_{\S^{d-1}\times[-1,1]}\sigma(\bv^T\bx+b)\,d\alpha(\bv,b):\int_{\S^{d-1}\times[-1,1]}d|\alpha|(\bv,b)<\infty\biggr\}.
\]
\end{corollary}

\begin{proof}
If we confine $\bx$ to the unit ball $\B^d$, then $\bv^T\bx\le\|\bv\|_2$ and when $b<-\|\bv\|_2$ we have $\sigma(\bv^T\bx+b)=0$. Thus,
\[
\sigma(\bv^T\bx+b)-\sigma(b)=\bigl(\sigma(\bv^T\bx+b)-\sigma(b)\bigr)I(|b|\le\|\bv\|_2)+\bv^T\bx I(b>\|\bv\|_2).
\]
By \eqref{eq:lifting}, we can rewrite \eqref{eq:Gf} as
\begin{equation}\label{eq:parhi_nowak}
\begin{split}
f(\bx)&=\int_{\S^{d-1}\times\R}\bigl(\sigma(\bx^T\bv+b)-\sigma(b)\bigr)\,d\cL(\alpha)(\bv,b)+c\\
&=\int_{\S^{d-1}\times\R}\sigma(\bv^T\bx+b)I(|b|\le\|\bv\|_2)\,d\cL(\alpha)(\bv,b)\\
&\relph{}-\int_{\S^{d-1}\times\R}\sigma(b)I(|b|\le\|\bv\|_2)\,d\cL(\alpha)(\bv,b)\\
&\relph{}+ \bx^{T}\int_{\S^{d-1}\times\R}\bv I(b>\|\bv\|_2)\,d\cL(\alpha)(\bv,b)+c\\
&=\int_{\S^{d-1}\times[-1,1]}\sigma(\bv^T\bx+b)\,d\alpha_1(\bv,b)+\bx^T\bbeta+c_f
\end{split}
\end{equation}
where $d\alpha_1(\bv,b)=I(|b|\le\|\bv\|_2)\,d\cL(\alpha)(\bv,b)$, $\bbeta=\int_{\S^{d-1}\times\R}\bv I(b>\|\bv\|_2)\,d\cL(\alpha)(\bv,b)$, and $c_f=c-\int_{\S^{d-1}\times\R}\sigma(b)I(|b|\le\|\bv\|_2)\,d\cL(\alpha)(\bv,b)$. Note that
\begin{align*}
\int_{\S^{d-1}\times[-1,1]}d|\alpha_1|(\bv,b)&\le\int_{\S^{d-1}\times\R}d|\cL(\alpha)|(\bv,b)\le\|f\|_\cS,\\
\|\bbeta\|_2&\le\int_{\S^{d-1}\times\R}\|\bv\|_2\,d|\cL(\alpha)|(\bv,b)\le\|f\|_\cS.
\end{align*}
Define
\[
\alpha_{\bbeta}(\bv,b)=I(\bbeta\ne\bzero_d)\|\bbeta\|_2\{I(\bv=\bbeta/\|\bbeta\|_2,b\in[-1,1])-I(\bv=\bbeta/\|\bbeta\|_2,b\in[-1,1])\},
\]
and hence $\int_{\S^{d-1}\times[-1,1]}d|\alpha_{\bbeta}|=2\|\bbeta\|_2\le2\|f\|_\cS$. Substituting $\bbeta^T\bx=\int_{\S^{d-1}\times[-1,1]}\sigma(\bv^T\bx+b)\,d\alpha_{\bbeta}(\bv,b)$ into \eqref{eq:parhi_nowak}, we obtain
\[
f(\bx)=\int_{\S^{d-1}\times[-1,1]}\sigma(\bv^T\bx+b)\,d\widetilde{\alpha}(\bv,b)+c_f,
\]
where $\widetilde\alpha=\alpha_1+\alpha_{\bbeta}$. By the triangle inequality, we have
\[
\int_{\S^{d-1}\times[-1,1]}d|\widetilde{\alpha}|(\bw)\le(1+2)\int_{\R^{d+1}}\|\bv\|_2\,d\alpha(\bv,b)=3\|f\|_{{\cS}},
\]
completing the proof.
\end{proof}

We are now ready to give the proof of Theorem \ref{thm:approx}.

\begin{proof}[Proof of Theorem \ref{thm:approx}]
For any $f\in\cG$ with $\|f\|_\cS\le M$, by Proposition \ref{cor:restriction}, there exists some signed measure $\alpha$ with $\int_{\S^{d-1}\times[-1,1]}d|\alpha|(\bv,b)\le3\|f\|_\cS$ and a constant $c_f<\infty$ such that
\[
f(\bx)=\int_{\S^{d-1}\times[-1,1]}\sigma(\bv^T\bx+b)\,d\alpha(\bv,b)+c_f.
\]
As a result, $f(\bx)\in\cF_1$ with $\gamma_1(f)\le\int_{\S^{d-1}\times[-1,1]}d|\alpha|(\bv,b)\le3\|f\|_\cS$, where $\cF_1$ and its equipped norm $\gamma_1$ are defined in Section 2.1 of \citet{bach2017breaking}.

Consider the function $N(\bv,b)=(\|\bv\|_2^2+b^2)^{-1/2}(\bv^T,b)^T$ that maps $\S^{d-1}\times[-1,1]$ to $\S^d$. Define the push-forward measure $N_\natural\alpha$ associated with $N(\bv,b)$ by
\[
N_\natural\alpha(A)=\int_{\S^{d-1}\times[-1,1]}I\{N(\bv,b)\in A\}\sqrt{\|\bv\|_2^2+b^2}\,d\alpha(\bv,b)
\]
for all $A\in\cB(\S^{d-1}\times[-1,1])$. The push-forward measure is well-defined since
\[
\int_{\S^d}d|N_\natural\alpha|(\bv,b)\le\int_{\S^{d-1}\times[-1,1]}\sqrt{\|\bv\|_2^2+b^2}\,d|\alpha|(\bv,b)\le\int_{\S^{d-1}\times[-1,1]}2\,d|\alpha|(\bv,b)\le6\|f\|_\cS.
\]
Thus, by the change-of-variable formula and the positive homogeneity of ReLU, we have
\begin{align*}
f(\bx)&=\int_{\S^{d-1}\times[-1,1]}\sigma(\bv^T\bx+b)\,d\alpha(\bv,b)+c_f\\
&=\int_{\S^{d-1}\times[-1,1]}(\sigma_{\bx}\circ N)(\bv,b)\sqrt{\|\bv\|_2^2+b^2}\,d\alpha(\bv,b)\\
&=\int_{\S^d}\sigma(\bv^T\bx+b)\,d(N_\natural\alpha)(\bv,b),
\end{align*}
where $\sigma_\bx(\bv,b)=\sigma(\bv^T\bx+b)$. We then conclude that $f(\bx)\in\mathcal{G}_1$ with $\gamma_1(f)\le6\|f\|_\cS$, where $\mathcal{G}_1$ and its equipped norm $\gamma_1$ are defined in Section 3 of \citet{bach2017breaking}. Applying Proposition 1 of \citet{bach2017breaking} with $\varepsilon = m^{-(d+3)/(2d)}$ yields the desired result.
\end{proof}

\section{Proofs of lower bounds}\label{sec:lowerboundsupplementray}
In this section, we prove Theorem \ref{thm:minimax} and Proposition \ref{prop:rand_feat}. We begin by introducing some notation. Let $\ell_q=\{\ba=(a_1,a_2,\dots):\sum_{i=1}^\infty|a_i|^q<\infty\}$ for $q\in(0,\infty)$, and $\|\ba\|_{\ell_q}=\bigl(\sum_{i=1}^\infty|a_i|^q\bigr)^{1/q}$ the associated norm. Let $L^2(\mu)=\{f:\int_{\B^d}f^2(\bx)\,d\mu(\bx)<\infty\}$ be the class of square integrable functions with respect to the measure $\mu$. For any functions $f,g\in L^2(\mu)$, let $\|f\|_{\mu,2}=\bigl\{\int_{\B^d}f^2(\bx)\,d\mu(\bx)\bigr\}^{1/2}$ denote the $L^2_\mu$-norm of $f$, and $\langle f,g\rangle_\mu=\int_{\B^d}f(\bx)g(\bx)\,d\mu(\bx)$ the inner product. Let $\cN(\cF,\xi,\|\cdot\|)$ be the cardinality of a minimal $\xi$-covering for the class $\cF$ under some norm $\|\cdot\|$, and $\cP(\cF,\xi,\|\cdot\|)$ the cardinality of the minimal $\xi$-packing for $\cF$ under $\|\cdot\|$.
The proof of the minimax lower bound in Theorem \ref{thm:minimax} makes use of a general result in \cite{yang1999information}.

\begin{theorem}[Theorem 1 of \citet{yang1999information}]\label{thm:low}
Consider the regression model
\[
y_i=f(\bx_i)+\ve_i,\quad i=1,\dots,n,
\]
where $\ve_i\sim N(0,\sigma_\ve^2)$ are independent Gaussian noises and $f\in\cF$ for some function class $\cF$. Suppose that there exist $\delta,\xi>0$ such that
\[
\log\cN(\xi)\le\frac{n\xi^2}{2\sigma_\ve^2},\quad\log\cP(\xi)\ge\frac{2n\xi^2}{\sigma_\ve^2}+2\log2,
\]
where $\cN(\xi)=\cN(\cF,\xi,\|\cdot\|_{\mu,2})$ and $\cP(\xi)=\cP(\cF,\xi,\|\cdot\|_{\mu,2})$. Then we have
\[
\inf_{\widehat{f}}\sup_{f\in\cF}\E\|\widehat{f}-f\|^2_{\mu,2}\ge\frac{\delta^2}{8},
\]
where $\E$ is the expectation with respect to observed samples.
\end{theorem}

Our main idea is to show that a complex enough function class $\cI_\phi(C_1,C_2)$ is a subset of $\cG(\B^d)$, where $\cI_\phi$ denotes the sparse $\ell_1$-approximated set defined by
\begin{equation}\label{eq:sparseset}
\cI_\phi(C_1,C_2)=\biggl\{\sum_{j=1}^{\infty}a_j\phi_j:\sum_{j=1}^\infty|a_j|\le C_1,\sum_{j=K+1}^{\infty}a_j^2\le C_2K^{-1},K\in\N\biggr\},
\end{equation}
and $\{\phi_j\}_{j=1}^\infty$ is a family of orthogonal polynomials with respect to some weight function on $d$-dimensional unit balls. Using technical tools from \cite{hayakawa2020minimax} and by Theorem \ref{thm:low}, we prove that the minimax lower bound for estimating functions in $\cI_\phi(C_1,C_2)$ has the rate $O((n\log n)^{-1/2})$, and thus Theorem \ref{thm:minimax} follows.

Let $W(\cdot)$ be a nonnegative weight function supported on $\B^d$ such that $\int_{\B^d}W(\bx)\,d\bx<\infty$. For each $n\in\N$, the set of polynomials of degree $n$ that are orthogonal to all polynomials of lower degrees forms a vector space of dimension $r_n\equiv\binom{n+d-1}{d}$ \citep{xu1998orthogonal,dunkl2014orthogonal}. A family of orthogonal polynomials with respect to $W(\cdot)$ is defined by
$\{\phi_k^n\}, 1\le k\le r_n^d, n\in\N$, which satisfies
\[
\int_{\B^d}\phi_k^n(\bx)\phi_j^m(\bx)W(\bx)\,d\bx=I(j=k,m=n).
\]
The following lemma shows that all the orthogonal polynomials belong to $\cG(\B^d)$, and hence $\cI_\phi(C_1,C_2)\subset\cG(\B^d)$.

\begin{lemma}\label{lem:orth_poly}
Consider the weight function $W(\bx)=\bigl(\int_{\B^d}d\bx\bigr)^{-1}I(\bx\in\B^d)$. Then the family of orthogonal polynomials with respect to $W(\cdot)$ is a subset of $\cG(\B^d)$.
\end{lemma}

The proof of Lemma \ref{lem:orth_poly} can be found in Section \ref{sec:lem_low}. We are ready to prove Theorem \ref{thm:minimax}.

\begin{proof}[Proof of Theorem \ref{thm:minimax}]
By assumption, the weight function is $W(\bx)=\bigl(\int_{\B^d}d\bx\bigr)^{-1}I(\bx\in\B^d)$. By Lemma \ref{lem:orth_poly}, the family of orthogonal polynomials $\{\phi_k^n\}_{k,n}$ with respect to the weight function is contained in $\cG(\B^d)$. Denote by $\{\phi_j\}_{j=1}^\infty$ the set $\{\phi_k^n\}_{k,n}$ after proper enumeration in an increasing order of $\cS$-norm. By assumption, $\mu$ is chosen to be the uniform distribution of $\B^d$, and thus $\{\phi_j\}_{j=1}^\infty$ satisfies
\[
\frac{1}{\int_{\B^d}1d\bx}\int_{\B^d}\phi_i(\bx)\phi_j(\bx)\,d\bx=I(i=j).
\]
Since $\{\phi_j\}_{j=1}^{\infty}\subset\cG(\B^d)$, we have $\cI_\phi(C_1,C_2)\subset\cG(\B^d)$. We can then focus on the minimax lower bound for functions in $\cI_\phi(C_1,C_2)$ owing to the following inequality:
\begin{equation}\label{thm:inequality}
\inf_{\widehat{f}}\sup_{f^{*}\in\cG(\B^d)}\E|\widehat{f}(\bx)-f^{*}(\bx)|^2\ge\inf_{\widehat{f}}\sup_{f^{*}\in\cI_\phi(C_1,C_2)}\E|\widehat{f}(\bx)-f^{*}(\bx)|^2.
\end{equation}
Consider the covering $\xi$-entropy and the packing $\xi$-entropy of $\cI_\phi(C_1,C_2)$. Since the $\phi_j$s are orthonormal, we need only evaluate $\cN(\cA,\xi,\|\cdot\|_{\ell_2})$, where
\[
\cA=\biggl\{\ba\in\ell_2:\sum_{k=1}^\infty|a_k|\le C_1,\sum_{i=m+1}^\infty a_i^2\le C_2m^{-1},m\in\N\biggr\}.
\]
Following the proof of Lemma 4.7 in \citet{hayakawa2020minimax}, one can show that
\[
C_l\xi^{-2}\le\log\cN(\cA,\xi,\|\cdot\|_{\ell_2})\le C_u\xi^{-2}(1-\log\xi)
\]
for some constants $C_l,C_u>0$. For completeness, we provide the detailed arguments below.

For each integer $q\ge1$, define the sequence $\ba^{(q)}\in\ell_2$ by
\[
a^{(q)}_k=\left\{\begin{array}{cc}
q^{-1},&\quad1\le k\le q,\\
0,&\quad k\ge q+1.
\end{array}\right.
\]
Note that $\sum_{k=1}^{\infty}|a^{(q)}_k|= 1$ and
\[
k\sum_{i=k+1}^\infty(a^{(q)}_i)^2=k(q-k)q^{-2}\le\frac{1}{4}
\]
for all $k=1,\dots,q$. Define $
C=\min(C_1,2\sqrt{C_2})$, and hence each $C\ba^{(q)}$ is an element of $\cA$. For each $q$, define the hyperrectangle
\[
\cA_q=\{\ba\in\ell_2:|a_i|\le Ca_i^{(q)},i\in\N\},
\]
which is a subset of $\cA$. For each pair of distinct vertices of $\cA_q$, the $\ell_2$ distance between them is at least $Cq^{-1}$ for all $q\ge1$. Letting $\delta=Cq^{-1}$ in Lemma \ref{thm:donoho}, we have
\[
\log\cN\biggl(\cA,\frac{C}{2\sqrt{q}},\|\cdot\|_{\ell_2}\biggr)\ge\log\cN\biggl(\cA_q,\frac{C}{2\sqrt{q}},\|\cdot\|_{\ell_2}\biggr)\ge Aq.
\]
As a result, for $\xi\in[C2^{-(j+1)/2-1},C2^{-j/2-1}]$,
\begin{equation}\label{eq:lower_entropy}
\log\cN(\cA,\xi,\|\cdot\|_{\ell_2})\ge\log\cN(\cA,(C/2)2^{-j/2},\|\cdot\|_{\ell_2})\ge\frac{A}{2}\biggl(\frac{2\xi}{C}\biggr)^{-2}\equiv C_l\xi^{-2}.
\end{equation}

We then prove the upper bound. for an arbitrary $\ba\in\cA$, by definition we have
\[
\sum_{i\ge q+1}a_i^2\le C_2q^{-1}.
\]
Let $\bb=(a_1,\dots,a_q,0,0,\dots)$, so that $\|\ba-\bb\|_{\ell_2}^2\le C_2q^{-1}$. Define the $\ell_2$-grid approximation of $\bb$ by $\widetilde\bb=(\sgn(b_i)q^{-1}\lfloor q\,|b_i|\rfloor)_{i=1}^\infty$.  Note that $\|\bb-\widetilde\bb\|_{\ell_2}\le q^{-1/2}$, $\|\ba-\widetilde\bb\|_{\ell_2}\le (\sqrt{C_2}+1)q^{-1/2}$, and the number of such $\widetilde\bb$ is at most $(2C_1q+1)^q$. Letting $q=(\sqrt{C_2}+1)^2\xi^{-2}$, the covering $\xi$-entropy can be upper bounded by
\begin{align*}
\log\cN(\cA,\xi,\|\cdot\|_{\ell_2})&\le\log\{(2C_1q+1)^q\}\\
&\le q\log(2C_1+1)+2q\log{q}\le C_u\xi^{-2}(-\log\xi+1),
\end{align*}
where, for example, we can set $C_u=(\sqrt{C_2}+1)^2\max[4,\log\{(2C_1+1)(\sqrt{C_2}+1)^4\}]$.

Finally, we apply Theorem \ref{thm:low} to complete the proof. Letting $\xi_n=c_\xi(\log n/n)^{1/4}$ for some constant $c_\xi>0$ yields
\begin{align*}
\log\cN(\cA,\xi_n,\|\cdot\|_{\ell_2})&\le C_uc_\xi^{-2}\biggl(\frac{\log n}{n}\biggr)^{-1/2}\biggl\{1+\log\biggl(\frac{n}{\log n}\biggr)\biggr\}\\
&\le2C^uc_\xi^{-2}(n\log n)^{1/2}.
\end{align*}
Then we have
\[
\frac{\log\cN(\cA,\xi_n,\|\cdot\|_{\ell_2})}{n\xi_n^2}\le2C^uc_\xi^{-4}=\frac{1}{2\sigma_\ve^2}
\]
for $c_\xi=(4C_u\sigma_\ve^2)^{1/4}$. On the other hand, letting $\delta_n=c_\xi(n\log n)^{-1/4}$ in \eqref{eq:lower_entropy} for some constant $c_\xi>0$ gives
\[
\log\cP(\cA,\delta_n,\|\cdot\|_{\ell_2})\ge\log\cN(\cA,\delta_n,\|\cdot\|_{\ell_2})\ge C_lc_\xi^{-2}(n\log n)^{1/2}=\frac{C_l}{8C_u}\frac{2n\xi^2_n}{\sigma_\ve^2}.
\]
Since $n\xi_n^2\to\infty$, we have $\{C_l/(8C_u)\}2n\sigma_\ve^{-2}\xi^2_n\ge2n\sigma_\ve^{-2}\xi_n^2+2\log 2$ by choosing some appropriate $C_l$ and $C_u$. By Theorem \ref{thm:low} and \eqref{thm:inequality}, we conclude that
\[
\inf_{\widehat{f}}\sup_{f^{*}\in\cG}\E|\widehat{f}(\bx)-f^{*}(\bx)|^2\ge\frac{C'}{\sqrt{n\log n}},
\]
where $C'=\sigma_\ve\sqrt{C_u}/4$.
\end{proof}

In the rest of this section, we prove Proposition \ref{prop:rand_feat}, whose idea is inspired by the following lemma.

\begin{lemma}[Theorem 6 of \cite{barron1993universal} and Lemma 3.15 of \citet{e2020comparative}]\label{lem:theorem6ofbarron}
Let $\mathscr{F}(f)(\bomega)$ be the Fourier transform of a function $f$ defined on $\R^d$. Define
\[
\Gamma_M=\biggl\{f:\int_{\R^d}\|\bomega\|_1^2|\mathscr{F}(f)(\bomega)|d\bomega<M\biggr\}.
\]
Then, for any $p$ fixed basis functions $h_1,\dots,h_p$,
\[
\sup_{f\in\Gamma_M}\inf_{h\in\Span(h_1,\dots,h_p)}\E_\bx|f(\bx)-h(\bx)|^2\ge\frac{\kappa M}{dp^{1/d}}.
\]
\end{lemma}

The connection between Lemma \ref{lem:theorem6ofbarron} and Proposition \ref{prop:rand_feat} is that $\Gamma_M\subset\cG_M$, which is implied by the following lemma. Let $\cQ(\S^{d-1})$ be the space consisting of all probability distributions defined on the unit sphere $\S^{d-1}$.

\begin{lemma}[Barron space]\label{lem:barron_space}
Let
\[
\cH_{\rho,M}=\biggl\{\int_{\S^{d-1}}a(\bv)\sigma(\bv^T\bx)\,d\rho(\bv):\int_{\S^{d-1}}a^2(\bv)\,d\rho(\bv)\le M^2\biggr\}
\]
for some $\rho\in\cQ(\S^{d-1})$. Then $\cup_{\rho\in\cQ(\S^{d-1})}\cH_{\rho,M}\subset\cG_M$.
\end{lemma}

Lemma 3.15 of \citet{e2020comparative} showed that $\Gamma_M\subset\cup_{\rho\in\cQ(\S^{d-1})}\cH_{\rho,M}$, and hence, by Lemma \ref{lem:barron_space}, $\Gamma_M\subset\cup_{\rho\in\cQ(\S^{d-1})}\cH_{\rho,M}\subset\cG_M$. Now we are ready to prove Proposition \ref{prop:rand_feat}.

\begin{proof}[Proof of Proposition \ref{prop:rand_feat}] Let $\bh(\bx)=(\sigma(\bx^T\bv_1+b_1),\dots,\sigma(\bx^T\bv_m+b_m))^T$. For any prespecified $\lambda$, by the optimality of $\widehat\ba$ we have
\[
h_{\rho_0}(\bx;\widehat\ba(\lambda))=H_{\rho_0,m}(\bx,\bX)\{H_{\rho_0,m}(\bX,\bX)+n\lambda\bI_n\}^{-1}\by,
\]
where $H_{\rho_0,m}(\bx,\bz)=m^{-1}\bh(\bx)^T\bh(\bz)$, $H_{\rho_0,m}(\bx,\bX)$ is a $1\times n$ vector with the $i$th component being $H_{\rho_0,m}(\bx,\bx_i)$, and $H_{\rho_0,m}(\bX,\bX)$ is an $n\times n$ matrix with the $(i,j)$th entry being $H_{\rho_0,m}(\bx_i,\bx_j)$. In the following, we show that $h_{\rho_0}(\bx;\widehat\ba(\lambda))$ lies in the space spanned by at most $\min(n,m)$ basis functions, and thus the desired result follows from Lemma \ref{lem:theorem6ofbarron}.

Define the linear functional $\cT\colon L^{2}(\mu)\to L^{2}(\mu)$ by
\begin{align*}
(\cT f)(\bx)&=\int_{\B^d}f(\bz)H_{\rho_0,m}(\bx,\bz)\,d\mu(\bz)\\
&=\frac{1}{m}\sum_{k=1}^m\sigma(\bx^T\bv_k+b_k)\int_{\B^d}f(\bz)\sigma(\bz^T\bv_k+b_k)\,d\mu(\bz).
\end{align*}
By assumption, $\{\bv_k\}_k\subset\S^{d-1}$ and $\{b_k\}_k\subset[-1,1]$, and hence
\[
\sup_{\bx\in\B^d}\max_{1\le k\le m}|\sigma(\bx^T\bv_k+b_k)|\le2.
\]
By the triangle inequality and the Cauchy--Schwarz inequality, we have
\[
\|\cT f\|_{\mu,2}\le\frac{1}{m}\sum_{k=1}^{m}|\langle f(\bz),\sigma(\bz^T\bv_k+b_k)\rangle_\mu|\|\sigma(\bx^T\bv_k+b_k)\|_{\mu,2}\le4\|f\|_{\mu,2},
\]
which further implies that $\cT$ is bounded. For any $f,g\in L^2(\mu)$,
\[
\langle\cT g,f\rangle_\mu=\frac{1}{m}\sum_{k=1}^{m}\langle f(\bx),\sigma(\bx^T\bv_k+b_k)\rangle_\mu\langle g(\bx),\sigma(\bx^T\bv_k+b_k)\rangle_\mu=\langle\cT f,g\rangle_\mu,
\]
which, together with the boundedness of $\cT$, implies that $\cT$ is self-adjoint. Thus, there exist $e_1,\dots,e_p$ with $p$ possibly infinite such that $(\cT e_i)(\bx)=\lambda_ie_i(\bx)$ with $|\lambda_1|\ge|\lambda_2|\ge\dots\ge|\lambda_p|>0$ and $\langle e_i,e_j\rangle_\mu=I(i=j)$.

Our first claim is that $p\le m$. If not, choose $e_1,\dots,e_{m+1}$ corresponding to the top $m+1$ absolutely largest eigenvalues, and set $\bLambda=\diag(\lambda_1,\dots,\lambda_{m+1})$. For any $j=1,\dots,m+1$ and $k=1,\dots,m$, define $a_{jk}=m^{-1}\langle\sigma(\bz^T\bv_k+b_k),e_j(\bz)\rangle_\mu$. Write $\bA=(a_{jk})\in\R^{(m+1)\times m}$. Since $(\cT e_i)(\bx)=\lambda_ie_i(\bx)$ and $\langle e_i,e_j\rangle_\mu=I(i=j)$, we obtain $(e_1(\bx),\dots,e_{m+1}(\bx))^T=\bLambda\bA\bh(\bx)$ and further
\begin{equation}\label{eq:rank_contradict}
\bI_{m+1}=\bLambda\bA\int_{\B^d}\bh(\bx)\bh(\bx)^Td\mu(\bx)\bA^T\bLambda^T,
\end{equation}
where $\bI_{m+1}$ denotes the identity matrix. Note that $\rank(\bA)\le m$, and thus the right-hand side of \eqref{eq:rank_contradict} has rank at most $m$, leading to a contradiction.

Our second claim is that $h_{\rho_0}(\bx;\widehat\ba(\lambda))$ can be expressed as a linear combination of $e_1(\cdot),\dots,e_p(\cdot)$. Suppose that $\bh(\bx)=\sqrt{m}\bB\be(\bx)$ with $\be(\bx)=(e_1(\bx),\dots,e_{p+1}(\bx))^T$, where $e_{p+1}(\cdot)$ denotes any function such that $\langle e_{p+1},e_j\rangle_\mu=0$ for $j=1,\dots,p$ and $\|e_{p+1}\|_{\mu,2}=1$. Let $\bI_{p+1,j}$ be the $j$th column of the identity matrix $\bI_{p+1}$. By definition,
\begin{equation}\label{eq:H_express}
H_{\rho_0,m}(\bx,\bz)=m^{-1}\bh(\bx)^T\bh(\bz)=\be(\bx)^T\bB^T\bB\be(\bz).
\end{equation}
Using the facts that $\cT(e_j)=\lambda_je_j$ for $j=1,\dots,p$ and $\cT(e_{p+1})=0$, we obtain
\[
\be(\bx)^T\bB^T\bB\bigl\langle\be(\bz),e_j(\bz)\bigr\rangle_\mu=\lambda_j\be(\bx)^T\bI_{p+1,j}I(j\ne p+1)
\]
for all $j$. Since $\langle\be(\bz),e_j(\bz)\rangle_\mu=\bI_{p+1,j}$, this equality holds only if $\bB^T\bB\bI_{p+1,j}=\lambda_jI(j\ne p+1)$ for all $j$; that is, $\bB^T\bB=\diag(\lambda_1,\dots,\lambda_p,0)$. By \eqref{eq:H_express},
\[
H_{\rho_0,m}(\bx,\bz)=\be(\bx)^T\bB^T\bB\be(\bz)=\sum_{i=1}^p\lambda_i e_i(\bx)e_i(\bz).
\]
Let $\bLambda_0=\mbox{diag}(\lambda_1,\dots,\lambda_p)$ and $\bE(\bX)=(e_i(\bx_j))\in\R^{p\times n}$. Then we obtain $H_{\rho_0,m}(\bX,\bX)=\bE(\bX)^T\bLambda_0\bE(\bX)$, and consequently $h_{\rho_0}(\bx;\widehat\ba(\lambda))=\sum_{i=1}^pc_i(\lambda)e_i(\bx)$, where
\[
\bc(\lambda)=(c_1(\lambda),\dots,c_p(\lambda))^T=\bLambda_0\bE(\bX)\{\bE(\bX)^T\bLambda_0\bE(\bX)+n\lambda\bI_n\}^{-1}\by.
\]

Our last claim is that when $m\ge p>n$, $h_{\rho_0}(\bx;\widehat\ba(\lambda))$ can be expressed as a linear combination of $n$ basis functions. For simplicity, we write $\bc(\lambda)$ as $\bc$, but keep in mind that its coefficients $c_i$ are functions of $\lambda$. First, observe that when $m>n$, $r=\rank(\bE(\bX))\le n$. If $p>n$, then there exists some matrix $\bC\in\R^{p\times n}$ such that $\bc=\bC\bd$, where $\bd\in\R^n$. For example, one can choose $\bC=\bLambda_0\bQ_e\in\R^{p\times n}$ and $\bd=\bD_e\bR_e^T\{\bE(\bX)^T\bLambda_0\bE(\bX)+n\lambda\bI_n\}^{-1}\by\in\R^n$, where $\bE(\bX)=\bQ_e\bD_e\bR_e^T$ is the singular value decomposition of $\bE(\bX)$ with $\bD_e\in\R^{n\times n}$ being a diagonal matrix whose first $r$ diagonal entries are nonzero. Let $\bu_i=(u_{i1},\dots,u_{ip})^T\in\R^p$ be the $i$th orthonormal basis of the column space of $\bC$, and let
\[
\widetilde{e}_k(\bx)=u_{k1}e_1(\bx)+\dots+u_{kp}e_p(\bx),\quad k=1,\dots,n.
\]
Since $\int_{\B^d}\widetilde{e}_i(\bx)\widetilde{e}_j(\bx)\,d\mu(\bx)=\bu_i^T\bu_j=I(i=j)$, we obtain
\[
g_{m,\rho_0}(\bx;\widehat\ba)=\sum_{i=1}^nc_i'\widetilde{e}_i(\bx),
\]
where $(c_1',\dots,c_n')^T=\bO\bd$ with $\bO=\bU^T\bC$ and $\bU=(\bu_1,\dots,\bu_n)$.

Combining these claims, we conclude that $h_{\rho_0}(\bx;\widehat\ba(\lambda))$ is a linear combination of at most $\min(n,m)$ fixed basis functions. By Lemma \ref{lem:theorem6ofbarron}, we have
\begin{align*}
&\sup_{f^{*}\in\cG_M}\inf_{\lambda>0}\E\left\|f^{*}-g_{m,\rho_0}(\cdot;\widehat\ba(\lambda))\right\|_2^2\\
&\quad\ge\sup_{f\in\Gamma_M}\inf_{g\in\Span(e_1,\dots,e_{q})}\E|f^*(\bX)-g(\bX)|^2\ge\frac{\kappa M}{d\{\min{(m,n)}\}^{1/d}},
\end{align*}
for some universal constant $\kappa>0$.
\end{proof}

\section{Technical lemmas}\label{sec:tech_lemmas}

\subsection{Lemmas used to prove the upper bounds}\label{sec:upper_lemma_proof}
We first introduce a useful lemma to control the expectation of Gaussian maxima.

\begin{lemma}\label{lem:maxima_Gauss}
Suppose that $\ve_1,\dots,\ve_n$ are sub-Gaussian random variables such that $\E e^{\gamma\ve_i}\le e^{\sigma_\ve^2\gamma^2/2}$ for all $\gamma$. Then $\E\max_i|\ve_i|\le\sigma_\ve\sqrt{2\log(2n)}$.
\end{lemma}

\begin{proof}
By assumption,
\begin{align*}
\E\max_i|\ve_i|&=\frac{1}{\gamma}\E\log\Bigl\{\exp\Bigl(\gamma\max_i|\ve_i|\Bigr)\Bigr\}\le\frac{1}{\gamma}\E\log\biggl\{\sum_{i=1}^n\exp(\gamma|\ve_i|)\biggr\}\\
&\le\frac{1}{\gamma}\log\E\biggl\{\sum_{i=1}^n(e^{\gamma\ve_i}+e^{-\gamma\ve_i})\biggr\}=\frac{\log(2n)}{\gamma}+\frac{\sigma_\ve^2\gamma}{2}.
\end{align*}
Choosing $\gamma=\sigma_\ve^{-1}\sqrt{2\log(2n)}$ yields the desired result.
\end{proof}

The next lemma gives an upper bound on $\log\cN_m(\delta)$.

\begin{lemma}\label{lem:metric_entropy}
The metric entropy of $\cF(m,1)$ with respect to  the supremum norm is upper bounded by
\[
\log\cN(\delta,\cF(m,1),\|\cdot\|_\infty)\le(d+1)m\log(1+4\delta^{-1}).
\]
\end{lemma}

\begin{proof}
For any two-layer ReLU networks $g(\cdot;\btheta_1),g(\cdot;\btheta_2)\in\cF(m,1)$ with parameters $\btheta_1=\bigl(a_1^{(1)},\dots,a_m^{(1)},\bw_1^{(1)T},\dots,\bw_m^{(1)T}\bigr)^T$ and $\btheta_2=\bigl(a_1^{(2)},\dots,a_m^{(2)},\bw_1^{(2)T},\dots,\bw_m^{(2)T}\bigr)^T$, we can assume without loss of generality that $\|\bw_i^{(j)}\|_2=1$ for all $i,j$, in which case $g(\bx;\btheta_j)\in\cF(m,1)$ is equivalent to $\sum_{k=1}^m|a_k^{(j)}|\le1$ for both $j$. Then letting $(\bx^T,1)^T=\widetilde\bx$ gives
\begin{align*}
&|g(\bx;\btheta_1)-g(\bx;\btheta_2)|\\
&\quad=\biggl|\sum_{k=1}^m a_k^{(1)}\sigma(\widetilde\bx^T\bw_k^{(1)})-\sum_{k=1}^ma_k^{(2)}\sigma(\widetilde\bx^T\bw_k^{(2)})\biggr|\\
&\quad\le\biggl|\sum_{k=1}^m(a_k^{(1)}-a_k^{(2)})\sigma(\widetilde\bx^T\bw_k^{(1)})\biggr|+\biggl|\sum_{k=1}^ma_k^{(2)}\bigl(\sigma(\widetilde\bx^T\bw_k^{(1)})-\sigma(\widetilde\bx^T\bw_k^{(2)})\bigr)\biggr|\\
&\quad\le\max_{1\le k\le m}\sigma(\widetilde\bx^T\bw_k^{(1)})\sum_{k=1}^m|a_k^{(1)}-a_k^{(2)}|+2\sum_{k=1}^m|a_k^{(2)}|\max_{1\le k\le m}\|\bw_k^{(1)}-\bw_k^{(2)}\|_2\\
&\quad\le\sum_{k=1}^m|a_k^{(1)}-a_k^{(2)}|+2\max_{1\le k\le m}\|\bw_k^{(1)}-\bw_k^{(2)}\|_2.
\end{align*}
Denote the unit $\ell_1$-ball in $\R^n$ by $\B^n_1(1)$. To cover $\cF(m,1)$ with respect to $\|\cdot\|_\infty$, we need only cover $\B^m_1(1)$ with respect to $\|\cdot\|_1$ and $m$ many $\B^d$ with respect to $\|\cdot\|_2$ simultaneously. Using a volume argument as in Lemma 5.7 of \citet{wainwright2019high} yields
\[
\cN(\delta,\B^m_1(1),\|\cdot\|_1)\le(1+2\delta^{-1})^m,\quad\cN(\delta,\B^d,\|\cdot\|_2)\le(1+2\delta^{-1})^d,
\]
and thus
\[
\log\cN(\delta,\cF(m,1),\|\cdot\|_\infty)\le(d+1)m\log(1+4\delta^{-1}).\qedhere
\]
\end{proof}

In the following two lemmas, let $\vrho$ be a Rademacher random variable with $P(\vrho=1)=P(\vrho=-1)=1/2$, and let $\vrho_1,\dots,\vrho_n$ be independent copies of $\vrho$.

\begin{lemma}[Lemma 26.2 of \citet{shalev2014understanding}]\label{thm:262}
Let $\bx_1,\dots,\bx_n$ be independent copies of $\bx$. For any function class $\cF$, we have
\[
\E\sup_{f\in\cF}\biggl(\E f(\bx)-\frac{1}{n}\sum_{k=1}^nf(\bx_k)\biggr)\le\E\biggl\{\frac{2}{n}\E_{\vrho}\sup_{f\in\cF}\biggl(\sum_{k=1}^n\vrho_kf(\bx_k)\biggr)\biggr\}.
\]
\end{lemma}

\begin{lemma}\label{thm:rademachercomplexity}
Let $\bx_1,\dots,\bx_n$ be any vectors such that $\max_i\|\bx_i\|_2\le1$. For any $m\ge1$, we have
\[
\E_\vrho\sup_{f\in\cF(m,F)}\biggl(\sum_{k=1}^n\vrho_k f(\bx_k)\biggr)\le2F\sqrt{n}.
\]
\end{lemma}

\begin{proof}
By definition,
\begin{align*}
&\E_{\vrho}\sup_{f\in\cF(m,F)}\biggl(\sum_{k=1}^n\vrho_k f(\bx_k)\biggr)\\
&\quad=\E_{\vrho}\sup_{\nu(\btheta)\le F}\biggl(\sum_{i=1}^n\vrho_i\sum_{k=1}^{m}a_k\sigma(\bw_k^T\bx_i)\biggr)\\
&\quad\le\E_{\vrho}\sup_{\nu(\btheta)\le F}\sup_{\|\bu\|_2=1}\biggl(\sum_{i=1}^n\vrho_i\sum_{k=1}^{m} a_k \|\bw_k\|_2\sigma(\bu^T\bx_i)\biggr)\\
&\quad\le F\E_{\vrho}\sup_{\|\bu\|_2=1}\biggl(\sum_{i=1}^n\vrho_i\sigma(\bu^T\bx_i)\biggr)\le2F\E_{\vrho}\sup_{\|\bu\|_2=1}\biggl(\sum_{i=1}^n\vrho_i\bu^T\bx_i\biggr)\le 2F\sqrt{n},
\end{align*}
where the last line follows from Lemmas 26.9 and 26.10 of \citet{shalev2014understanding}.
\end{proof}

The rest of this subsection is devoted to the proofs of Lemmas \ref{lem:talagrand} and \ref{lem:maximal_small_width}.

\begin{proof}[Proof of Lemma \ref{lem:talagrand}]
By a standard symmetrization argument,
\begin{equation}\label{eq:A1}
\E Z_n\le2\E_{\vrho,\bx}\sup_{f\in\cF^*(m,1)}\biggl|\frac{1}{n}\sum_{i=1}^n\vrho_if^2(\bx_i)\biggr|,
\end{equation}
where $\vrho_i$ are independent Rademacher variables. Since $\phi(x)=x^2$ is $2$-Lipschitz continuous for $x\in[-1,1]$ and $\sup_{f\in\cF^*(m,1)}\sup_{\bx\in\B^d}|f(\bx)|\le2$, by Lemma 26.9 of \citet{shalev2014understanding} we have
\[
\E Z_n\le2\E_{\vrho,\bx}\sup_{f\in\cF^*(m,1)}\biggl|\frac{1}{n}\sum_{i=1}^n\vrho_i\phi(f(\bx_i))\biggr|\le16\E_{\vrho,\bx}\sup_{f\in\cF^*(m,1)}\biggl|\frac{1}{n}\sum_{i=1}^n\vrho_if(\bx_i)\biggr|.
\]
Let $\widetilde{f}\in\cF(\widetilde{m},1)$ be the two-layer ReLU network with $\widetilde{m}$ hidden units that best approximates $f^*$ under the $L_\infty(\B^d)$-norm in Theorem \ref{thm:approx}, where $\widetilde{m}\ge n^{(d+3)/d}$. Thus, $\|\widetilde{f}-f^*\|_{L_\infty(\B^d)}\le C_1\|f^*\|_\cS \widetilde{m}^{-(d+3)/(2d)}\le C_1/\sqrt{n}$ for some constant $C_1>0$. By decomposing $f=f-\widetilde{f}+\widetilde{f}$ and noting that $f-\widetilde{f}\in\cF(m+\widetilde{m},2)$, we obtain
\begin{align*}
\E Z_n&\le16\E_{\vrho,\bx}\sup_{f\in\cF(m,1)}\biggl|\frac{1}{n}\sum_{i=1}^n\vrho_i\bigl(f(\bx_i)-\widetilde{f}(\bx_i)\bigr)\biggr| +16\E_{\vrho,\bx}\biggl|\frac{1}{n}\sum_{i=1}^n\vrho_i\bigl(\widetilde{f}(\bx_i)-f^*(\bx_i)\bigr)\biggr|\\
&\le16\E_{\vrho,\bx}\sup_{f\in\cF(m+\widetilde{m},2)}\biggl|\frac{1}{n}\sum_{i=1}^n\vrho_if(\bx_i)\biggr|+\frac{16}{n}\sum_{i=1}^n\sqrt{\E_{\vrho}\vrho_i^2}\|\widetilde{f}-f^*\|_2\\
&\le16\E_{\vrho,\bx}\sup_{f\in\cF(m+\widetilde{m},2)}\biggl|\frac{1}{n}\sum_{i=1}^n\vrho_if(\bx_i)\biggr|+\frac{16C_1}{\sqrt{n}}\le\frac{16C_1+64}{\sqrt{n}}\equiv\frac{C_{\cF}}{\sqrt{n}},
\end{align*}
where the last inequality follows from Lemma \ref{thm:rademachercomplexity}.

Define
\[
U=\sup_{\bx\in\B^d}\sup_{f\in\cF^*(m,1)}|f(\bx)|^2,\quad\eta^2=\sup_{f\in\cF^*(m,1)}\E|f(\bx)|^4,\quad K_n=2U\E Z_n+\eta^2,
\]
and note that for $\sqrt{n}\ge C_{\cF}$,
\[
U\le2\sup_{\bx\in\B^d}\sup_{f\in\cF(m,1)}(|f(\bx)|^2+|f^*(\bx)|^2)\le4,\quad\eta^2\le U^2\le16,\quad K_n\le\frac{8C_{\cF}}{\sqrt{n}}+16\le24.
\]
By Talagrand's concentration inequality \citep[Theorem 3.27]{wainwright2019high},
\[
P(Z_n-\E Z_n\ge t)\le2\exp\biggl(-\frac{nt^2}{8eK_n+4Ut}\biggr)\le2\exp\biggl(-\frac{nt^2}{192e+16t}\biggr).
\]
Note that $-nt^2/(192e+16t)\le-nt/32$ if $t\ge12e$, and $-nt^2/(192e+16t)\le-nt^2/(384e)$ otherwise. We then conclude that
\[
P\biggl(Z_n\ge\frac{C_{\cF}}{\sqrt{n}}+t\biggr)\le\exp\biggl\{-\frac{n}{32}\min\biggl(\frac{t^2}{12e},t\biggr)\biggr\}.\qedhere
\]
\end{proof}

\begin{proof}[Proof of Lemma \ref{lem:maximal_small_width}]
First note that $\sup_{\bx\in\B^d}\sup_{f\in\cF^*(m,1)}|f(\bx)|\le2$. By \eqref{eq:A1}, we have
\[
\E Z_n(\gamma)\le16\E_{\vrho,\bx}\sup_{f\in\cB_\cF(\gamma)}\frac{1}{n}\biggl|\sum_{i=1}^n\vrho_if(\bx_i)\biggr|,
\]
where $\vrho_i$ are independent Rademacher variables. Let $\{g_j\}_{j=1}^M$ be a minimal $(1/n)$-covering of $\cB_\cF(\gamma)$ with respect to the $L_\infty(\B^d)$-norm. For a given $f\in\cB_\cF(\gamma)$, let $g_{j^*}$ be the function closest to $f$.  By the triangle inequality, we obtain
\begin{equation}\label{eq:AA2}
\begin{split}
\biggl|\sum_{i=1}^n\vrho_if(\bx_i)\biggr|&\le\biggl|\sum_{i=1}^n\vrho_i\bigl(f(\bx_i)-g_{j^*}(\bx_i)\bigr)\biggr|+\max_{1\le j\le M}\biggl|\sum_{i=1}^n\vrho_ig_j(\bx_i)\biggr|\\
&\le1+\max_{1\le j\le M}\biggl|\sum_{i=1}^n\vrho_i\frac{g_j(\bx_i)}{\|g_j\|_n}\biggr|\sqrt{\max_{1\le j\le M}\|g_j\|_n^2}\\
&\equiv1+I_1\sqrt{I_2}.
\end{split}
\end{equation}
Since $\vrho_i$ are sub-Gaussian with $\E e^{\gamma\vrho_i}\le e^{\gamma^2/2}$ for all $\gamma$, it follows from Lemma \ref{lem:maxima_Gauss} that
\[
\E I_1\le\sqrt{2n\log(2M)}.
\]
Moreover, since $g_j\in\cB_\cF(\gamma)$, we have $\max_j\|g_j\|_2\le\gamma$, and thus
\[
I_2\le\gamma^2+\max_j\bigl|\|g_j\|_n^2-\|g_j\|_2^2\bigr|.
\]
Note that $\max_j\sup_\bx|g_j(\bx)|\le2$ and $\Var(|g_j(\bx)|^2)\le\E|g_j(\bx)|^4\le4\gamma^2$. Applying Bernstein's inequality \citep[Lemma 2.2.9]{van1996weak} and the union bound yields
\[
P\Bigl(\max_j\bigl|\|g_j\|_n^2-\|g_j\|_2^2\bigr|\ge t\gamma\Bigr)\le2M\exp\biggl(-\frac{nt^2}{16t/(3\gamma)+8}\biggr)\le 2M\exp\biggl(-\frac{nt\gamma}{6}\biggr)
\]
for $t\ge12\gamma$. Using the identity $\E X=\int_0^\infty P(X\ge t)\,dt$ for nonnegative $X$ gives, for $M\ge4$,
\begin{align*}
\E\max_j\bigl|\|g_j\|_n^2-\|g_j\|_2^2\bigr|/\gamma&\le\int_0^{12\gamma}1\,dt+\int_{12\gamma}^\infty2M\exp\biggl(-\frac{nt\gamma}{6}\biggr)\,dt\\
&\le12\gamma+\frac{12M}{n\gamma}e^{-2n\gamma^2}\le15\gamma
\end{align*}
since $\gamma\ge\sqrt{\log M/(2n)}$, which is due to the assumption $\gamma\ge\sqrt{6md\log n/n}$ and the fact that $\log M\le12md\log n$ to be shown later. Combining these pieces, by Jensen's inequality we have
\begin{equation}\label{eq:AA3}
\frac{1}{n}\E_{\vrho,\bx}(I_1\sqrt{I_2})=\frac{1}{n}\E_\bx\{\E_{\vrho}(I_1\mid(\bx_i)_i)\sqrt{I_2}\}\le8\gamma\sqrt{\frac{\log M}{n}}.
\end{equation}
It remains to find a $(1/n)$-covering of $\cB_\cF(\gamma)$ with respect to the $L_\infty(\B^d)$-norm. Consider a $(1/n^{3/2})$-covering of $\cF^*(m,1)$ with respect to the $L_\infty(\B^d)$-norm, which we denote by $\{f_j\}_{j=1}^{M'}$. In the following, we prove that $\{f_j/\max(\|f_j\|_2/\gamma,1)\}_{j=1}^{M'}$ is a $(2/n)$-covering of $\cB_\cF(\gamma)$.

Since $\cB_\cF(\gamma)\subset\cF^*(m,1)$, for any $f\in\cB_\cF(\gamma)$ there exists some $g_j\subset\{f_j\}_{j=1}^{M'}$ such that $\|f-g_j\|_\infty\le1/n^{3/2}$, and hence
\begin{equation}\label{eq:bound_2}
\bigl|\|g_j\|_2-\|f\|_2\bigr|\le\|g_j-f\|_2\le\frac{1}{n^{3/2}}.
\end{equation}
If $\|g_j\|_2\le\gamma$, then $g_j$ also belongs to $\{f_j/\max(\|f_j\|_2/\gamma,1)\}_{j=1}^{M'}$. If $\|g_j\|_2>\gamma$, then by the triangle inequality, \eqref{eq:bound_2}, and the assumption that $\gamma\ge1/\sqrt{n}$ we have
\[
\biggl|f-\frac{\gamma g_j}{\|g_j\|_2}\biggr|\le|f-g_j|+\frac{\bigl|\|g_j\|_2-\gamma\bigr|}{\|g_j\|_2}|g_j|\le\frac{1}{n^{3/2}}+\frac{n^{-3/2}}{1/\sqrt{n}}\le\frac{2}{n},
\]
where we have used the fact that $0<\|g_j\|_2-\gamma\le\|g_j\|_2-\|f\|_2$. By Lemma \ref{lem:metric_entropy}, we have
\begin{align*}
\log M&\le\log\cN\bigl(1/(2n)^{3/2},\cF(m,1),\|\cdot\|_\infty\bigr)\le(d+1)m\log(1+12n^{3/2})\\
&\le2md\log(13n^{3/2})\le6md+3md\log n\le12md\log n.
\end{align*}
Substituting into \eqref{eq:AA3} yields
\[
\frac{1}{n}\E_{\vrho,\bx}(I_1\sqrt{I_2})\le16\gamma\sqrt{\frac{3md\log n}{n}}\le29\gamma\sqrt{\frac{md\log n}{n}}.
\]
Combining with \eqref{eq:A1} and \eqref{eq:AA2}, we conclude that
\[
\E Z_n(\gamma)\le16\biggl(\frac{1}{n}+29\gamma\sqrt{\frac{md\log n}{n}}\biggr)\le480\gamma\sqrt{\frac{md\log n}{n}},
\]
where we have used the fact that $1/n\le\gamma\sqrt{md\log n/n}$.
\end{proof}

\subsection{Lemmas used to prove the lower bounds}\label{sec:lem_low}
The following lemma is useful for lower bounding the metric entropy of the set defined in \eqref{eq:sparseset}. We state it without proof.

\begin{lemma}[Lemma 4 of \cite{donoho1993unconditional}]\label{thm:donoho}
Let $C_k\subset \ell_2$ be a $k$-dimensional hypercube of side $2\delta>0$ defined by
\[
C_k=\{\ba\in\ell_2:\max(|a_1|,\dots,|a_k|)\le\delta,|a_{k+1}|=|a_{k+2}|=\dots=0\}.
\]
Then there exists some constant $A>0$ such that
\[
\log\cN\biggl(C_k,\frac{\delta\sqrt{k}}{2},\|\cdot\|_{\ell_2}\biggr)\ge Ak
\]
for all $k\in\N$.
\end{lemma}

We then prove Lemma \ref{lem:orth_poly}.

\begin{proof}[Proof of Lemma \ref{lem:orth_poly}]
Consider the smooth function
\[
h_{\ve}(\bx)=\ve^{-d}\exp\biggl(-\frac{\ve^2}{\ve^2-\|\bx\|_2^2}\biggr)I(\|\bx\|_2<\ve),
\]
known as a mollifier. The cut-off function $I_{\B^d}*h_\ve$ for the set $\B^d$ is defined as a convolution of the indicator function of $\B^d$ and $h_\ve$. Then, for a sufficiently small $\ve$,
\[
I_{\B^d}*h_\ve\in C_c^\infty(\B^d_2(2)),\quad I(\bx\in\B^d)*h_{\ve}(\bx)=1\text{ for all }\bx\in\B^d,
\]
where $C_c^\infty(\B^d_2(2))$ denotes the class of smooth functions compactly supported on $\B^d_2(2)$. Define $g_{\ve,k,n}(\bx)=\phi_k^n(\bx)\{I(\bx\in\B^d)*h_{\ve}(\bx)\}$, and hence $g_{\ve,k,n}\in C_c^\infty(\B^{d}_2(2))$. It follows from Proposition 3 of \citet{ongie2020function} that $\|g_{\ve,k,n}\|_{\cR}<\infty$. By Proposition \ref{prop:functionspace2}, there exists some signed measure $\alpha_{\ve,k,n}$ with $\int_{\R^d}\|\bv\|_2\,d|\alpha_{\ve,k,n}|(\bv,b)<\infty$ such that
\[
g_{\ve,k,n}(\bx)=\int_{\R^d}\bigl(\sigma(\bv^T\bx+b)-\sigma(b)\bigr)\,d\alpha_{\ve,k,n}(\bv,b)+c
\]
for all $\bx\in\R^d$. If we restrict $\bx$ to $\B^d$, then
\[
\phi_k^n(\bx)=g_{\ve,k,n}(\bx)=\int_{\R^{d}}\bigl(\sigma(\bv^T\bx+b)-\sigma(b)\bigr)\,d\alpha_{\ve,k,n}(\bv,b)+c,
\]
for all $1\le k\le r_n^d$ and $n$. We conclude that $\{\phi_k^n\}_{k,n}\subset \cG(\B^d)$.
\end{proof}

We close this subsection with the proof of Lemma \ref{lem:barron_space}.

\begin{proof}[Proof of Lemma \ref{lem:barron_space}]
For any $\rho\in\cQ(\S^{d-1})$, define the measure $\alpha_{a,\rho}(\bv,b)$ on $\S^{d-1}\times[-1,1]$ such that for any Borel measurable set $A\subset\S^{d-1}\times[-1,1]$,
\[
\int_{\S^{d-1}\times[-1,1]}I\{(\bv^T,b)^T\in A\}\,d\alpha_{a,\rho}(\bv,b)=\int_{\S^{d-1}}I\{b=0,(\bv^T,b)^T\in A\}a(\bv)\,d\rho(\bv).
\]
By the Cauchy--Schwarz inequality, we have
\[
\int_{\S^{d-1}\times[-1,1]}\|\bv\|_2\,d|\alpha_{a,\rho}|(\bv,b)\le\int_{\S^{d-1}}|a(\bv)|\,d\rho(\bv)\le\biggl(\int_{\S^{d-1}}|a(\bv)|^2\,d\rho(\bv)\biggr)^{1/2}\le M,
\]
and for any $\bx\in\B^d$,
\begin{align*}
&\int_{\S^{d-1}\times[-1,1]}\bigl(\sigma(\bv^T\bx+b)-\sigma(b)\bigr)\,d\alpha_{a,\rho}(\bv,b)\\
&\quad=\int_{\S^{d-1}\times[-1,1]}I(b\ne0)\bigl(\sigma(\bv^T\bx+b)-\sigma(b)\bigr)\,d\alpha_{a,\rho}(\bv,b)\\
&\quad\relph{}+\int_{\S^{d-1}\times[-1,1]}I(b=0)\bigl(\sigma(\bv^T\bx+b)-\sigma(b)\bigr)\,d\alpha_{a,\rho}(\bv,b)\\
&\quad=\int_{\S^{d-1}}a(\bv)\sigma(\bv^T\bx)\,d\rho(\bv).
\end{align*}
We conclude that $\cH_{\rho,M}\subset\cG_M$ for any $\rho\in\cQ(\S^{d-1})$. Then $\cup_{\rho\in\cQ(\S^{d-1})}\cH_{\rho,M}\subset\cG_M$.
\end{proof}

\bibliographystyle{imsart-nameyear}
\bibliography{tradeoff}